\documentclass[twoside,11pt]{article}

\usepackage[nohyperref,preprint]{jmlr2e}

\usepackage{mdframed}
\usepackage{booktabs}
\usepackage{multirow}
\usepackage{subcaption}
\usepackage{verbatim}
\usepackage{amssymb}
\usepackage{pifont}
\usepackage{tablefootnote}
\usepackage{enumitem}
\usepackage{amsmath}
\usepackage{amssymb}
\usepackage{amsfonts}
\usepackage{dsfont}
\usepackage{bm}
\usepackage{xcolor}
\usepackage[T1]{fontenc}

\usepackage[colorlinks=False, pdfborder={0 0 0}]{hyperref}

\newcommand*{\system}{\textcolor[HTML]{D55E00}}
\newcommand*{\fsexample}{\textcolor[HTML]{029E73}} 
\newcommand*{\question}{\textcolor[HTML]{CC78BC}} 

\newcommand*{\llamathreeseventy}{\textcolor[HTML]{0173B2}}
\newcommand*{\llamatwoseventy}{\textcolor[HTML]{DE8F05}}

\newcommand*{\mistral}{\textcolor[HTML]{CC78BC}}
\newcommand*{\llamatwosevenchat}{\textcolor[HTML]{CA9161}}
\newcommand*{\llava}{\textcolor[HTML]{BCBD22}}
\newcommand*{\flamingo}{\textcolor[HTML]{17BECF}}

\newcommand{\cmark}{\ding{51}}%
\newcommand{\tri}{\ding{115}}%
\newcommand{\starr}{\ding{72}}%

\newcommand*{\navytriangle}{\color[HTML]{1A85FF}{\tri}}
\newcommand*{\magentastar}{\color[HTML]{D41159}{\starr}}

\newmdenv[
    backgroundcolor=gray!10,
    linecolor=black,
    roundcorner=5pt,
    innerleftmargin=10pt,
    innerrightmargin=10pt,
    innertopmargin=10pt,
    innerbottommargin=10pt,
    frametitlebackgroundcolor=gray!30,
    frametitlefont=\bfseries
]{custombox}

\usepackage{lastpage}
\jmlrheading{24}{2025}{1-\pageref{LastPage}}{2/25}{}{21-0000}{Daniel P. Jeong, Pranav Mani, Saurabh Garg, Zachary C. Lipton, Michael Oberst}

\ShortHeadings{Limited Impact of Medical Adaptation of LLMs/VLMs}{Jeong, Mani, Garg, Lipton, Oberst}
\firstpageno{1}

\begin{document}

\title{The Limited Impact of Medical Adaptation of Large Language and Vision-Language Models}

\author{\name Daniel P. Jeong\textsuperscript{1} \email danielje@cs.cmu.edu
\AND
\name Pranav Mani\textsuperscript{2} \email pranav@abridge.com
\AND
\name Saurabh Garg\textsuperscript{3}\email garg.saurabh.2014@gmail.com
\AND
\name Zachary C. Lipton\textsuperscript{1,2} \email zlipton@cmu.edu
\AND
\name Michael Oberst\textsuperscript{2,4} \email moberst@jhu.edu
\AND
\addr \textsuperscript{1}Machine Learning Department, Carnegie Mellon University\\
\addr \textsuperscript{2}Abridge\\
\addr \textsuperscript{3}Mistral AI\\
\addr \textsuperscript{4}Department of Computer Science, Johns Hopkins University
}

\editor{My editor}

\maketitle

\vspace{35pt}

\begin{abstract}
Several recent works seek to adapt general-purpose large language models (LLMs) and vision-language models (VLMs) for medical applications through continued pretraining on publicly available biomedical corpora. 
These works typically claim that such domain-adaptive pretraining improves performance on various downstream medical tasks, such as answering medical exam questions. 
In this paper, we compare ten ``medical'' LLMs and two VLMs against their corresponding base models, arriving at a different conclusion: all medical VLMs and nearly all medical LLMs fail to consistently improve over their base models in the zero-/few-shot prompting and supervised fine-tuning regimes for medical question answering (QA).
For instance, on clinical-note-based QA tasks in the 3-shot setting, medical LLMs outperform their base models in only 26.7\% of cases, reach a (statistical) tie in 16.7\% of cases, and perform significantly worse in the remaining 56.7\% of cases. 
Our conclusions are based on (i) comparing each medical model directly against its base model; (ii) optimizing the prompts for each model separately in zero-/few-shot prompting; and (iii) accounting for statistical uncertainty in comparisons. 
Our findings suggest that state-of-the-art general-domain models may already exhibit strong medical knowledge and reasoning capabilities, and offer recommendations to strengthen the conclusions of future studies.
\end{abstract}


\section{Introduction}\label{sec:intro}
Recent advances in autoregressive large language models (LLMs)
and vision-language models (VLMs) have attracted interest
from practitioners in medicine, 
where these models hold great potential to transform
various aspects of clinical practice 
(e.g., medical diagnosis, information retrieval from clinical documents, patient triaging) 
\citep{how-ai-can-transform,generalist-medai}.
State-of-the-art performance on various medical benchmarks 
is typically achieved by massive-scale closed-source models,
such as \textsc{GPT-4} \citep{gpt4,gpt4v}, \textsc{Med-Gemini} \citep{med-gemini,med-gemini-2}, 
and \textsc{Med-PaLM} \citep{llm-clinical,med-palm-2,med-palm-m}, 
often performing on par with humans on medical licensing exams 
and open-ended consumer health question-answering (QA) tasks.
However, the general lack of transparency in these models, 
high API usage costs, and patient data privacy concerns 
make their integration into routine clinical workflows challenging \citep{llm-hipaa}. 

To address such concerns, recent works 
have proposed cheaper, open-source alternatives
through \textit{domain-adaptive pretraining}
\citep[DAPT;][]{dapt}, where a pretrained open-source general-domain model---such 
as \textsc{Llama} \citep{llama-1,llama-2,llama-3} or \textsc{Mistral} \citep{mistral} in the language space; 
and \textsc{LLaVA} \citep{llava} or \textsc{Open-Flamingo} \citep{open-flamingo} in the vision-language space---is continually pretrained on biomedical (image-)text corpora 
from public sources such as PubMed. 
While some works show that medical models pretrained from scratch 
only using domain-specific corpora can outperform those trained via DAPT, 
both in the context of BERT-style encoder-only models \citep{bert,pubmedbert,gatortron} 
and decoder models \citep{galactica,biogpt,do-we-still-need-clinical-lms,biomedlm}, 
the DAPT approach has become common practice, 
resulting in a trend where the release of a more capable general-domain model
is typically followed by the release of its medical counterpart.  

Despite the widespread adoption of medical DAPT, 
the claimed improvements in performance are worth scrutinizing. 
While the central story (improved domain-specific performance through domain-specific adaptation) is intuitive, more recent base models 
often already exhibit strong off-the-shelf performance 
on medical benchmarks without any adaptation.
For instance, as of the time of writing,
the general-domain \textsc{Llama-3-8B}~\citep{llama-3} outperforms 
other medically specialized models such as \textsc{MediTron-70B}~\citep{meditron} and \textsc{BioMistral-7B}~\citep{biomistral} on the Open Medical LLM Leaderboard~\citep{open-medical-llm-leaderboard}, which evaluates each model on standard medical QA benchmark datasets such as MedQA \citep{medqa} and MedMCQA \citep{medmcqa}.
Moreover, given the general lack of transparency about the pretraining corpora 
used to train the general-domain model in the first place, 
it is possible that they may already be trained on relevant medical text
(or even the very same text as used for medical DAPT).

Perhaps more concerning is the lack of apples-to-apples comparisons in the literature. 
First, medical models resulting from DAPT are often only compared 
against other baselines with different architectures and model scale
(e.g., \textsc{Clinical-Camel-70B} \citep{clinical-camel} vs.~\textsc{GPT-4} \citep{gpt4}).
Second, even for the same model scale, models are often evaluated under inconsistent evaluation setups.
For example, \citet{meditron} fine-tune \textsc{MediTron-70B} on MedMCQA with gradient updates
and compare it to~\textsc{Clinical-Camel-70B} \citep{clinical-camel} zero-shot prompted on MedMCQA. 
Third, the common practice of using a single, 
fixed prompting setup (e.g., prompt format, choice of few-shot examples) 
for all models under evaluation also warrants concern,
as LLM/VLM behavior is extremely sensitive to such design decisions 
\citep{how-can-we-know-what-lms-know,calibrate-before-use,open-clinical-llms-sensitive}, 
and prompt designs optimized for a proposed method are seldom optimal for the baselines \citep{quantifying-lm-prompt-design}.
These issues can undermine the validity 
of claimed performance benefits 
in the medical DAPT literature.

Meanwhile, medical LLMs trained via DAPT are often solely evaluated on QA tasks designed to assess medical knowledge (e.g., MedQA \citep{medqa}, MedMCQA \citep{medmcqa}), although it is generally unclear whether strong performance on such tasks necessarily implies high clinical utility \citep{shaky-foundations}. 
Notably, clinical notes (e.g., outpatient SOAP notes and discharge summaries) look different from academic biomedical articles (PubMed is the primary source of data used in many medical adaptation papers (Table \ref{tab:models})) both in form, content, and surface-level use of language (e.g., syntactical differences, use of jargon, abundance of naturally occurring grammatical errors \citep{llms-are-clinical-info,do-we-still-need-clinical-lms}).
To assess whether medical DAPT contributes to higher clinical utility, 
it is essential to also evaluate medical models resulting from DAPT 
on QA tasks grounded on real-world clinical notes
in direct comparison against their base models.

\begin{figure*}[t!]
    \centering
    \includegraphics[width=\linewidth]{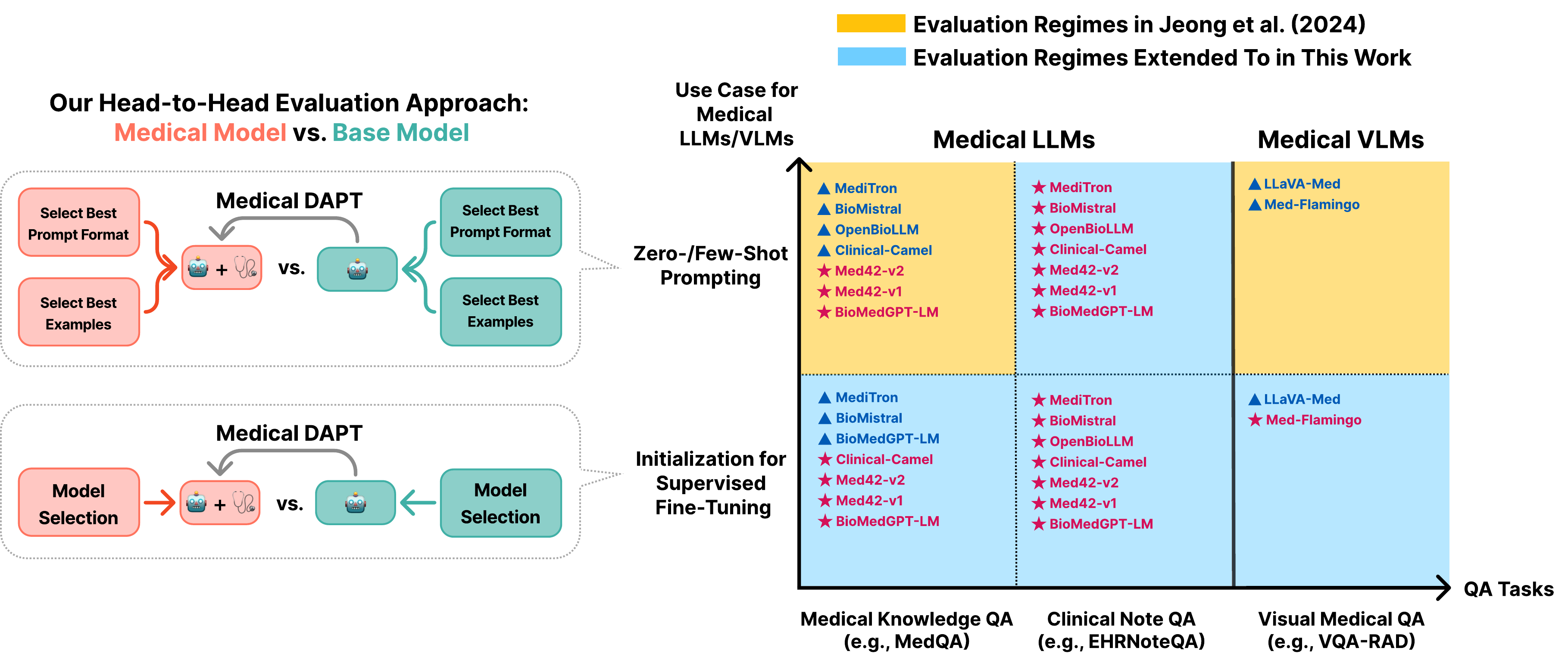}
    \caption{Overview of our head-to-head evaluation approach for each pair of medical LLM/VLM (\textcolor[HTML]{FF735E}{red}) and its base model (\textcolor[HTML]{40B0A6}{turquoise}), across a wide variety of medical QA tasks and model use cases. 
    (Left) We compare each model pair based on their zero-/few-shot QA capabilities (top) and their QA performances after supervised fine-tuning (bottom), \textit{while ensuring that their only differences lie in medical DAPT}.
    (Right) We substantially extend the results of \citet{eval-medical-dapt-emnlp} (\textcolor[HTML]{FFC20A}{yellow}) to include QA tasks based on clinical notes, as well as a comparison of downstream QA performance when using medical vs.~general-domain models as an initialization for fine-tuning (\textcolor[HTML]{22B4FF}{cyan}).
    Within each evaluation regime, a {\navytriangle} indicates that a model is evaluated in that regime both in the original paper and in our work, while a {\magentastar} indicates that it is \textit{only} evaluated in our work\protect\footnotemark. 
    \textit{None of the medical LLMs we evaluate have previously been evaluated on QA tasks based on real-world clinical notes.}
    }
    \label{fig:intro-fig-1}
\end{figure*}
\footnotetext{Although \citet{med42-v1,med42-v2} describe their evaluation of the proposed \textsc{Med42} models on medical knowledge QA tasks as being in the ``zero-shot'' setting, we treat the zero-/few-shot prompting evaluations as missing, 
since they were already fine-tuned on these datasets via medical DAPT.}

In this paper, we perform an apples-to-apples comparison that addresses these concerns, 
comparing ten medical LLMs and two medical VLMs against their general-domain base models on various medical (visual) QA tasks (Figure \ref{fig:intro-fig-1}). 
For each pair of general-domain and medically adapted LLMs/VLMs, 
\textbf{whose only differences lie in medical DAPT} (i.e., one model is the base model, from which the other is derived via medical DAPT), 
we compare their downstream performances from (i) zero-/few-shot prompting \citep{gpt-2,gpt-3} and (ii) supervised fine-tuning (SFT). 
For the former, we follow \citet{eval-medical-dapt-emnlp} and compare the performances after independently selecting the ``best'' prompt format and few-shot examples for each model based on the validation set (Section \ref{sec:prompting}). 
For the latter, we compare the performances after fine-tuning each model on the training set of each downstream QA dataset, using the best hyperparameters selected via grid search (Section \ref{sec:sft}). 
In both cases, we use the percentile bootstrap to assess whether the perceived improvements in performance from medical DAPT are attributable to chance.
In Table \ref{tab:models}, we list all of the LLM/VLM pairs included in our study, which were selected to cover a wide range of general-domain base models and model scales (7B--70B). 

\begin{figure*}[t!]
    \centering
    \includegraphics[width=\linewidth]{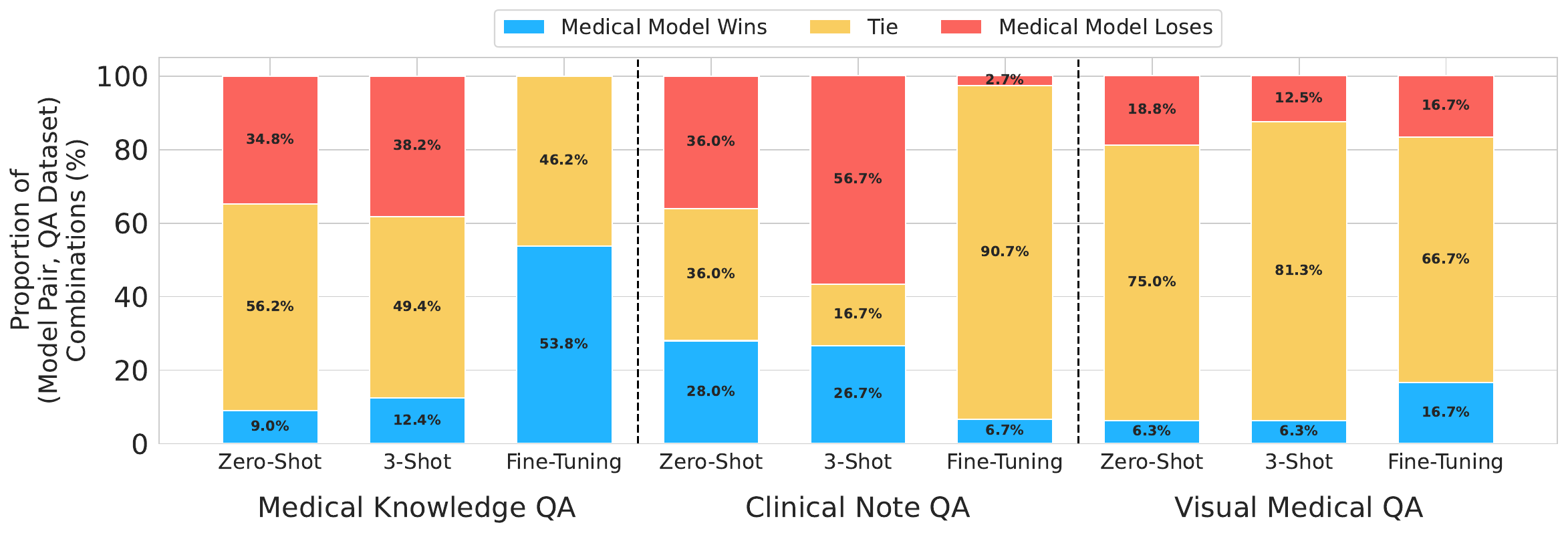}
    \vspace{-15pt}
    \caption{Medical LLMs and VLMs trained via DAPT show limited improvement over their general-domain counterparts in the zero-/few-shot prompting and supervised fine-tuning (SFT) regimes.
    Here, we show the win, tie, and loss rates (\%) of medical models vs.~their corresponding base models across all (model pair, QA dataset) combinations.
    Win rate refers to the proportion of (model pair, QA dataset) combinations where a medical model shows a \textit{statistically significant} improvement.
    We do observe that on textual medical knowledge QA tasks, medical LLMs outperform their base models in the SFT regime, although the improvements are marginal (1.6\% average gain in accuracy; Section \ref{sec:sft-results-knowledge}).
    }
    \label{fig:intro-fig-2}
\end{figure*}

\begin{table*}[t!]
    \centering
    \caption{
    Prior works overlook aspects critical to assessing the impact of medical DAPT on downstream zero-/few-shot performance, potentially leading to overly optimistic conclusions about its benefits. 
    For each medical LLM/VLM included in our study (Table \ref{tab:models}), we summarize the following details:
    (i) whether the medical model was compared against its base model; 
    (ii) whether the sensitivity of the medical model and its base model to the choice of prompt format and few-shot examples was considered in their comparison; 
    and (iii) whether statistical tests were performed to determine the significance of any measured performance improvement.
    {\tri} indicates that the corresponding aspect was only partially considered (e.g., accounted for sensitivity to few-shot examples but not the prompt format).
    }
    \label{tab:eval-comparison}
    \resizebox{0.9\linewidth}{!}{
    \begin{tabular}{@{}l@{\hskip 7pt}c@{\hskip 7pt}c@{\hskip 7pt}c@{\hskip 7pt}c@{}}
        \toprule
        & \shortstack{Compared to\\Base Model?} & \shortstack{Accounted for\\Prompt Sensitivity?} & \shortstack{Tested for\\Statistical Significance?} \\
        \midrule
        \textsc{Med42-v2} \citep{med42-v2} & \cmark & & \\
        \textsc{Med42-v1} \citep{med42-v1} & \cmark & & \\
        \textsc{OpenBioLLM} \citep{OpenBioLLMs} & & & \\
        \textsc{Clinical-Camel} \citep{clinical-camel} & & & \\
        \textsc{BioMistral} \citep{biomistral} & \cmark & \tri & \\
        \textsc{MediTron} \citep{meditron} & \cmark & \tri & \\
        \textsc{BioMedGPT-LM} \citep{biomedgpt} & & & \\
        \midrule
        \textsc{LLaVA-Med} \citep{llava-med} & \cmark & & \\
        \textsc{Med-Flamingo} \citep{med-flamingo} & \cmark & & \\
        \midrule
        \textbf{Our Evaluation} & \cmark & \cmark & \cmark \\
        \bottomrule
    \end{tabular}
    }
\end{table*}

\textbf{Overall, we find that both medical LLMs and VLMs show limited improvement over their base models, across all of the medical QA tasks and evaluation settings that we consider (Figure \ref{fig:intro-fig-2}).}
In the zero-/few-shot prompting regime (Section \ref{sec:results}), we find that all medical VLMs and the majority of medical LLMs fail to consistently outperform their base models across all datasets, including QA tasks focused on assessing medical knowledge and those based on real-world clinical notes. 
In the SFT regime (Section \ref{sec:sft-results}), we find that the medical LLMs overall \textit{do} show statistically significant improvements on the medical knowledge QA tasks but not on the clinical note QA tasks, while the medical VLMs show little to no improvement on all of the visual medical QA tasks. 
Our findings suggest that a rigorous pairwise comparison of models, including tests for statistical significance, is essential to draw reliable conclusions about the performance benefits from medical DAPT, while these basic practices are not consistently adopted in the literature (Table \ref{tab:eval-comparison}).

Our main contributions can be summarized as follows:
\begin{enumerate}[topsep=0.5ex,itemsep=-0.5ex]
    \item We provide a comprehensive head-to-head comparison between state-of-the-art general-domain LLMs/VLMs and their medical DAPT counterparts on a wide range of medical QA tasks, to investigate the effectiveness of DAPT for medical specialization.
    \item We find that when the prompts are optimized for each medical and general-domain model independently, the majority of medical models fail to improve over their general-domain counterparts in the zero-/few-shot prompting regime (Section \ref{sec:prompting-results-1}).
    \item We show that using a single, fixed prompt format and choice of few-shot examples for all models without statistical testing can lead to overly optimistic conclusions about the benefits from medical DAPT in the zero-/few-shot prompting regime (Section \ref{sec:prompting-results-2}).
    \item We find that in the SFT regime, all medical VLMs fail to show improvement, while medical LLMs show improvement on textual medical knowledge QA tasks but not on QA tasks based on real-world clinical notes (Section \ref{sec:sft-results}).
\end{enumerate}

\section{Related Work}\label{sec:related-work}
DAPT \citep{dapt} is a transfer learning approach, 
where a pretrained model is further pretrained on domain-specific data
for better alignment to a target domain of interest (e.g., medicine, law). 
Several studies show that language models trained via DAPT 
often outperform their general-domain counterparts on domain-specific tasks,
such as claim detection from blog posts \citep{chakrabarty-etal-2019-imho}, 
named entity recognition from German novels \citep{german-novel}, 
and judgment prediction for legal cases \citep{lawformer}.
In the medical domain, prior works based on BERT-style encoder models \citep{bert}, 
such as \textsc{BioBERT} \citep{biobert} and \textsc{ClinicalBERT} \citep{clinical-bert-alsentzer}, 
show that medical DAPT improves fine-tuning performance on tasks 
such as medical concept extraction from patient reports \citep{i2b2-2010}, 
identification of gene-disease relations from PubMed abstracts \citep{ncbi-disease,gad,chemprot},
and natural language inference on clinical notes \citep{mednli}. 

More recent works suggest that decoder-based autoregressive LLMs and VLMs trained via medical DAPT 
also show strong performance on various downstream medical tasks. 
Medical LLMs such as \textsc{MediTron} \citep{meditron}, adapted from \textsc{Llama-2} \citep{llama-2};
and \textsc{BioMistral} \citep{biomistral}, adapted from \textsc{Mistral-7B-Instruct-v0.1} \citep{mistral}; 
perform well on knowledge-intensive QA tasks derived from medical licensing 
and academic exams \citep{medqa,medmcqa,mmlu} and PubMed abstracts \citep{pubmedqa}. 
Medical VLMs such as \textsc{LLaVA-Med} \citep{llava-med}, 
adapted from \textsc{LLaVA} \citep{llava}; 
and \textsc{Med-Flamingo} \citep{med-flamingo}, adapted from \textsc{Open-Flamingo} \citep{open-flamingo}; also perform well on visual medical QA tasks based on radiology \citep{vqa-rad,slake} 
and pathology images \citep{pvqa} 
and academic exams \citep{mmmu}.
These encouraging results have established DAPT as a go-to approach 
for training medically specialized LLMs and VLMs,
a conclusion that we re-examine across different evaluation settings (zero-/few-shot prompting, SFT) and across both knowledge-intensive and clinically relevant QA tasks.

\section{Experimental Setup}\label{sec:eval-setup}
To investigate the effectiveness of medical DAPT in improving (i) zero-/few-shot and (ii) SFT performances,
we compare ten medical LLMs and two medical VLMs against their general-domain counterparts in \textit{pairs} on 22 textual medical QA datasets (8 based on clinical notes) and 8 visual medical QA datasets, respectively (Figure \ref{fig:intro-fig-1}). 
The models in each pair are \textit{exactly identical} in model architecture and scale, and their only difference lies in whether they were additionally pretrained on medical data. 
We also note that while some of the datasets considered contain both closed-ended questions (with objective ground-truth answers) and open-ended questions, we focus our evaluations on the former, where an objective, quantitative assessment of medical knowledge and reasoning capabilities is possible.
To ensure the reproducibility of our results, we open-source the source code used for all of our evaluations described below via our GitHub repository\footnote{\href{https://github.com/taekb/eval-medical-dapt}{github.com/taekb/eval-medical-dapt}}.

\begin{table*}[t!]
    \centering
    \caption{Summary of open-source autoregressive VLM and LLM pairs used for evaluation. For \textsc{Med42}, we only list the top-five adaptation datasets with the highest mixture ratios.}
    \label{tab:models}
    \resizebox{\linewidth}{!}{
    \begin{tabular}{@{}c@{\hskip 10pt}l@{\hskip 7pt}l@{\hskip 7pt}l@{\hskip 3pt}}
        \toprule
        \textbf{Model Class} & \textbf{General Domain} & \textbf{Medical Domain} & \textbf{Medical Adaptation Corpora} \\
        \midrule
        \multirow{30}{*}{\textbf{LLM}} & \multirow{5}{*}{\textsc{Llama-3-70B-Instruct}} & \multirow{5}{*}{\textsc{Med42-v2-70B}} & Medical QA Datasets (e.g., MedQA, MedMCQA) \\
        & & & Medical Instruction 120k \citep{med-instruction-120k} \\
        & & & OpenGPT (OpenChat) \citep{openchat} \\
        & & & StackExchange \citep{stack-exchange} \\
        & & & Medical Flashcards \citep{medalpaca} \\
        \cmidrule(){2-4}
        & \textsc{Llama-3-70B-Instruct} & \textsc{OpenBioLLM-70B} & Undisclosed \\
        \cmidrule(){2-4}
        & \multirow{2}{*}{\textsc{Llama-2-70B}} & \multirow{2}{*}{\textsc{MediTron-70B}} & Clinical Practice Guidelines (e.g., CDC, WHO) \\
        & & & PubMed Articles \citep[S2ORC;][]{s2orc} \\
        \cmidrule(){2-4}
        & \multirow{3}{*}{\textsc{Llama-2-70B}} & \multirow{3}{*}{\textsc{Clinical-Camel-70B}} & ShareGPT \\
        & & & 20k PubMed Articles Published Before 2021 \\
        & & & Random 4k Subset of MedQA \citep{medqa} \\
        \cmidrule(){2-4}
        & \multirow{5}{*}{\textsc{Llama-2-70B}} & \multirow{5}{*}{\textsc{Med42-v1-70B}} & Medical QA Datasets (e.g., MedQA, MedMCQA) \\
        & & & OpenGPT (OpenChat) \citep{openchat} \\
        & & & StackExchange \citep{stack-exchange} \\
        & & & Medical Flashcards \citep{medalpaca} \\
        & & & CORD-19 \citep{cord-19} \\
        \cmidrule(){2-4}
        & \multirow{5}{*}{\textsc{Llama-3-8B-Instruct}} & \multirow{5}{*}{\textsc{Med42-v2-8B}} & Medical QA Datasets (e.g., MedQA, MedMCQA) \\
        & & & Medical Instruction 120k \citep{med-instruction-120k} \\
        & & & OpenGPT (OpenChat) \citep{openchat} \\
        & & & StackExchange \citep{stack-exchange} \\
        & & & Medical Flashcards \citep{medalpaca} \\
        \cmidrule(){2-4}
        & \textsc{Llama-3-8B} & \textsc{OpenBioLLM-8B} & Undisclosed \\
        \cmidrule(){2-4}
        & \multirow{2}{*}{\textsc{Llama-2-7B}} & \multirow{2}{*}{\textsc{MediTron-7B}} & Clinical Practice Guidelines (e.g., CDC, WHO) \\
        & & & PubMed Articles \citep[S2ORC;][]{s2orc} \\
        \cmidrule(){2-4}
        & \textsc{Mistral-7B-Instruct-v0.1} & \textsc{BioMistral-7B} & PubMed Articles (PMC Open Access Subset) \\
        \cmidrule(){2-4}
        & \textsc{Llama-2-7B-Chat} & \textsc{BioMedGPT-LM-7B} & PubMed Articles \citep[S2ORC;][]{s2orc} \\
        \midrule
        \multirow{3}{*}{\textbf{VLM}} & \textsc{LLaVA-v0-7B} & \textsc{LLaVA-Med-7B} & PubMed Articles \citep[PMC-15M;][]{biomedclip} \\
        \cmidrule(){2-4}
        & \multirow{2}{*}{\textsc{Open-Flamingo-9B}} & \multirow{2}{*}{\textsc{Med-Flamingo-9B}} & Medical Textbooks \citep[MTB;][]{med-flamingo} \\
        & & & PubMed Articles \citep[PMC-OA;][]{pmcclip}\\
        \bottomrule
    \end{tabular}
    }
\end{table*}

\paragraph{Models (Table \ref{tab:models}).} 
For medical LLM evaluations, we consider the following model families: \textsc{Med42-v2} \citep{med42-v2}, \textsc{Med42-v1} \citep{med42-v1}, \textsc{OpenBioLLM} \citep{OpenBioLLMs}, \textsc{MediTron} \citep{meditron}, \textsc{Clinical-Camel} \citep{clinical-camel}, \textsc{BioMistral} \citep{biomistral}, and \textsc{BioMedGPT-LM} \citep{biomedgpt}. 
For medical VLM evaluations, we consider \textsc{LLaVA-Med} \citep{llava-med} and \textsc{Med-Flamingo} \citep{med-flamingo}. 
For all models, we use the checkpoints made available via HuggingFace. 
In all zero-/few-shot prompting experiments, we generate predictions from each model via (i) greedy decoding (i.e., sampling with temperature $T=0$) and (ii) constrained decoding. For constrained decoding, we constrain the token vocabulary to be one of the answer choice letters (e.g., one of [``A'', ``B'', ``C'', ``D''] for a four-choice QA dataset) and treat the answer choice with the highest token probability as a given model's prediction. For the SFT experiments, we only generate predictions via greedy decoding, as it best reflects the setup used for fine-tuning (Section \ref{sec:sft}).

\paragraph{Textual Medical Knowledge QA Datasets.} For textual medical knowledge QA, we use MedQA \citep{medqa}, MedMCQA \citep{medmcqa}, PubMedQA \citep{pubmedqa}, and MMLU-Medical \citep{mmlu} for evaluation. MMLU-Medical refers to a subset of MMLU corresponding to 9 subjects related to medicine: anatomy, clinical knowledge, college biology, college medicine, high school biology, medical genetics, nutrition, professional medicine, and virology. For MedQA, we use the official train--validation--test splits as provided through BigBio \citep{bigbio}. 
We note that MedQA has two versions, one with four answer choices per question and the other with five, and we use both for evaluation.
For MedMCQA, which does not have a public test set, we follow the approach taken by \citet{pmc-llama} and \citet{biomistral}, taking a random 80--20 train--validation split of the official training set and using the official validation set for testing. For PubMedQA, we follow \citet{llm-clinical}, using the 211k artificially generated QA samples for training, and taking a 50--50 split on the 1k expert-labeled examples. For MMLU-Medical, we use the official split as provided. 
Meanwhile, as the \textsc{Med42} models---\textsc{Med42-v2-70B}, \textsc{Med42-v2-8B}, and \textsc{Med42-v1-70B}---were trained on all of the QA datasets mentioned above as part of DAPT, \textit{we exclude these models from our discussion on textual medical knowledge QA}.

\paragraph{Textual Clinical Note QA Datasets.} 
As strong performance on medical knowledge QA tasks may not necessarily imply high clinical utility \citep{shaky-foundations}, we also perform evaluations with the following textual QA datasets based on clinical notes: 
MedNLI \citep{mednli}, EHRNoteQA \citep{ehrnoteqa}, the 2008 i2b2 Obesity Comorbidity Detection Challenge dataset \citep{i2b2-2008} (4 classification tasks), the CASI Clinical Acronym Sense Disambiguation dataset \citep{casi}, and the MIMIC-III Clinical Acronym Sense Disambiguation dataset \citep{mimic-iii,lmc}. For MedNLI, we frame each natural language inference task as a three-way closed-ended QA task, where the answer to each question is one of [``entailment'', ``contradiction'', and ``neutral''], and use the official train--validation--test splits provided via PhysioNet~\citep{physionet} under a data use agreement. 
For EHRNoteQA, we use the ``Level 1'' subset, which only contains QA pairs with clinical notes at most 3k tokens in length, to account for the limited context window sizes of the LLMs that we evaluate, and take a random 60--20--20 split of the data to obtain the train, validation, and test sets.
For the 2008 i2b2 Obesity Comorbidity Detection Challenge dataset, we follow the preprocessing steps in \citet{open-clinical-llms-sensitive} to obtain 4 binary classification (i.e., two-way multiple-choice QA) datasets, each focused on detecting the presence of (i) asthma, (ii) coronary artery disease (CAD), (iii) diabetes, and (iv) obesity from a given clinical note.
We then select the QA pairs with clinical notes at most 3k tokens in length, as done for EHRNoteQA.
As the preprocessing pipeline by \citet{open-clinical-llms-sensitive} outputs a train--test split, we take a random 80--20 split on the training set to obtain the train and validation sets.
For the CASI and MIMIC-III sense disambiguation datasets, we follow the preprocessing steps detailed in \citet{lmc} for consistency with prior works \citep{llms-are-clinical-info}.

\paragraph{Visual Medical QA Datasets.} For visual medical QA, we use VQA-RAD \citep{vqa-rad}, PathVQA \citep{pvqa}, SLAKE \citep{slake}, and MMMU-Medical \citep{mmmu} for evaluation. MMMU-Medical is a subset of MMMU with 5 subjects relevant to medicine: basic medical science, clinical medicine, diagnostics and lab medicine, pharmacy, and public health. For VQA-RAD, we address the train--test leakage and duplication issues in the official train--test split \citep{med-flamingo} by removing the training examples repeated in the test set and removing all duplicates in both sets. 
As the official split does not include a validation set, we take a random 80--20 split on the training set to create a new train--validation split. For MMMU-Medical, which lacks a public test set, we randomly select 5 examples from the official validation set for validation, and reserve the remaining 25 examples for testing. For all other datasets, we use the official split as provided. 

\paragraph{Evaluation Metric.} Since we focus on closed-ended QA tasks, we use exact-match accuracy as our main evaluation metric. Following the Holistic Evaluation of Language Models (HELM) benchmark \citep{helm}, we treat the text generated by a model (without any constraints on the vocabulary) to be its prediction, and check for an exact match between the prediction and the correct answer up to primitive string operations (e.g., lower-casing, removing white space/punctuation). If multiple matches occur, we treat the prediction as incorrect (even if one of the matches is the correct answer), in order to handle cases where the model simply repeats the list of answer choices or produces an ambiguous answer (e.g., selecting multiple answer choices).
Meanwhile, to quantify the extent of \textit{improvement} from medical DAPT, we also consider the \textit{relative} accuracy of the medical model with respect to the general-domain model. Formally, we define relative exact-match accuracy as $\mathbb{E}[\mathds{1}[f_{\text{medical}}(x) = y] - \mathds{1}[f_{\text{general}}(x) = y]] \in [-1,1]$, where $f_{\text{medical}}$ and $f_{\text{general}}$ denote the medical and general-domain models, $x$ and $y$ denote the input prompt and answer in a QA pair from the test set, and $\mathds{1}[\cdot]$ denotes the indicator function. 
This metric quantifies the difference in accuracy between the medical model and the general-domain model.
We refer to the accuracy of a particular model as the \textit{absolute} exact-match accuracy in subsequent discussions, and refer to the metric above (for a pair of models) as the \textit{relative} accuracy.

\paragraph{Assessing Statistical Significance.} Given the relatively small size of test datasets in medical QA benchmarks, it is important to assess whether the perceived improvements in performance from medical DAPT are attributable to chance.  To account for statistical uncertainty, we use the percentile bootstrap, re-sampling (with replacement) questions from the test set to get a sample of the same size as the original test set.  Within each resample, we compute the difference in accuracy for the paired models, and repeat this process for 10,000 iterations.  The resulting distribution of relative accuracy is used to derive a 95\% confidence interval, and we judge a difference to be statistically significant if this interval does not cross zero.  We do not perform any type of multiple-testing correction, which would have the effect of lowering the number of comparisons deemed to be significant.

\subsection{Zero-/Few-shot Prompting with Model-Specific Prompt Selection}\label{sec:prompting}

In this section, we provide an overview of our approach to assess whether medical DAPT leads to statistically significant improvements in zero-/few-shot medical QA performance. 
For few-shot prompting, we consider the 3-shot setting to ensure that the input prompt is shorter than the context window sizes for all models evaluated.
Meanwhile, we exclude the EHRNoteQA and i2b2 datasets from few-shot prompting evaluations for medical LLMs, as most of the clinical notes in these datasets already occupy the full context window for most models even in the zero-shot setting.
For evaluation, we pay special attention to two aspects. 
First, language models are highly sensitive to the choice of prompting strategy (e.g., prompt format, choice of few-shot examples), where seemingly insignificant changes to the prompt can lead to idiosyncratic model behavior \citep{how-can-we-know-what-lms-know,calibrate-before-use,open-clinical-llms-sensitive}. 
Second, prior works show that the ``optimal'' choice of prompt format is rarely the same between different models \citep{quantifying-lm-prompt-design}, suggesting that using a single, fixed prompt for all models for comparison can result in misleading conclusions.

\begin{figure*}[t!]
    \centering
    \includegraphics[width=1\linewidth]{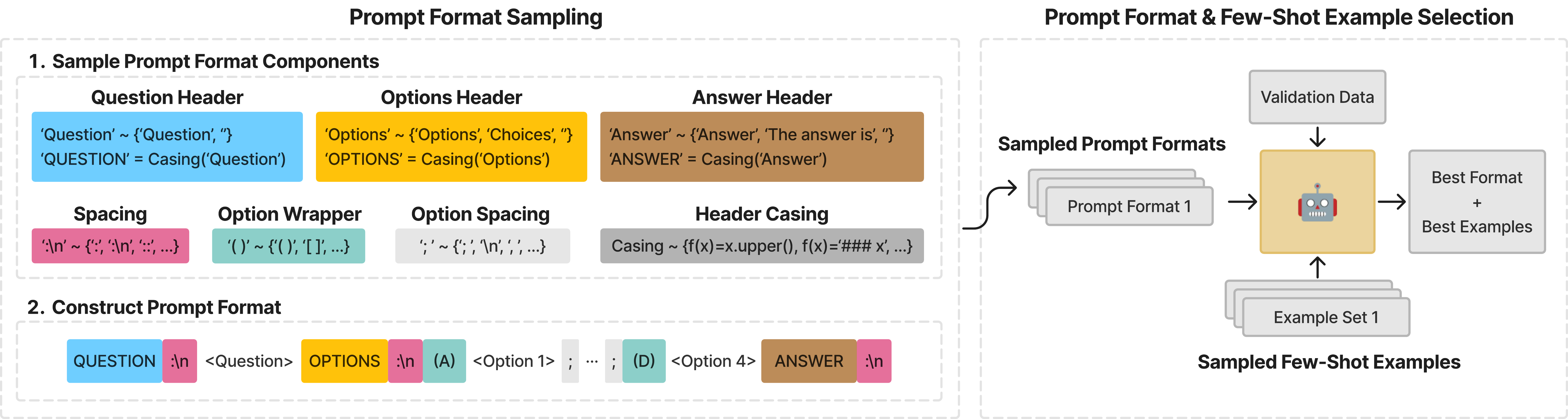}
    \caption{Overview of the prompt format sampling (left) and model-specific prompt selection (right) process considered in our evaluations (Section \ref{sec:prompting}).}
    \label{fig:prompt-selection}
\end{figure*}

To ensure a fair comparison that isolates the impact of medical DAPT, we treat the choice of prompt format and few-shot examples as additional hyperparameters when generating predictions, and tailor them to each model \textit{independently} (Figure \ref{fig:prompt-selection}). 
We first randomly sample 10 plausible prompt formats from a predefined search space and 10 different sets of few-shot examples from the training set of each dataset. We then search over all pairs of prompt formats (plus one additional manually designed default format) and few-shot examples, and select the best pair out of $(10+1) \times 10 = 110$ that results in the highest validation exact-match accuracy. 
Given that a grid search at this scale can be computationally expensive, especially for datasets like MedMCQA that contain 37k validation QA pairs (Table \ref{tab:datasets}), we randomly subsample 500 validation QA pairs for datasets that have more than 500. Using the vLLM framework \citep{vllm} for sampling model outputs, this leads to a runtime of around 5--15 minutes per trial, when using 4 NVIDIA A6000 GPUs for the 70B models and 2 GPUs for the others.
We then generate the final predictions on the test set using the prompt format and few-shot samples selected for each model. 
In the zero-shot setting, we only search over the prompt formats. 
Meanwhile, we acknowledge that in the ``truly'' zero-shot setting, it may not be possible to optimize the prompt with respect to a dedicated validation set. Nonetheless, we implement our prompt selection approach even in the zero-shot setting to mitigate the potential impact that choosing a particular prompt format has on QA performance, and ensure a fair comparison between models in each pair.

To define the prompt format search space, we follow \citet{quantifying-lm-prompt-design} and construct a context-free grammar of semantically equivalent yet syntactically distinct prompt formats (Figure~\ref{fig:prompt-selection}, left).
For the medical models that have a specific prompt format designed and recommended for closed-ended QA tasks (e.g., \textsc{BioMistral} \citep{biomistral}), we fix the prompt format to what is provided and only search over the choice of few-shot examples. When such information is missing or only partially available (Table~\ref{tab:prompting-details}), we search over both the prompt formats and few-shot examples. For instruction-tuned models, which typically have a structured conversational format that is expected (e.g., ``\texttt{\#\#\# User:\ldots \#\#\# Assistant:\ldots}''), we use the sampled question and answer templates to format each ``user'' query and ``assistant'' response. We provide the remaining details in Appendices \ref{sec:details-prompting-selection}--\ref{sec:details-prompting}.

\subsection{Supervised Fine-Tuning (SFT)}\label{sec:sft}
To assess whether medical DAPT leads to a better initialization of the model parameters for fine-tuning on downstream medical QA tasks, we compare the performances of models in each pair after fine-tuning each on the training set of each QA dataset. 
Due to computational constraints, we focus on parameter-efficient fine-tuning methods (detailed below) instead of full fine-tuning. 
For the textual medical QA evaluations, we exclude the MMLU-Medical datasets given that the training sets only include 5 examples per subject.
For the visual medical QA evaluations, we exclude the MMMU-Medical datasets for the same reason.

\paragraph{Parameter-efficient fine-tuning.} For \textsc{Med-Flamingo-9B} and \textsc{Open-Flamingo-9B}, we fine-tune the gated cross-attention layers, the Perceiver resampler \citep{perceiver}, and the embeddings of the special \texttt{<image>} and \texttt{<|endofchunk|>} tokens, 
which correspond to a only subset of all model parameters.
For all other LLMs and VLMs, we fine-tune each model with low-rank adapters \citep[LoRA;][]{lora}. 
We add the low-rank adapters to all of the linear layers (i.e., the self-attention and feedforward network layers in each Transformer block), following prior works that demonstrate its effectiveness over adapting a subset of the linear layers \citep{qlora,lora-learns-less-forgets-less}.
We fine-tune the trainable parameters by minimizing the cross-entropy loss on the output tokens generated conditional on the input context (i.e., the task instruction, image, and question). 
We train all models for a maximum of 10 epochs and apply early stopping regularization, monitoring the validation loss after every epoch of training and enforcing a patience of 1.
To apply the gradient updates, we use the AdamW optimizer \citep{adamw} with the recommended momentum hyperparameters $\beta_1=0.9,\beta_2=0.999$, and $\epsilon=10^{-8}$ \citep{adam}; and a cosine learning rate schedule with a linear warmup of 0.05 $\times$ maximum number of training steps.
For \textsc{Med-Flamingo-9B} and \textsc{Open-Flamingo-9B}, we perform a grid search over the learning rates and weight decay coefficients and use the validation loss to select the best setting for final evaluation.
For all other models, we perform a grid search over the LoRA ranks and learning rates and use the validation loss to select the best setting. 
We perform all experiments using the DeepSpeed ZeRO \citep{deepspeed-zero} and PyTorch FSDP \citep{pt-fsdp} frameworks for distributed training. 
Meanwhile, for fine-tuning the 70B-parameter LLMs on PubMedQA and i2b2 (Diabetes), we also apply 4-bit NormalFloat double quantization \citep[QLoRA;][]{qlora} to the model parameters to address out-of-memory issues.
We provide the remaining details in Appendix \ref{sec:details-sft}.

\section{Zero-/Few-Shot Prompting Evaluation Results}\label{sec:results}
We summarize the main findings from the zero-/few-shot prompting 
experiments outlined in Section \ref{sec:prompting}. 
Unless specified otherwise, we focus on the greedy decoding results in subsequent discussions and include the results for constrained decoding in Appendix \ref{sec:constrained-decoding}.
\textbf{Overall, we find that all medical VLMs and the majority of medical LLMs fail to consistently improve over their general-domain counterparts in the zero-/few-shot prompting regime (Section \ref{sec:prompting-results-1}).}
We find that the medical LLMs do not show consistent improvements on \textit{either} medical knowledge QA or clinical note QA datasets in this setting. 
Moreover, we demonstrate the importance of rigorous experimental design in surfacing this finding---performing pairwise model comparison with a single, fixed prompt optimized only for the medical model, while ignoring statistical uncertainty, can paint a misleadingly optimistic picture of off-the-shelf performance benefits from medical DAPT (Section \ref{sec:prompting-results-2}).

\subsection{Performance Benefits from Medical DAPT Largely Diminish After Model-Specific Prompt Selection and Statistical Testing}
\label{sec:prompting-results-1}
We first show that all medical VLMs and the majority of medical LLMs fail to consistently outperform their corresponding base models in terms of zero-shot and 3-shot performance, after model-specific prompt selection and statistical testing. 
We calculate the 95\% confidence intervals in relative exact-match accuracy via bootstrapping on the test set, as described in Section \ref{sec:eval-setup}. Below, we present the results on the textual medical knowledge QA datasets, the textual clinical note QA datasets, and the visual medical QA datasets one-by-one.

\begin{figure*}[t!]
    \centering
    \begin{tabular}{@{}c@{}c@{}}
        \multicolumn{2}{c}{
            \begin{subfigure}{0.95\linewidth}
                \includegraphics[width=\linewidth]{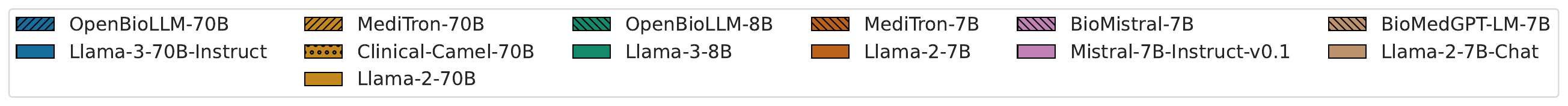}
            \end{subfigure}
        }
        \\
        \begin{subfigure}{0.02\linewidth}
            \makebox[\linewidth]{\raisebox{75pt}{{(a)}}}
        \end{subfigure} &
        \begin{subfigure}{0.95\linewidth}
            \includegraphics[width=\linewidth]{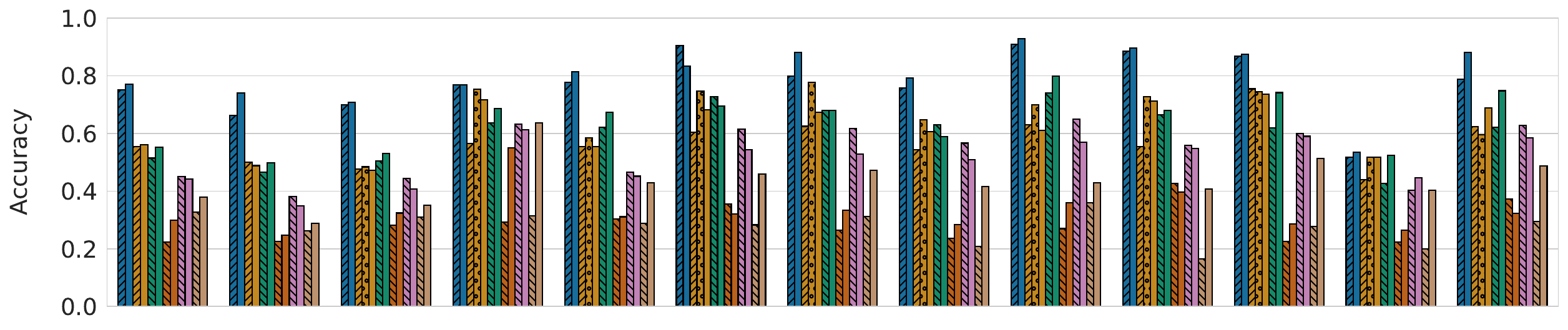}
        \end{subfigure}
        \\
        \begin{subfigure}{0.02\linewidth}
        \end{subfigure} &
        \begin{subfigure}{0.95\linewidth}
            \includegraphics[width=\linewidth]{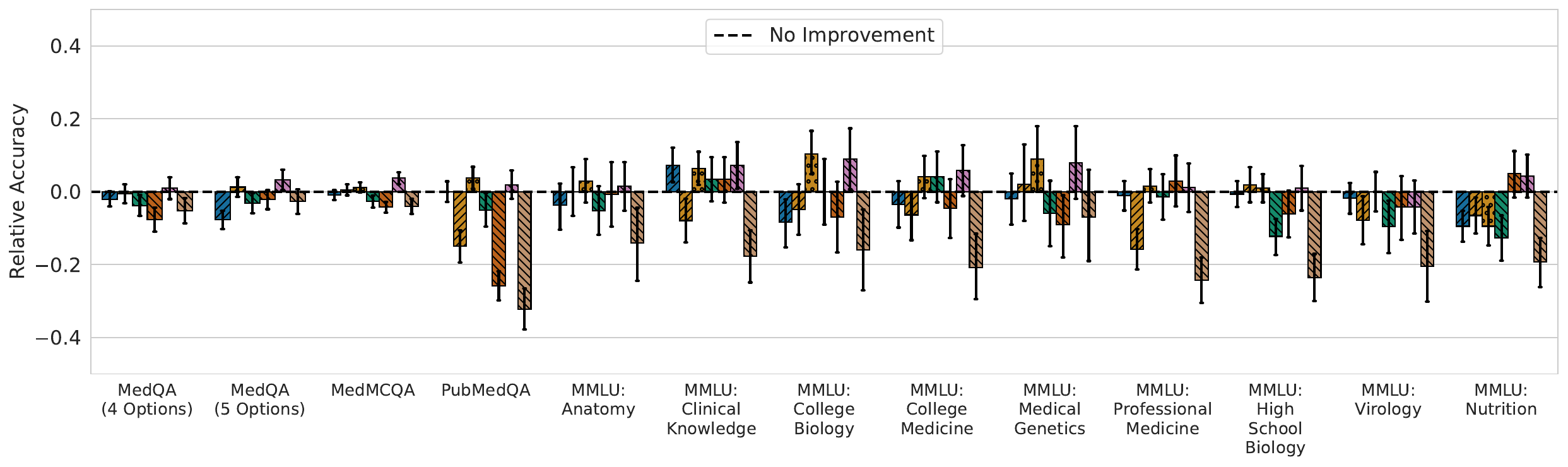}
        \end{subfigure}
        \\
        \begin{subfigure}{0.02\linewidth}
            \makebox[\linewidth]{\raisebox{75pt}{{(b)}}}
        \end{subfigure} &
        \begin{subfigure}{0.95\linewidth}
            \includegraphics[width=\linewidth]{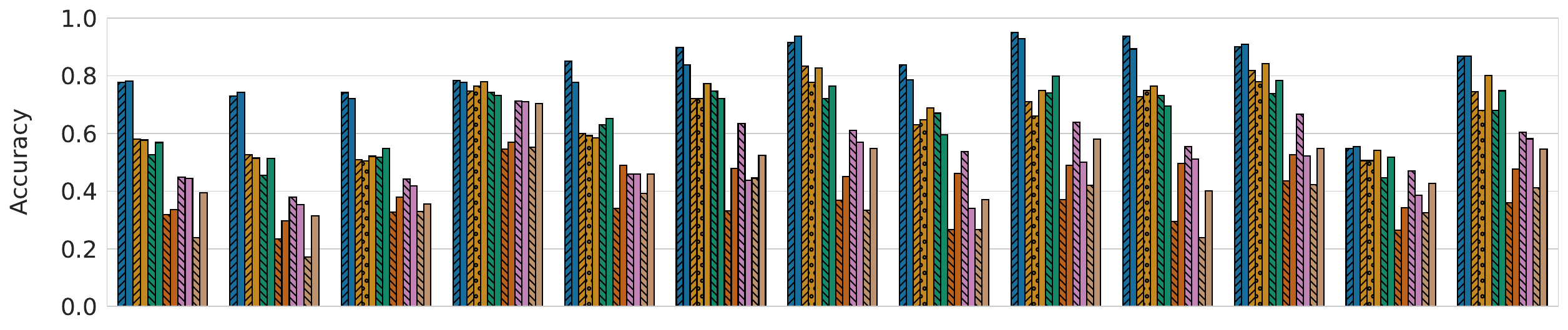}
        \end{subfigure}
        \\
        \begin{subfigure}{0.02\linewidth}
        \end{subfigure} &
        \begin{subfigure}{0.95\linewidth}
            \includegraphics[width=\linewidth]{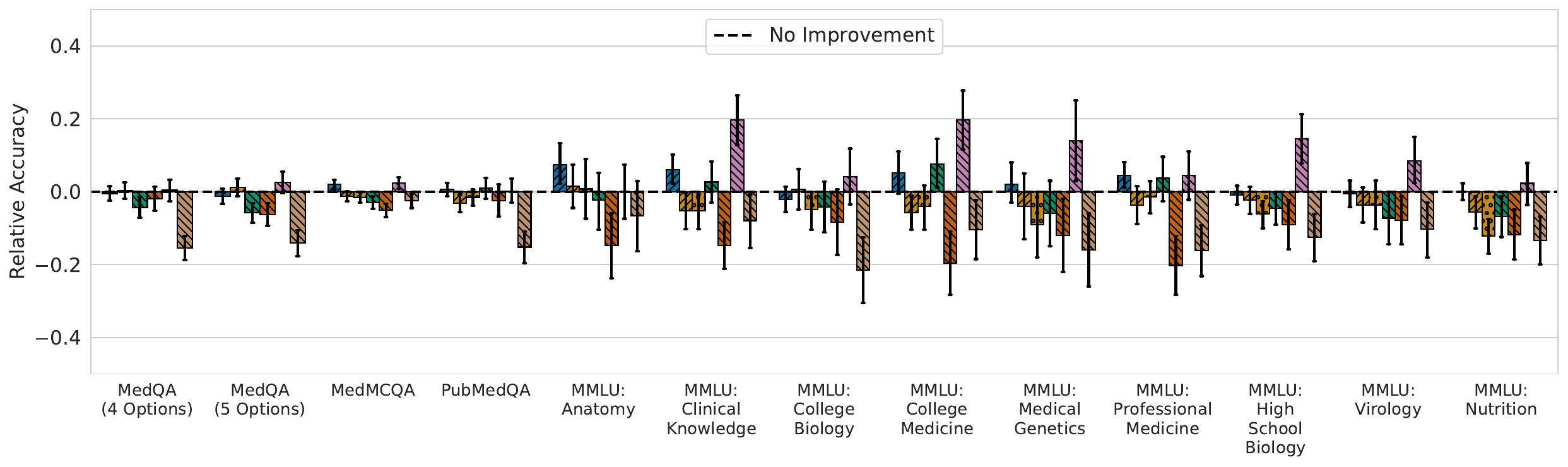}
        \end{subfigure}
    \end{tabular}
    \vspace{-10pt}
    \caption{After independently selecting the best prompt format and examples for each model, medical LLMs (textured bars) fail to consistently improve over their base models (solid bars) on textual medical knowledge QA tasks, in both (a) zero-shot and (b) 3-shot settings. 
    In each panel, the top row shows the absolute exact-match accuracies on the test set, and the bottom row shows the relative exact-match accuracies along with 95\% confidence intervals derived via bootstrapping on the test set (Section~\ref{sec:eval-setup}). 
    Here, model predictions are generated via greedy decoding. 
    Improvements are also limited with constrained decoding (Figure \ref{fig:llm-acc-ci-logprob-knowledge}).}
    \label{fig:llm-acc-ci-knowledge}
\end{figure*}

\subsubsection{Evaluation of Medical LLMs on Textual Medical Knowledge QA}
\label{sec:prompting-results-1-knowledge}
In Figures \ref{fig:llm-acc-ci-knowledge}(a) and (b), we show the absolute and relative exact-match accuracies achieved by the medical and general-domain LLMs on the textual medical knowledge QA datasets, from zero-shot and 3-shot prompting, respectively.
In Table \ref{tab:win-tie-loss-rates-knowledge}, we show the win, tie, and loss rates (\%) of the medical LLMs, where win rate refers to the proportion of QA datasets where a medical model shows a statistically significant improvement over its base model.
We exclude the results for \textsc{Clinical-Camel-70B} on both versions of MedQA, and the \textsc{Med42} models on all textual medical knowledge QA datasets, as they have already been trained on these datasets as part of medical DAPT (Table \ref{tab:models}). 

Overall, we find that the majority of medical models show little to no improvement over their base models, with the 95\% confidence intervals in relative accuracy crossing or lying below zero (bottom rows of Figure \ref{fig:llm-acc-ci-knowledge}(a) and (b)).
\begin{table}[t!]
    \centering
    \caption{The zero-shot and 3-shot win/tie/loss rates (\%) for all medical LLMs on textual medical knowledge QA, after independently optimizing the prompt for each model. For each medical model, we boldface the win rate if it wins more than it loses to its general-domain base model, and vice versa. Here, we show the results when model predictions are generated via greedy decoding. The results for constrained decoding are shown in Table \ref{tab:win-tie-loss-rates-logprob-knowledge}.}
    \label{tab:win-tie-loss-rates-knowledge}
    \resizebox{0.65\linewidth}{!}{
    
    \begin{tabular}{@{}l@{\hskip 20pt}c@{\hskip 10pt}c@{\hskip 10pt}c@{\hskip 15pt}c@{\hskip 10pt}c@{\hskip 10pt}c@{\hskip 10pt}}
        \toprule
        \multirow{2}{*}{\textbf{Model}} & \multicolumn{3}{c}{\textbf{Zero-Shot}} & \multicolumn{3}{c}{\textbf{3-Shot}} \\
        \cmidrule{2-4} \cmidrule{5-7}
        & Win & Tie & Loss & Win & Tie & Loss \\
        \midrule
        \textsc{OpenBioLLM-70B}        & 7.7 & 69.2 & \textbf{23.1} & \textbf{30.8} & 69.2 & 0 \\
        \textsc{MediTron-70B}             & 0 & 61.5 & \textbf{38.5} & 0 & 69.2 & \textbf{30.8} \\
        \textsc{Clinical-Camel-70B} & \textbf{27.3} & 63.6 & 9.1 & 0 & 63.6 & \textbf{36.4} \\
        \midrule
        \textsc{OpenBioLLM-8B}         & 0 & 46.2 & \textbf{53.8} & 7.7 & 61.5 & \textbf{30.8} \\
        \textsc{MediTron-7B}              & 0 & 69.2 & \textbf{30.8} & 0 & 23.1 & \textbf{76.9} \\
        \textsc{BioMistral-7B}          & \textbf{30.8} & 69.2 & 0 & \textbf{46.2} & 53.8 & 0 \\
        \textsc{BioMedGPT-LM-7B}         & 0 & 15.4 & \textbf{84.6} & 0 & 7.7 & \textbf{92.3} \\
        \midrule
        \textbf{Aggregate}                                 & 9.0 & 56.2 & \textbf{34.8} & 12.4 & 49.4 & \textbf{38.2} \\
        \bottomrule
    \end{tabular}
    }
\end{table}
In fact, \textbf{in each setting (zero-shot and 3-shot), only 2 out of 7 medical LLMs show statistically significant improvements} in performance (i.e., win more than they lose to their base models), and \textbf{only one medical LLM shows statistically significant improvements in both settings} (Table \ref{tab:win-tie-loss-rates-knowledge}). 
In the zero-shot setting, only \textsc{Clinical-Camel-70B} (\llamatwoseventy{orange}) and \textsc{BioMistral-7B} (\mistral{purple}) show improvements, each achieving win/tie/loss rates of 27.3\%/63.6\%/9.1\% and 30.8\%/69.2\%/0\%, respectively.
In the 3-shot setting, only \textsc{OpenBioLLM-70B} (\llamathreeseventy{blue}) and \textsc{BioMistral-7B} (\mistral{purple}) show significant improvements, each achieving win/tie/loss rates of 30.8\%/69.2\%/0\% and 46.2\%/53.8\%/0\%, respectively. 
Notably, \textsc{BioMistral-7B} is the only medical LLM that consistently outperforms its base model (\textsc{Mistral-7B-Instruct-v0.1}) across both settings, albeit with relatively low absolute performance. 

Meanwhile, we observe that some medical LLMs actually perform significantly worse than their base models---e.g., \textsc{MediTron-7B} and \textsc{BioMedGPT-LM-7B} with loss rates of 76.9\% and 92.3\% in the 3-shot setting. 
When we aggregate the results over all (model pair, textual medical knowledge QA dataset) combinations (last row of Table \ref{tab:win-tie-loss-rates-knowledge}), we find that the medical LLMs achieve win/tie/loss rates of 9.0\%/56.2\%/34.8\% in the zero-shot setting and 12.4\%/49.4\%/38.2\% in the 3-shot setting, indicating that the medical LLMs overall reach a statistical tie with their base models in most cases and only win in a small number of cases.

We similarly observe limited improvements overall with constrained decoding (Appendix \ref{sec:prompting-results-1-knowledge-logprob}). 
When we aggregate the results over all (model pair, textual medical knowledge QA dataset) combinations, medical LLMs achieve win/tie/loss rates of 16.9\%/68.5\%/14.6\% in the zero-shot setting and 11.2\%/74.2\%/14.6\% in the 3-shot setting (Table \ref{tab:win-tie-loss-rates-logprob-knowledge}).
While the aggregate loss rates are generally lower with constrained decoding, the majority of cases still result in a tie, with similar aggregate win and loss rates. 
Meanwhile, we observe that some medical LLMs show larger improvements with constrained decoding. For example, when switching from greedy to constrained decoding, the zero-shot win/tie/loss rates for \textsc{MediTron-70B} change from 0\%/61.5\%/38.5\% to 30.8\%/46.2\%/23.1\%, showing a substantial increase in the win rate. However, the results are mixed, as some other models (e.g., \textsc{Clinical-Camel-70B}, \textsc{OpenBioLLM-70B}) perform worse with constrained decoding.

\textit{In summary, these results show that after appropriately accounting for prompt sensitivity and statistical uncertainty, medical LLMs trained via DAPT show limited improvements in zero-/few-shot prompting performance over their general-domain base models, on textual QA tasks focused on evaluating medical knowledge.}

\subsubsection{Evaluation of Medical LLMs on Textual Clinical Note QA} 
\label{sec:prompting-results-1-clinical}
\begin{figure*}[t!]
    \centering
    \begin{tabular}{@{}c@{}c@{\hskip 7pt}c@{}c@{}}
        \multicolumn{4}{c}{
            \begin{subfigure}{0.95\linewidth}
                \includegraphics[width=\linewidth]{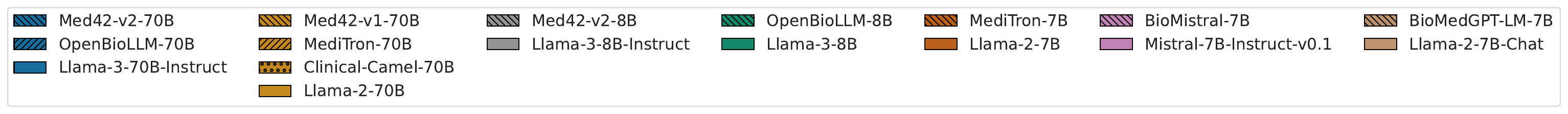}
            \end{subfigure}
        }
        \\
        \begin{subfigure}{0.03\linewidth}
            \makebox[\linewidth]{\raisebox{75pt}{{(a)}}}
        \end{subfigure} &
        \begin{subfigure}{0.64\linewidth}
            \includegraphics[width=\linewidth]{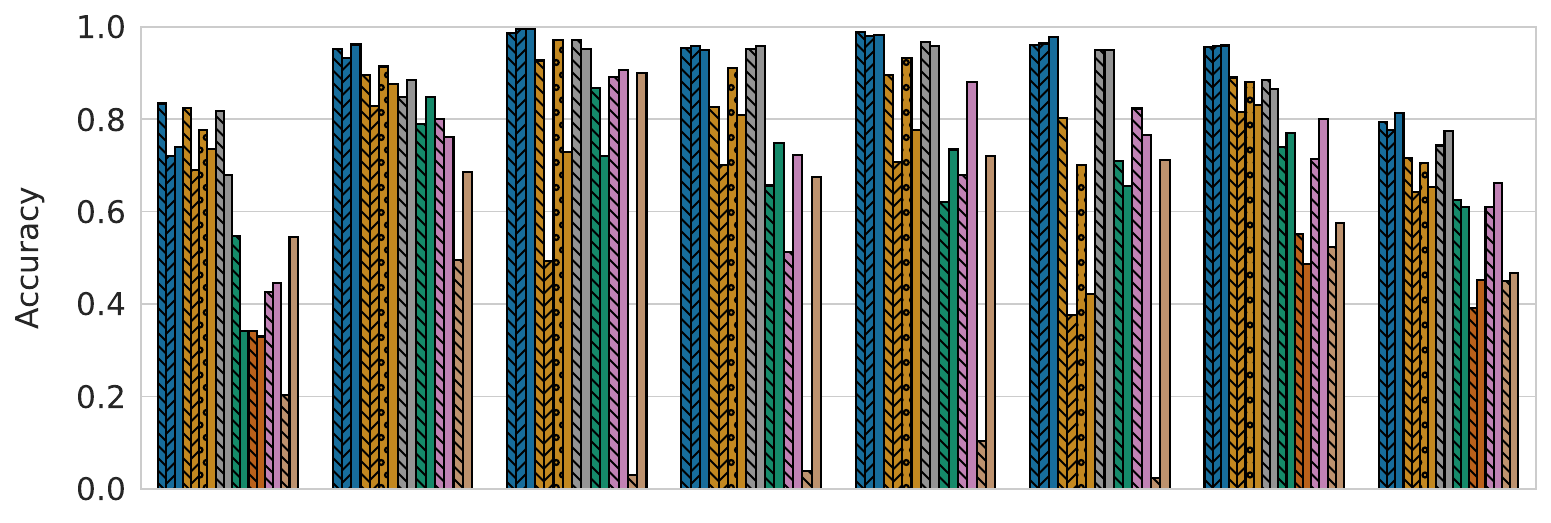}
        \end{subfigure} &
        \begin{subfigure}{0.03\linewidth}
            \makebox[\linewidth]{\raisebox{75pt}{{(b)}}}
        \end{subfigure} &
        \begin{subfigure}{0.26\linewidth}
            \includegraphics[width=\linewidth]{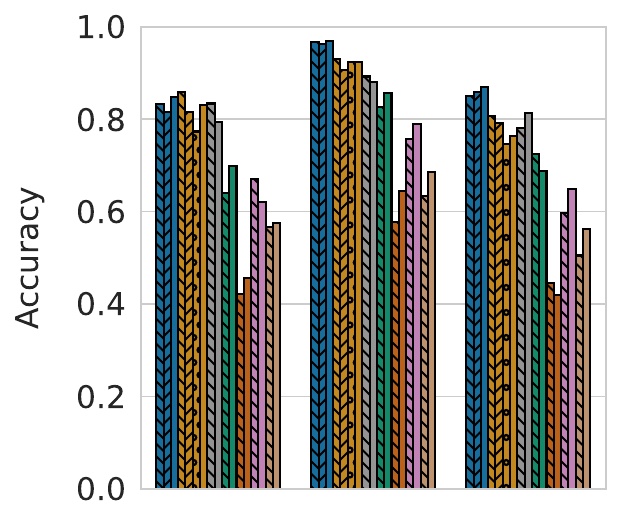}
        \end{subfigure}
        \\
        \begin{subfigure}{0\linewidth}
        \end{subfigure} &
        \begin{subfigure}{0.64\linewidth}
            \includegraphics[width=\linewidth]{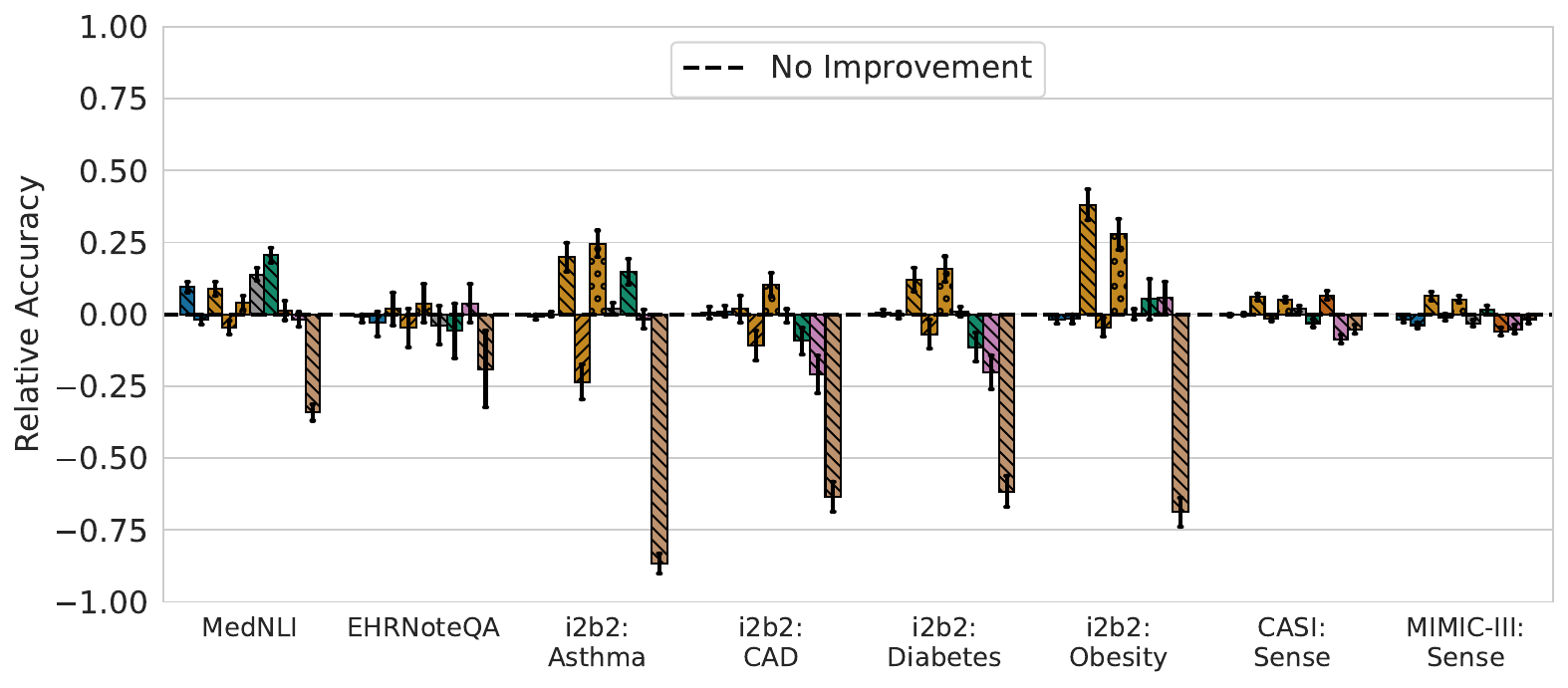}
        \end{subfigure} &
        \begin{subfigure}{0\linewidth}
        \end{subfigure} &
        \begin{subfigure}{0.26\linewidth}
            \includegraphics[width=\linewidth]{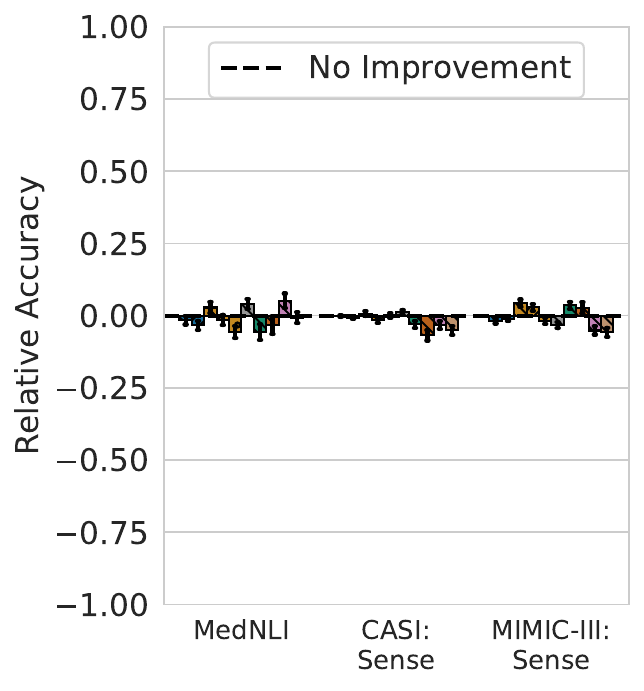}
        \end{subfigure}
    \end{tabular}
    \vspace{-10pt}
    \caption{After independently selecting the best prompt format and examples for each model, medical LLMs (textured bars) fail to consistently improve over their base models (solid bars) on textual clinical note QA tasks, in both (a) zero-shot and (b) 3-shot settings. 
    In the 3-shot setting, we exclude the results on the EHRNoteQA and i2b2 datasets given the context window limitations (Section \ref{sec:prompting}).
    In each panel, the top row shows the absolute exact-match accuracies on the test set, and the bottom row shows the relative exact-match accuracies along with 95\% confidence intervals derived via bootstrapping on the test set (Section~\ref{sec:eval-setup}). 
    Here, model predictions are generated via greedy decoding. 
    Improvements are also limited with constrained decoding (Figure \ref{fig:llm-acc-ci-clinical-logprob}).}
    \label{fig:llm-acc-ci-clinical}
\end{figure*}

\begin{figure*}[t!]
    \centering
    \begin{tabular}{@{}c@{}c@{}c@{}}
        \multicolumn{3}{c}{
            \begin{subfigure}{0.95\linewidth}
                \includegraphics[width=\linewidth]{figs/llm-acc-ci-legend-clinical.pdf}
            \end{subfigure}
        }
        \\
        \begin{subfigure}{0.05\linewidth}
            \makebox[\linewidth]{\raisebox{65pt}{{(a)}}}
        \end{subfigure} &
        \begin{subfigure}{0.45\linewidth}
            \includegraphics[width=\linewidth]{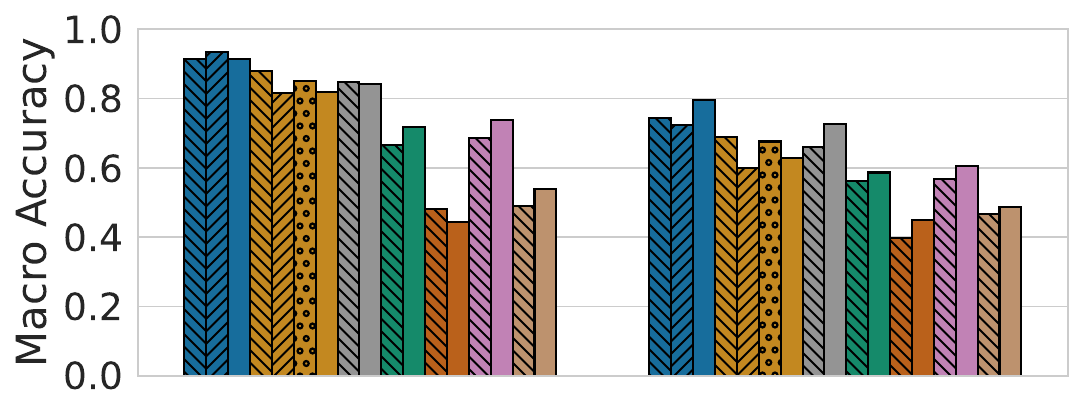}
        \end{subfigure} &
        \begin{subfigure}{0.45\linewidth}
            \includegraphics[width=\linewidth]{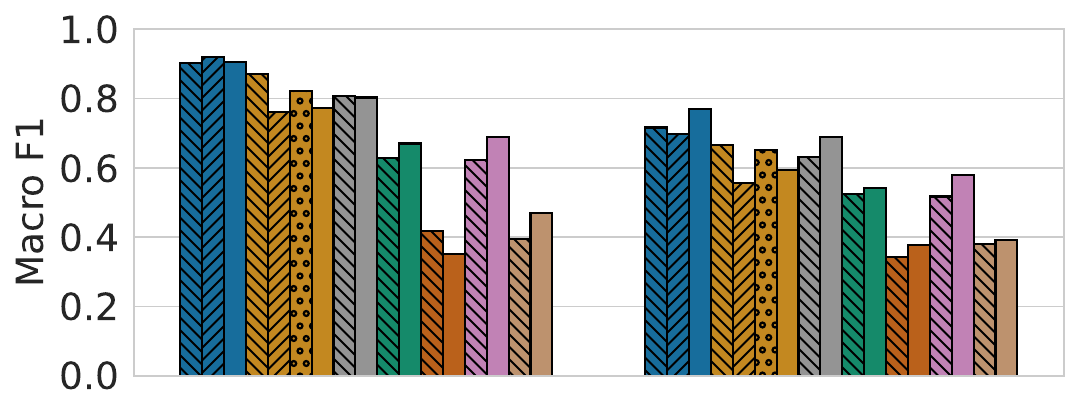}
        \end{subfigure}
        \\
        \begin{subfigure}{0\linewidth}
        \end{subfigure} &
        \begin{subfigure}{0.45\linewidth}
            \includegraphics[width=\linewidth]{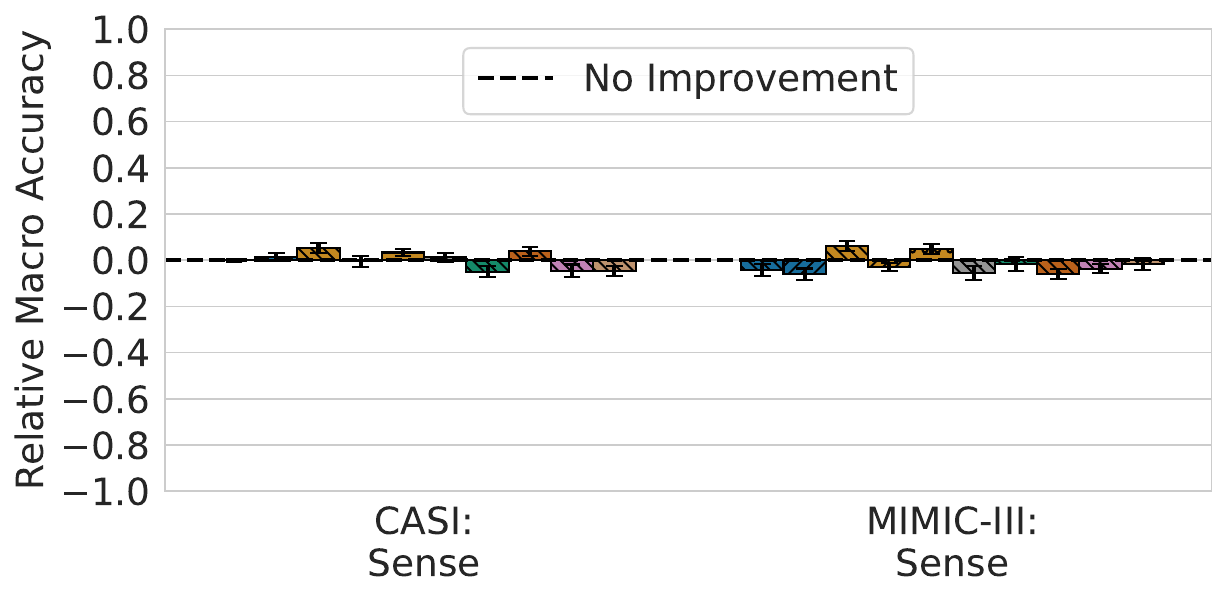}
        \end{subfigure} &
        \begin{subfigure}{0.45\linewidth}
            \includegraphics[width=\linewidth]{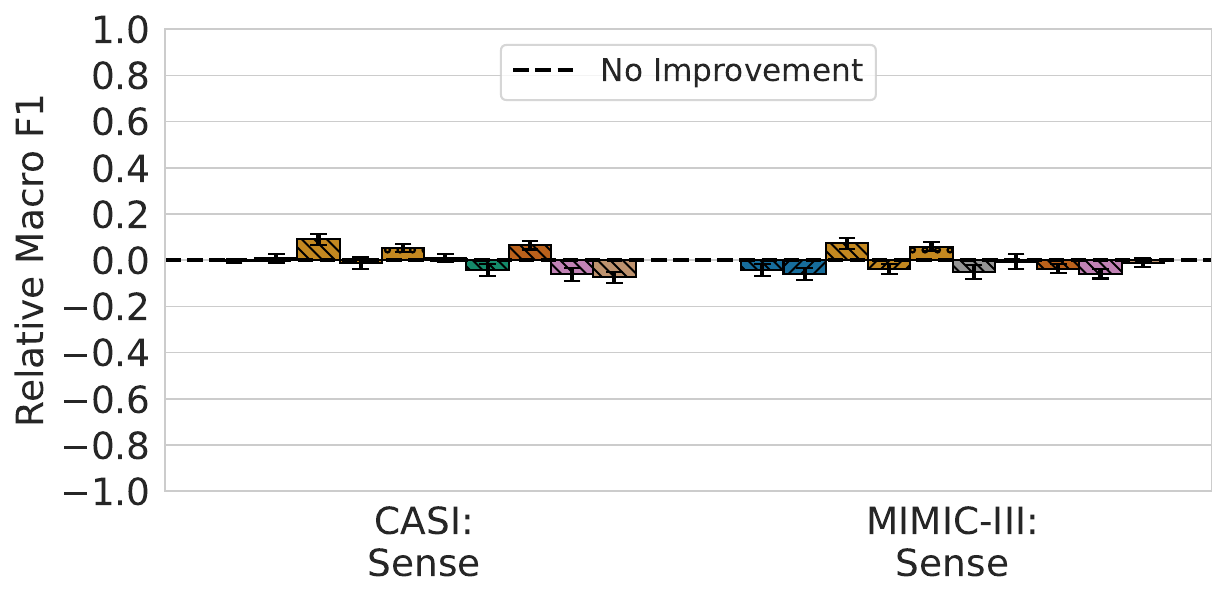}
        \end{subfigure} \\
        \begin{subfigure}{0.05\linewidth}
            \makebox[\linewidth]{\raisebox{65pt}{{(b)}}}
        \end{subfigure} &
        \begin{subfigure}{0.45\linewidth}
            \includegraphics[width=\linewidth]{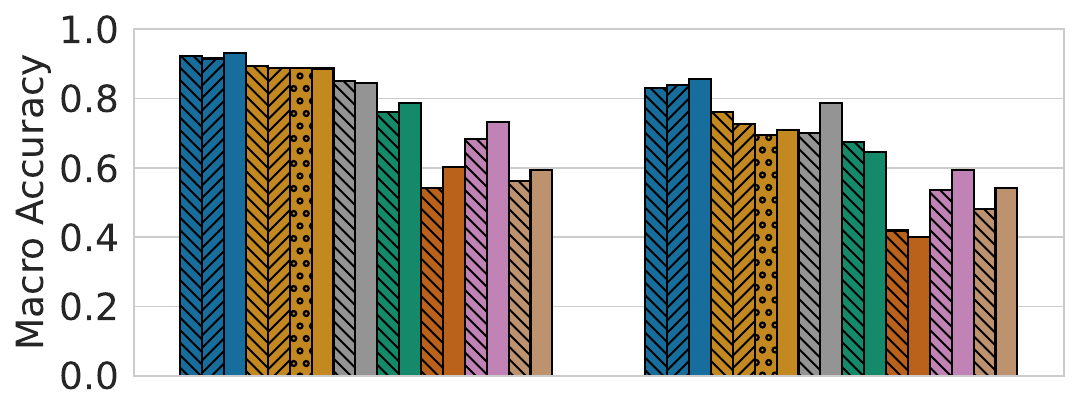}
        \end{subfigure} &
        \begin{subfigure}{0.45\linewidth}
            \includegraphics[width=\linewidth]{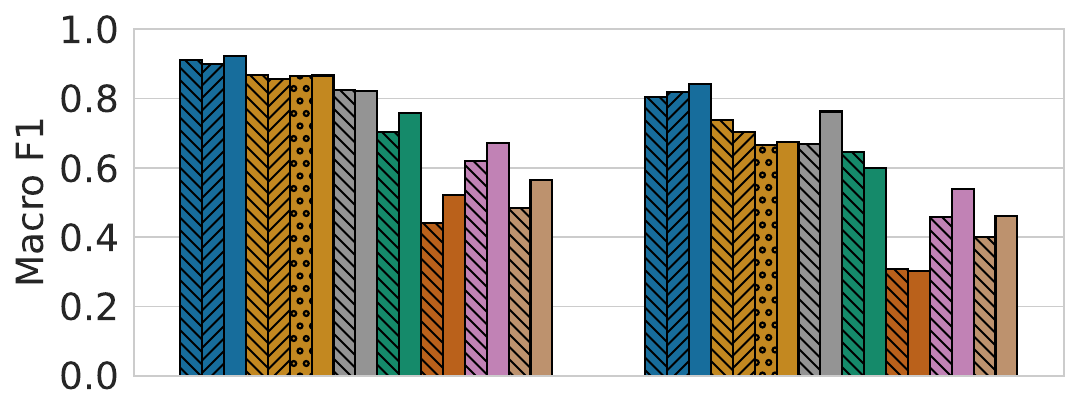}
        \end{subfigure} \\
        \begin{subfigure}{0\linewidth}
        \end{subfigure} &
        \begin{subfigure}{0.45\linewidth}
            \includegraphics[width=\linewidth]{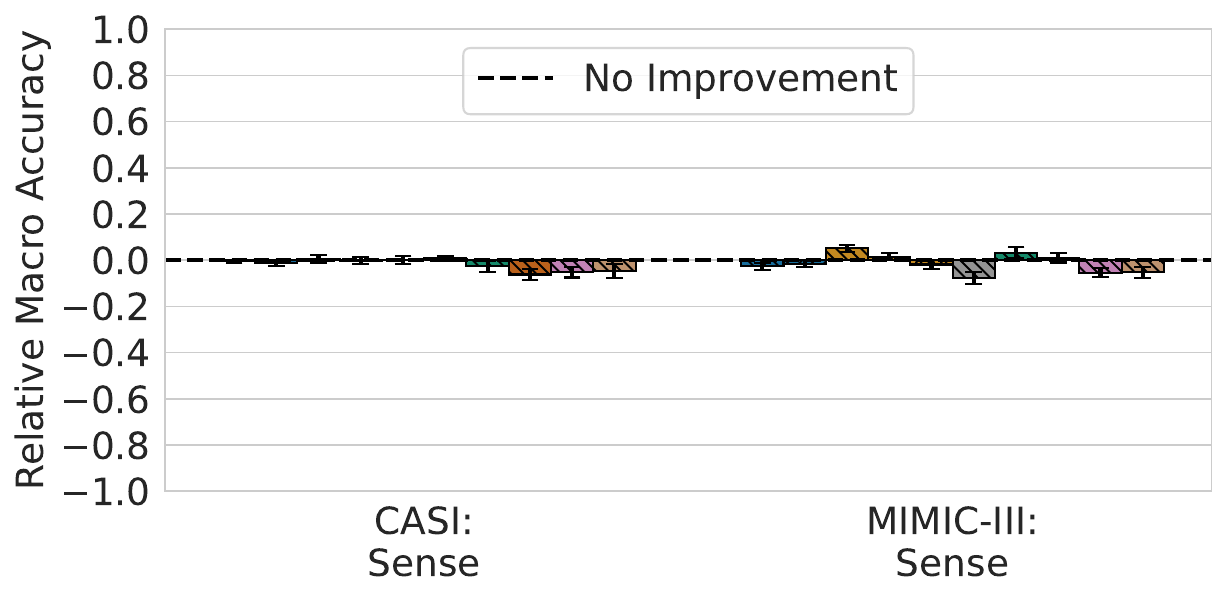}
        \end{subfigure} &
        \begin{subfigure}{0.45\linewidth}
            \includegraphics[width=\linewidth]{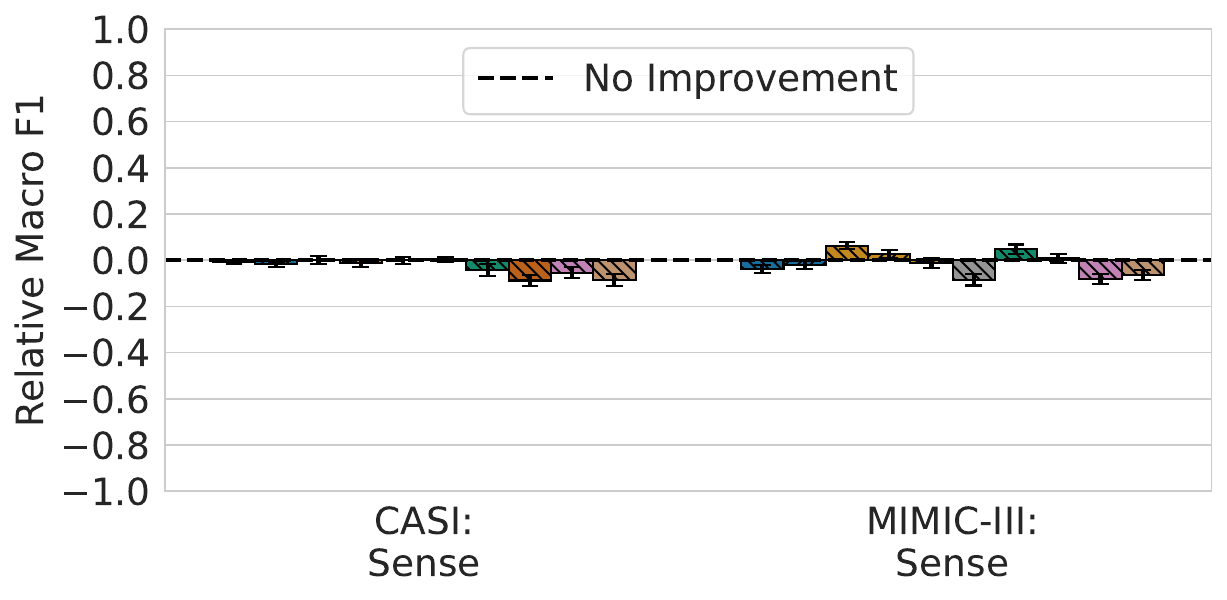}
        \end{subfigure}
    \end{tabular}
    \vspace{-10pt}
    \caption{Even after accounting for the imbalance in the distribution of clinical acronyms in the CASI and MIMIC-III datasets, medical LLMs (textured bars) fail to consistently improve over their base models (solid bars), in both (a) zero-shot and (b) 3-shot settings. 
    We show the results when the prompts are optimized for each model independently. 
    In each panel, the top row shows the absolute macro exact-match accuracies and F1 scores---averaged over clinical acronyms---on the test set, and the bottom row shows the relative macro exact-match accuracies and F1 scores along with 95\% confidence intervals derived via bootstrapping on the test set (Section~\ref{sec:eval-setup}). 
    Here, model predictions are generated via greedy decoding. 
    Improvements are also limited with constrained decoding (Figure \ref{fig:llm-acc-ci-sense-logprob}).}
    \label{fig:llm-acc-ci-sense}
\end{figure*}

In Figures \ref{fig:llm-acc-ci-clinical}(a) and (b), we show the absolute and relative exact-match accuracies achieved by the medical and general-domain LLMs on the textual clinical note QA datasets, from zero-shot and 3-shot prompting, respectively. 
For the CASI and MIMIC-III sense disambiguation datasets, we also show the absolute and relative \textit{macro} exact-match accuracies and F1 scores---averaged over clinical acronyms---in Figure \ref{fig:llm-acc-ci-sense}, as the distribution of clinical acronyms in these datasets is highly imbalanced \citep{lmc}. 
In Table \ref{tab:win-tie-loss-rates-clinical}, we show the win, tie, and loss rates (\%) of the medical LLMs, where win rate refers to the proportion of QA datasets where a medical model shows a statistically significant improvement over its base model. 
As described in Section \ref{sec:prompting}, we exclude the 3-shot results for the EHRNoteQA and i2b2 datasets, as a clinical note from even a single QA pair from these datasets on average already fills the context window for most models. 
Additionally, we exclude the zero-shot results on these datasets for \textsc{MediTron-7B} (and its base model \textsc{Llama-2-7B}), as its small context window size of 2k tokens is insufficient even in the zero-shot setting.

In contrast to the zero-shot results on textual medical knowledge QA tasks (Figure \ref{fig:llm-acc-ci-knowledge}), the extent of zero-shot performance improvement achieved by the medical LLMs varies more significantly across different model pairs and clinical note QA tasks (Figure \ref{fig:llm-acc-ci-clinical}(a)). 
In particular, the results vary the most on the four i2b2 datasets, where \textsc{Med42-v1-70B} and \textsc{Clinical-Camel-70B} (\llamatwoseventy{orange}) show large improvements (around 20+\% in absolute terms), while \textsc{MediTron-70B} (\llamatwoseventy{orange}), \textsc{BioMistral-7B} (\mistral{purple}), and \textsc{BioMedGPT-LM-7B} (\llamatwosevenchat{brown}) tend to perform worse than their general-domain counterparts. 
In part, such variability across datasets results from significant differences in the overall syntax, length, and format of the clinical notes provided as context. 
For the i2b2 datasets, the clinical notes are provided in full with minimal preprocessing, while other datasets often only include short snippets with just a few sentences extracted from the full note (e.g., MedNLI, CASI, MIMIC-III) or apply more extensive preprocessing (e.g., EHRNoteQA).

Meanwhile, \textsc{BioMedGPT-LM-7B} (\llamatwosevenchat{brown}) generally fails to output an appropriate response on these datasets even after performing model-specific prompt selection as in Section \ref{sec:prompting}, often refusing to produce an answer or generating irrelevant text completions. 
We note that such behavior is partially attributable to the lack of details on the recommended prompting setup for \textsc{BioMedGPT-LM-7B} (Appendix \ref{sec:reproducibility}).
\textsc{BioMedGPT-LM-7B} does achieve higher accuracy upon constraining its output vocabulary (i.e., constrained decoding), albeit with limited performance improvements over its base model (Figure \ref{fig:llm-acc-ci-clinical-logprob}).

\begin{table}[t!]
    \centering
    \caption{The zero-shot and 3-shot win, tie, and loss rates (\%) of all medical LLMs on textual clinical note QA, after independently optimizing the prompt for each model. For each medical model, we boldface the win rate if it wins more than it loses to its general-domain base model, and vice versa. Here, we show the results when model predictions are generated via greedy decoding. The results for constrained decoding are shown in Table \ref{tab:win-tie-loss-rates-logprob-clinical}.}
    \label{tab:win-tie-loss-rates-clinical}
    \resizebox{0.65\linewidth}{!}{
    
    \begin{tabular}{@{}l@{\hskip 20pt}c@{\hskip 10pt}c@{\hskip 10pt}c@{\hskip 15pt}c@{\hskip 10pt}c@{\hskip 10pt}c@{\hskip 10pt}}
        \toprule
        \multirow{2}{*}{\textbf{Model}} & \multicolumn{3}{c}{\textbf{Zero-Shot}} & \multicolumn{3}{c}{\textbf{3-Shot}} \\
        \cmidrule{2-4} \cmidrule{5-7}
        & Win & Tie & Loss & Win & Tie & Loss \\
        \midrule
        \textsc{Med42-v2-70B}             & 12.5 & 62.5 & \textbf{25.0} & 0 & 33.3 & \textbf{66.7} \\
        \textsc{OpenBioLLM-70B}        & 0 & 75.0 & \textbf{25.0} & 0 & 0 & \textbf{100.0} \\
        \textsc{Med42-v1-70B}             & \textbf{75.0} & 25.0 & 0 & \textbf{66.7} & 33.3 & 0 \\
        \textsc{MediTron-70B}             & 0 & 25.0 & \textbf{75.0} & 33.3 & 33.3 & 33.3 \\
        \textsc{Clinical-Camel-70B} & \textbf{87.5} & 12.5 & 0 & 0 & 33.3 & \textbf{66.7} \\
        \midrule
        \textsc{Med42-v2-8B}              & \textbf{37.5} & 50.0 & 12.5 & \textbf{66.7} & 0 & 33.3 \\
        \textsc{OpenBioLLM-8B}         & 25.0 & 37.5 & \textbf{37.5} & 33.3 & 0 & \textbf{66.7} \\
        \textsc{MediTron-7B}              & 33.3 & 33.3 & 33.3 & 33.3 & 0 & \textbf{66.7} \\
        \textsc{BioMistral-7B}          & 12.5 & 37.5 & \textbf{50.0} & 33.3 & 0 & \textbf{66.7} \\
        \textsc{BioMedGPT-LM-7B}         & 0 & 0 & \textbf{100.0} & 0 & 33.3 & \textbf{66.7} \\
        \midrule
        \textbf{Aggregate}                                 & 28.0 & 36.0 & \textbf{36.0} & 26.7 & 16.7 & \textbf{56.7} \\
        \bottomrule
    \end{tabular}
    }
\end{table}

In \textit{aggregate}, we find that the medical models do not consistently improve over their base models in the zero-shot setting (Figure \ref{fig:llm-acc-ci-clinical}(a)).
In fact, \textbf{only 3 out of 10 medical LLMs}---\textsc{Med42-v1-70B}, \textsc{Clinical-Camel-70B}, and \textsc{Med42-v2-8B}---\textbf{win more than they lose to their base models in the zero-shot setting} (Table \ref{tab:win-tie-loss-rates-clinical}).
Across all (model pair, textual clinical note QA dataset) combinations, the medical LLMs achieve zero-shot win/tie/loss rates of 28.0\%/36.0\%/36.0\%. 
Meanwhile, \textsc{Med42-v1-70B} and \textsc{Clinical-Camel-70B} show substantial improvements in the zero-shot setting, respectively achieving win rates of 75.0\% and 87.5\%, and both achieving a loss rate of 0\%.
The two models also tend to show strong performance on the i2b2 datasets, 
which, unlike other datasets, are based on minimally preprocessed, full-length clinical notes. 
A notable difference between these two models and others is that their corresponding medical adaptation corpora include datasets beyond PubMed articles (e.g., medical knowledge QA datasets), often in \textit{conversational formats} (Table \ref{tab:models}).
Given that both were adapted from \textsc{Llama-2-70B}, which has not been instruction-tuned and thus may not reliably generate an appropriate response in the zero-shot setting, we note that the perceived performance improvements may be overestimated in this setting.

In the 3-shot setting, 
where even non-instruction-tuned LLMs generally
output a relevant response in the correct format, 
the majority of medical LLMs 
show little to no improvement over their base models, 
with the 95\% confidence intervals in relative exact-match accuracy crossing or lying below zero
(Figure \ref{fig:llm-acc-ci-clinical}(b)).
\textbf{Only 2 out of 10 medical LLMs}---\textsc{Med42-v1-70B} and \textsc{Med42-v2-8B}---\textbf{show significant improvements in the 3-shot setting} 
(Table \ref{tab:win-tie-loss-rates-clinical}). 
When aggregated across all (model pair, textual clinical note QA dataset) combinations, the medical LLMs achieve 3-shot win/tie/loss rates of 26.7\%/16.7\%/56.7\%. 

On the CASI and MIMIC-III sense disambiguation datasets, the medical LLMs fail to show significant improvements over their base models \textit{even after accounting for the imbalance in the distribution of clinical acronyms}, in both zero-shot (Figure \ref{fig:llm-acc-ci-sense}(a)) and 3-shot (Figure \ref{fig:llm-acc-ci-sense}(b)) settings. 
On the CASI dataset, only 3 out of 10 medical LLMs show improvement in the zero-shot setting in terms of macro-average accuracy, and none of the medical LLMs show improvement in the 3-shot setting. 
In terms of macro-average F1 score, only 2 out of 10 medical LLMs show improvement in the zero-shot setting, while none of the medical LLMs show improvement in the 3-shot setting.
On the MIMIC-III dataset, only 2 out of 10 medical LLMs show improvement in the zero-shot setting in terms of each metric, and only 3 out of 10 medical LLMs show improvement in the 3-shot setting in terms of each metric.

We similarly observe limited improvements overall with constrained decoding (Appendix \ref{sec:prompting-results-1-clinical-logprob}). When we aggregate the results over all (model pair, textual clinical note QA dataset) combinations, medical LLMs achieve win/tie/loss rates of 34.7\%/37.3\%/28.0\% in the zero-shot setting and 33.3\%/23.3\%/43.3\% in the 3-shot setting (Table \ref{tab:win-tie-loss-rates-logprob-clinical}). 
While the aggregate loss rates are generally lower with constrained decoding, 
the win rates are roughly equivalent to the loss rates modulo the statistical ties. 
We also see limited improvements in terms of macro-average accuracy and F1 score on the CASI and MIMIC-III datasets (Figure \ref{fig:llm-acc-ci-sense-logprob}).
Meanwhile, as for the textual medical knowledge QA tasks, some medical LLMs show larger improvements with constrained decoding. 
For example, when switching from greedy to constrained decoding, we observe that the zero-shot win/tie/loss rates for \textsc{BioMedGPT-LM-7B} change from 0\%/0\%/100.0\% to 50.0\%/12.5\%/37.5\%, showing a substantial increase in the win rate. 
However, the results are again mixed, as some other models (e.g., \textsc{MediTron-7B} and \textsc{BioMistral-7B}) perform worse with constrained decoding.

\textit{In summary, these results demonstrate that on textual QA tasks based on real-world clinical notes, medical LLMs trained via DAPT show limited improvements in zero-/few-shot prompting performance over their general-domain counterparts after appropriately accounting for prompt sensitivity and statistical uncertainty.}

\subsubsection{Evaluation of Medical VLMs on Visual Medical QA} 
\label{sec:prompting-results-1-vqa}
\begin{figure*}[t!]
    \centering
    \begin{tabular}{@{}c@{}c@{\hskip 5pt}c@{}c@{}}
        \multicolumn{4}{c}{
            \begin{subfigure}{0.8\linewidth}
                \includegraphics[width=\linewidth]{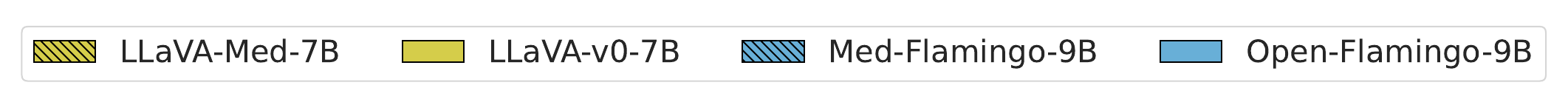}
            \end{subfigure}
        }
        \\
        \begin{subfigure}{0.03\linewidth}
            \makebox[\linewidth]{\raisebox{45pt}{{(a)}}}
        \end{subfigure} &
        \begin{subfigure}{0.45\linewidth}
            \includegraphics[width=\linewidth]{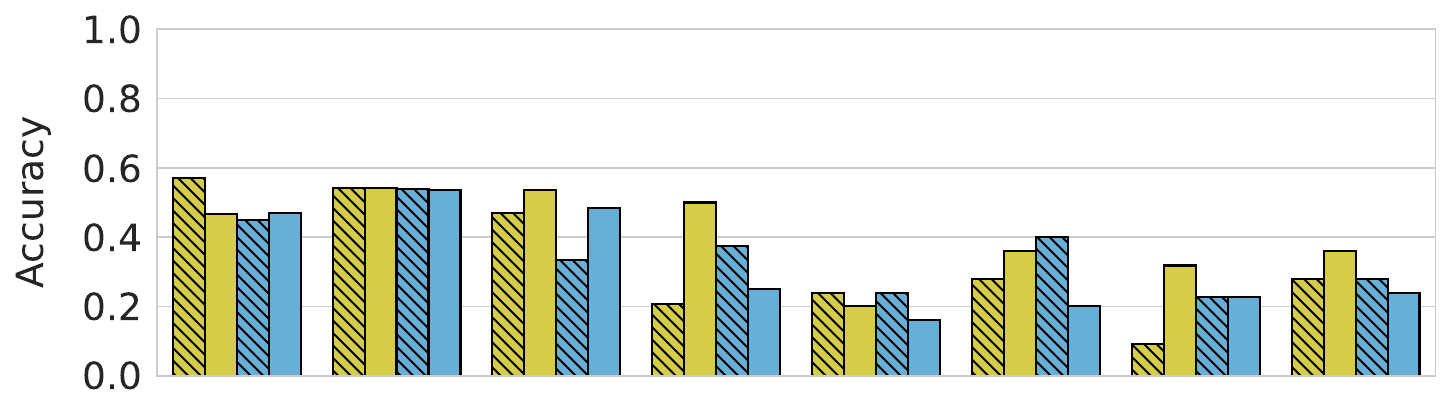}
        \end{subfigure} &
        \begin{subfigure}{0.03\linewidth}
            \makebox[\linewidth]{\raisebox{45pt}{{(b)}}}
        \end{subfigure} &
        \begin{subfigure}{0.45\linewidth}
            \includegraphics[width=\linewidth]{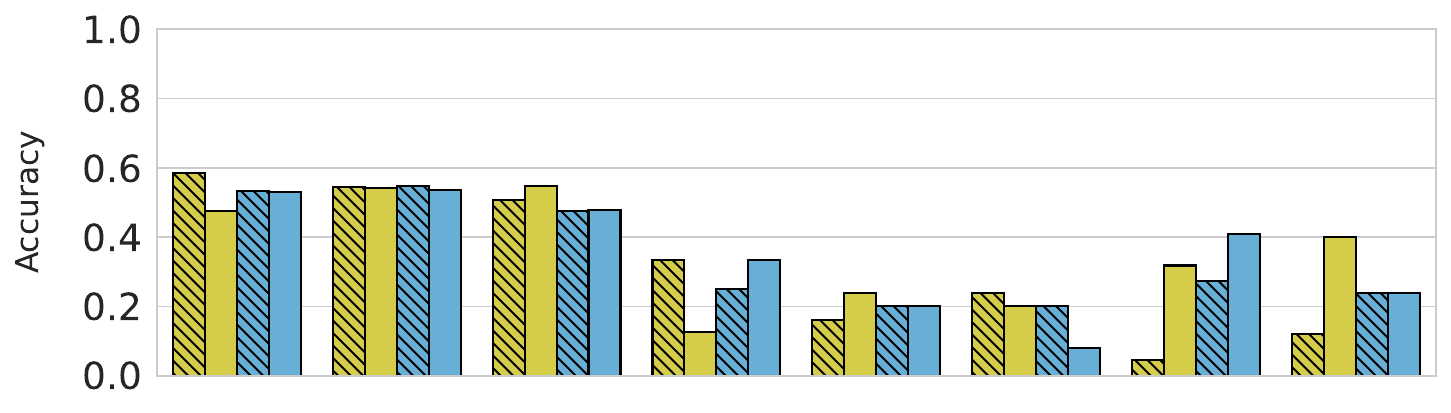}
        \end{subfigure}
        \\
        \begin{subfigure}{0.03\linewidth}
        \end{subfigure} &
        \begin{subfigure}{0.45\linewidth}
            \includegraphics[width=\linewidth]{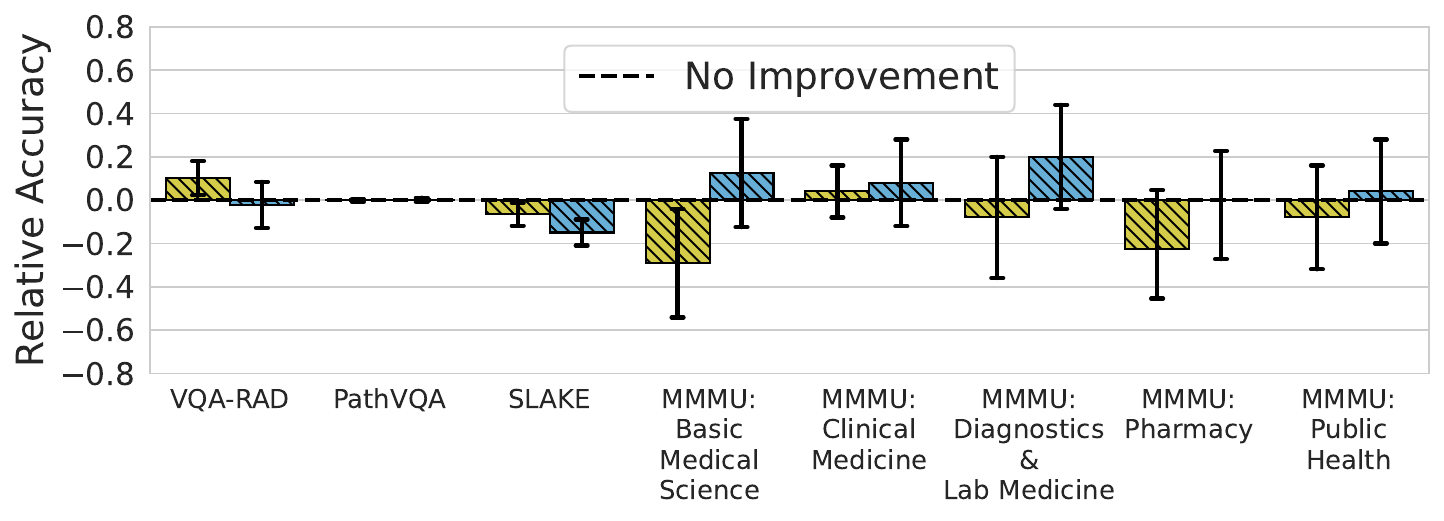}
        \end{subfigure} &
        \begin{subfigure}{0.03\linewidth}
        \end{subfigure} &
        \begin{subfigure}{0.45\linewidth}
            \includegraphics[width=\linewidth]{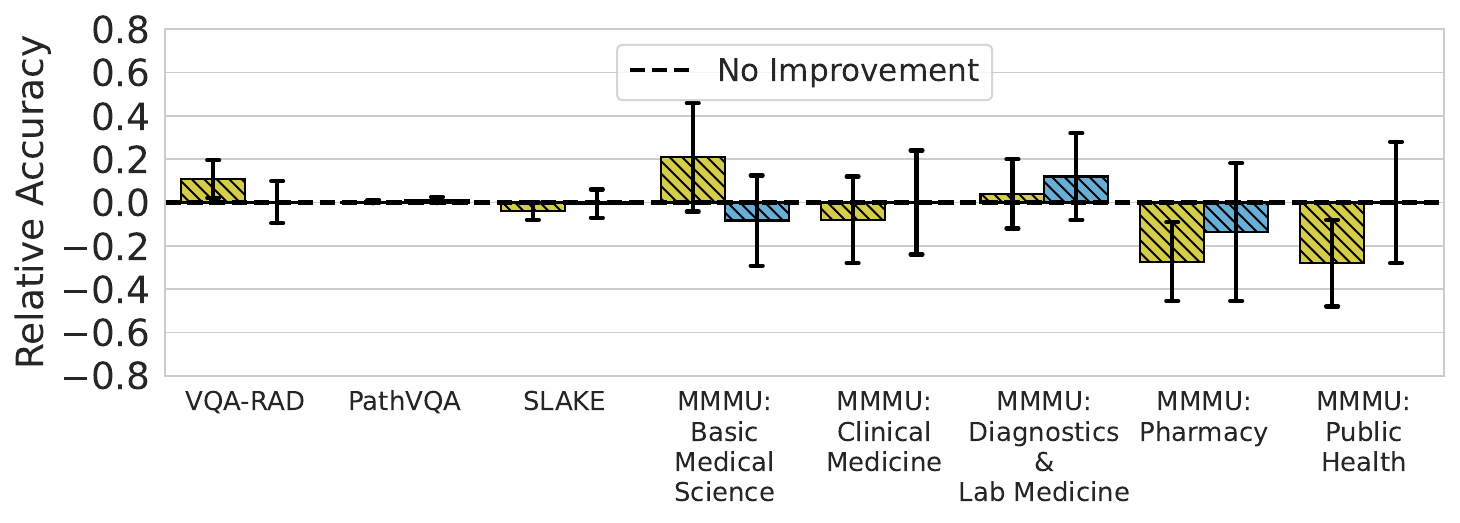}
        \end{subfigure}
    \end{tabular}
    \vspace{-10pt}
    \caption{
    After independently selecting the best prompt format and examples for each model, medical VLMs (textured bars) fail to consistently improve over their base models (solid bars) on visual medical QA tasks, in both (a) zero-shot and (b) 3-shot settings. 
    In each panel, the top row shows the absolute exact-match accuracies on the test set, and the bottom row shows the relative exact-match accuracies along with 95\% confidence intervals derived via bootstrapping on the test set (Section~\ref{sec:eval-setup}). 
    Here, model predictions are generated via greedy decoding. 
    Improvements are also limited with constrained decoding (Figure \ref{fig:vlm-logprob-acc-ci}).}
    \label{fig:vlm-acc-ci}
\end{figure*}

In Figures \ref{fig:vlm-acc-ci}(a) and (b), we show the absolute and relative exact-match accuracies achieved by the medical and general-domain VLMs on the visual medical QA datasets, from zero-shot and 3-shot prompting, respectively. 
Table \ref{tab:win-tie-loss-rates-vlm} shows the win, tie, and loss rates (\%) of the medical VLMs, where win rate refers to the proportion of visual medical QA datasets where a medical model shows a statistically significant improvement over its base model. 
We note that both \textsc{LLaVA-Med-7B} and its base model \textsc{LLaVA-v0-7B} were not pretrained to handle multi-image inputs and may not perform better with more in-context examples.

In both zero-shot and 3-shot settings, we find that the absolute exact-match accuracies are mostly similar between each model pair on VQA-RAD, PathVQA, and SLAKE (Figure \ref{fig:vlm-acc-ci}). 
On the three datasets, the absolute accuracies tend to be higher for all models in the 3-shot vs.~the zero-shot setting, but the pairwise differences remain similarly small in both settings. 
For MMMU-Medical on the other hand, we see that the extent of performance improvement (or lack thereof) varies significantly across the two model pairs and datasets, and that the absolute exact-match accuracies do not exhibit an increasing trend going from zero-shot to 3-shot prompting.
For example, when comparing the absolute accuracies between each VLM pair in the zero-shot setting, \textsc{LLaVA-Med-7B} generally performs significantly worse than \textsc{LLaVA-v0-7B} (\llava{light green}), while \textsc{Med-Flamingo-9B} generally performs significantly better than \textsc{Open-Flamingo-9B} (\flamingo{cyan}) across all subjects in MMMU-Medical. 
Meanwhile, on the MMMU-Medical datasets, the 95\% confidence intervals in relative exact-match accuracy almost always cross zero, indicating that the observed differences are \textit{not} statistically significant (bottom rows of Figure \ref{fig:vlm-acc-ci}(a) and (b)).
We note that the confidence intervals are much wider on MMMU-Medical than on the other datasets, as the test sets only include 25 QA examples per subject in MMMU-Medical (Table \ref{tab:datasets}).

When we aggregate the results across all model pairs and visual medical QA datasets while accounting for such statistical uncertainty, we find that the medical VLMs only achieve win/tie/loss rates of 6.3\%/75.0\%/18.8\% in the zero-shot setting and 6.3\%/81.3\%/12.5\% in the 3-shot setting (Table \ref{tab:win-tie-loss-rates-vlm}).
In fact, we find that both \textsc{LLaVA-Med-7B} and \textsc{Med-Flamingo-9B} are \textbf{virtually indistinguishable from their base models in terms of performance}: \textsc{LLaVA-Med-7B} achieves win/tie/loss rates of 12.5\%/62.5\%/25.0\% in both zero-shot and 3-shot settings, while \textsc{Med-Flamingo-9B} achieves win/tie/loss rates of 0\%/87.5\%/12.5\% in the zero-shot setting and 0\%/100\%/0\% in the 3-shot setting. 

We observe that these conclusions also hold when the model predictions are generated via constrained decoding (Appendix \ref{sec:prompting-results-1-vqa-logprob}). 
The medical VLMs almost always reach a tie with their base models in both zero-shot and 3-shot settings, with the 95\% confidence intervals in relative exact-match accuracy crossing zero (Figure \ref{fig:vlm-logprob-acc-ci}).
Collectively, the medical VLMs achieve win/tie/loss rates of 6.3\%/87.5\%/6.3\% in the zero-shot setting and 0\%/93.8\%/6.3\% in the 3-shot setting (Table \ref{tab:win-tie-loss-rates-logprob-vlm}).
In fact, \textbf{no medical VLM shows improvement over its general-domain base model, regardless of the decoding strategy.}

\begin{table}[t!]
    \centering
    \caption{The zero-shot and 3-shot win, tie, and loss rates (\%) of all medical VLMs on visual medical QA, after independently optimizing the prompt for each model. For each medical model, we boldface the win rate if it wins more than it loses to its general-domain base model, and vice versa. Here, we show the results when model predictions are generated via greedy decoding. The results for constrained decoding are shown in Table \ref{tab:win-tie-loss-rates-logprob-vlm}.}
    \label{tab:win-tie-loss-rates-vlm}
    \resizebox{0.65\linewidth}{!}{
    
    \begin{tabular}{@{}l@{\hskip 20pt}c@{\hskip 10pt}c@{\hskip 10pt}c@{\hskip 15pt}c@{\hskip 10pt}c@{\hskip 10pt}c@{\hskip 10pt}}
        \toprule
        \multirow{2}{*}{\textbf{Model}} & \multicolumn{3}{c}{\textbf{Zero-Shot}} & \multicolumn{3}{c}{\textbf{3-Shot}} \\
        \cmidrule{2-4} \cmidrule{5-7}
        & Win & Tie & Loss & Win & Tie & Loss \\
        \midrule
        \textsc{LLaVA-Med-7B}    & 12.5 & 62.5 & \textbf{25.0} & 12.5 & 62.5 & \textbf{25.0} \\
        \textsc{Med-Flamingo-9B} & 0 & 87.5 & \textbf{12.5} & 0 & 100.0 & 0 \\
        \midrule
        \textbf{Aggregate}                            & 6.3 & 75.0 & \textbf{18.8} & 6.3 & 81.3 & \textbf{12.5} \\
        \bottomrule
    \end{tabular}
    }
\end{table}

\textit{In summary, these results show that after appropriately accounting for prompt sensitivity and statistical uncertainty, medical VLMs trained via DAPT show limited improvements in zero-/few-shot prompting performance over their general-domain counterparts, on visual medical QA tasks spanning various medical imaging modalities (e.g., pathology, radiology).}

\subsection{Overlooking Prompt Sensitivity and Statistical Uncertainty May Overestimate the Performance Benefits from Medical DAPT}
\label{sec:prompting-results-2}

Based on our findings in Section \ref{sec:prompting-results-1}, we further investigate whether the conclusions differ if the same prompt is used for each pair of medical and general-domain models. 
In particular, we consider whether selecting a prompt only for the medical model, following Section \ref{sec:prompting}, and using it for the corresponding general-domain base model can widen the performance gap between each pair.
We also assess whether this gap becomes amplified when models are compared without accounting for statistical uncertainty, which is often observed in practice.
More concretely, we evaluate how the win/tie/loss rates (\%) of the medical models change across all QA tasks, as we vary the following aspects of the experimental setup:
\begin{enumerate}[topsep=0.5ex,itemsep=-0.5ex]
    \item select prompts for each model independently vs.~only based on the medical model;
    \item determine the winner for each model pair based on statistical testing (allowing for ties) vs.~comparing only based on absolute accuracy.
\end{enumerate}
Note that when comparing each model pair based on raw absolute accuracy, there are no ties, as the real-valued absolute accuracies are rarely identical. 

\begin{figure*}[t!]
    \centering
    \begin{tabular}{@{}c@{}c@{}}
        \begin{subfigure}{0.03\linewidth}
            \makebox[\linewidth]{\raisebox{55pt}{{(a)}}}
        \end{subfigure} &
        \begin{subfigure}{0.9\linewidth}
            \includegraphics[width=\linewidth]{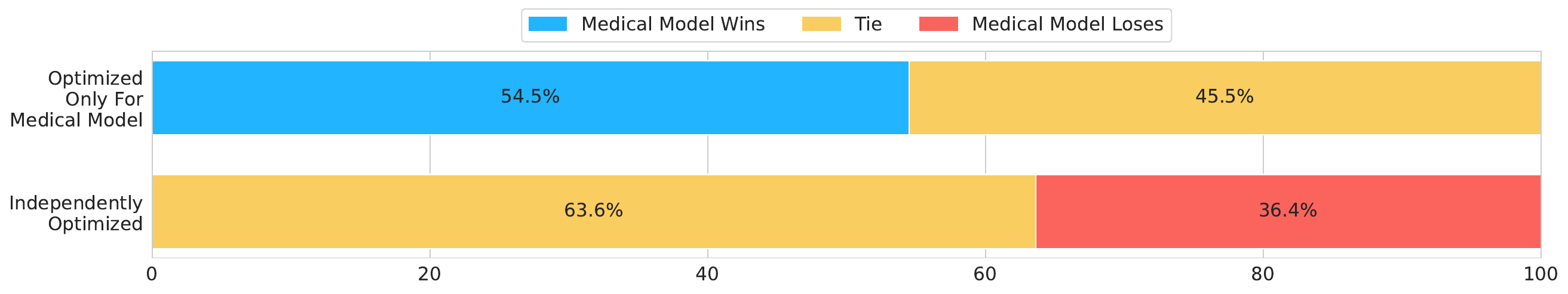}
        \end{subfigure}
        \\
        \begin{subfigure}{0\linewidth}
            \makebox[\linewidth]{\raisebox{55pt}{{(b)}}}
        \end{subfigure} &
        \begin{subfigure}{0.9\linewidth}
            \includegraphics[width=\linewidth]{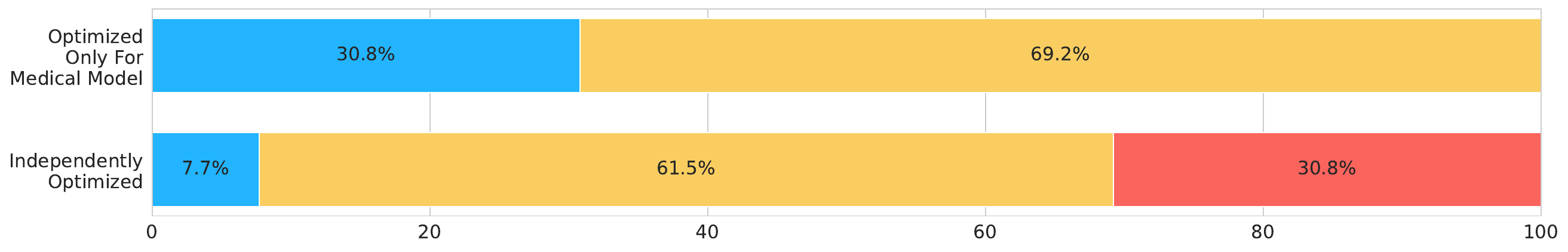}
        \end{subfigure} \\
        \begin{subfigure}{0\linewidth}
            \makebox[\linewidth]{\raisebox{65pt}{{(c)}}}
        \end{subfigure} &
        \begin{subfigure}{0.9\linewidth}
            \includegraphics[width=\linewidth]{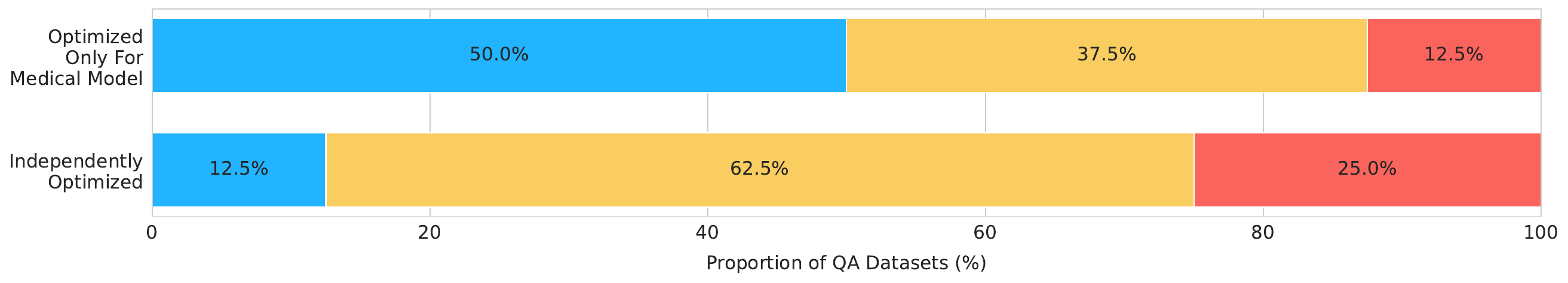}
        \end{subfigure} 
    \end{tabular}
    \vspace{-10pt}
    \caption{Using a single, fixed prompt only optimized for the medical model can overestimate the performance improvements from medical DAPT: (a) \textsc{Clinical-Camel-70B}, (b) \textsc{OpenBioLLM-8B}, (c) \textsc{LLaVA-Med-7B}. 
    ``Optimized Only For Medical Model'' refers to the setting where we perform model-specific prompt selection (Section \ref{sec:prompting}) only for the medical model and use the selected prompt for both models in each pair.
    ``Independently Optimized'' refers to the setting where we separately perform model-specific prompt selection for each model in a given pair.
    For \textsc{Clinical-Camel-70B} and \textsc{OpenBioLLM-8B}, we show the 3-shot win/tie/loss rates (\%) on the textual medical knowledge QA tasks (trends are similar for the textual clinical note QA tasks; see Section \ref{sec:prompting-results-2-aggregate}).
    For \textsc{LLaVA-Med-7B}, we show the 3-shot win/tie/loss rates (\%) on all visual medical QA tasks.}
    \label{fig:medopt-prompt}
\end{figure*}

\textbf{Overall, we find that the perceived performance benefits from medical DAPT can be significantly overestimated without appropriately accounting for prompt sensitivity and statistical uncertainty in the results.}
We consistently observe this across all LLM/VLM pairs and medical QA tasks.
Using specific model pairs as examples, we illustrate our findings regarding each aspect in Sections \ref{sec:prompting-results-2-prompt}--\ref{sec:prompting-results-2-uncertainty}, and discuss the aggregate result across all model pairs and QA datasets in Section \ref{sec:prompting-results-2-aggregate}.

\subsubsection{Selecting the ``Right'' Prompt Independently is Crucial for a Fair Comparison Between Models}
\label{sec:prompting-results-2-prompt}

In Figure \ref{fig:medopt-prompt}, we show how the win, tie, and loss rates (\%) change when we optimize the prompt only for the medical model, using \textsc{Clinical-Camel-70B}, \textsc{OpenBioLLM-8B}, and \textsc{LLaVA-Med-7B} as illustrative examples. 
For \textsc{Clinical-Camel-70B} and \textsc{OpenBioLLM-8B}, we show the 3-shot results on the textual medical knowledge QA datasets (see Section \ref{sec:prompting-results-2-aggregate} for discussion of results on textual clinical note QA, which are similar).
For \textsc{LLaVA-Med-7B}, we show the 3-shot results on all visual medical QA datasets. 
We note that the results here are shown \textit{with} statistical testing, i.e., the win/tie/loss rates are calculated based on the 95\% bootstrapping confidence intervals, as described in Section \ref{sec:eval-setup}.

In all three cases, we observe that when we optimize the prompt only for the medical model, the win rates increase significantly and the loss rates decrease significantly.
For example, we find that the win rate for \textsc{Clinical-Camel-70B} increases from 0\% to 54.5\% while its loss rate decreases from 36.4\% to 0\% under this setting, completely reversing the conclusion about \textsc{Clinical-Camel-70B} (Figure \ref{fig:medopt-prompt}(a)).
We also observe that such trends exist regardless of model scale: across both 70B-parameter and 7/8B-parameter models, the results are highly sensitive to the choice of prompt format and few-shot examples.

\textbf{Notably, we find that such differences in performance can result from seemingly insignificant differences in the prompt.} Below, we show the prompt formats optimized for \textsc{Clinical-Camel-70B} and \textsc{Llama-2-70B} on MMLU (Clinical Knowledge):

\begin{custombox}[frametitle={\textsc{Clinical-Camel-70B} vs.~\textsc{Llama-2-70B} on MMLU (Clinical Knowledge)}]
\textbf{Prompt Format Optimized for \textsc{Clinical-Camel-70B}:}\\
**Question**= Glycolysis is the name given to the pathway involving the conversion of:\\
(A) glycogen to glucose-1-phosphate.\\
(B) glycogen or glucose to fructose.\\
(C) glycogen or glucose to pyruvate or lactate.\\
(D) glycogen or glucose to pyruvate or acetyl CoA.\\
**Answer**=

\noindent \textbf{Prompt Format Optimized for \textsc{Llama-2-70B}:}\\
\#\#\# Question: Glycolysis is the name given to the pathway involving the conversion of:\\
(A) glycogen to glucose-1-phosphate.\\
(B) glycogen or glucose to fructose.\\
(C) glycogen or glucose to pyruvate or lactate.\\
(D) glycogen or glucose to pyruvate or acetyl CoA.\\
\#\#\# Answer: 
\end{custombox}
We find that the main difference in the prompt, modulo the selected few-shot examples, lies on how the question and answer headers are formatted, with the overall semantics of the overall prompt remaining completely intact. 
We also note that the task instruction provided to \textsc{Clinical-Camel-70B} and \textsc{Llama-2-70B} is also identical: ``The following is a multiple-choice question about medical knowledge. Answer the question by choosing one of the options from A to D.'' 
These results are consistent with the conclusions of prior works that demonstrate that the behavior of LLMs/VLMs are highly sensitive to the specifics of the input prompt \citep{how-can-we-know-what-lms-know,calibrate-before-use,quantifying-lm-prompt-design}, including works specifically focused on clinical tasks \citep{open-clinical-llms-sensitive}.

\textit{Our findings suggest that 
selecting the ``right'' prompt format and few-shot examples for each model separately is crucial for a fair comparison in the zero-/few-shot prompting regime.}

\subsubsection{Ignoring Statistical Uncertainty Can Lead to Misleading Conclusions About Performance Improvements from Medical DAPT}
\label{sec:prompting-results-2-uncertainty}

\begin{figure*}[t!]
    \centering
    \begin{tabular}{@{}c@{}c@{}}
        \begin{subfigure}{0.03\linewidth}
            \makebox[\linewidth]{\raisebox{55pt}{{(a)}}}
        \end{subfigure} &
        \begin{subfigure}{0.9\linewidth}
            \includegraphics[width=\linewidth]{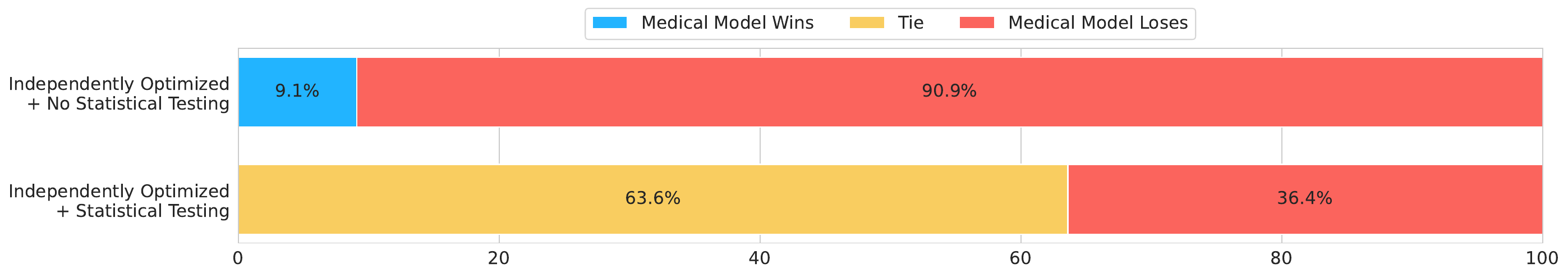}
        \end{subfigure}
        \\
        \begin{subfigure}{0\linewidth}
            \makebox[\linewidth]{\raisebox{55pt}{{(b)}}}
        \end{subfigure} &
        \begin{subfigure}{0.9\linewidth}
            \includegraphics[width=\linewidth]{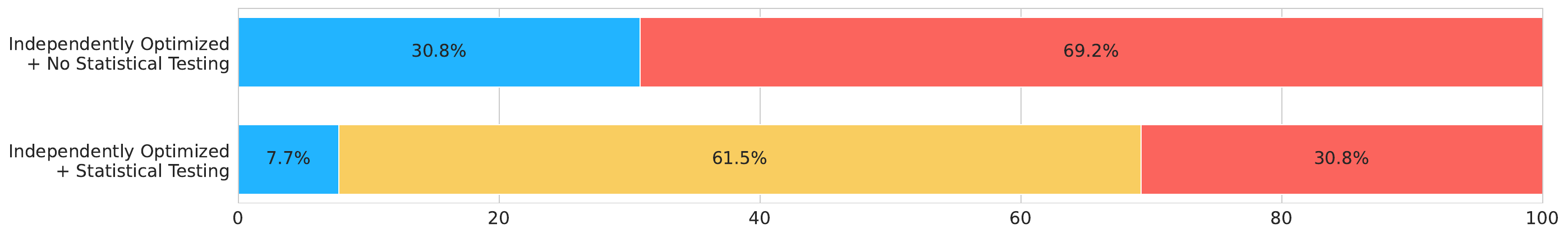}
        \end{subfigure} \\
        \begin{subfigure}{0\linewidth}
            \makebox[\linewidth]{\raisebox{65pt}{{(c)}}}
        \end{subfigure} &
        \begin{subfigure}{0.9\linewidth}
            \includegraphics[width=\linewidth]{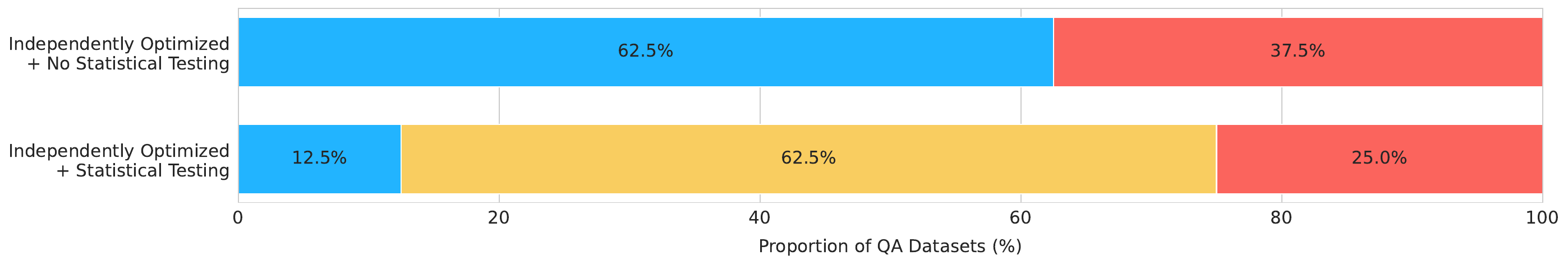}
        \end{subfigure} 
    \end{tabular}
    \vspace{-10pt}
    \caption{Ignoring statistical uncertainty in model comparison can lead to overly optimistic or pessimistic conclusions about the performance benefits from medical DAPT: (a) \textsc{Clinical-Camel-70B}, (b) \textsc{OpenBioLLM-8B}, (c) \textsc{LLaVA-Med-7B}. 
    Here, we show the results when we separately perform model-specific prompt selection for each model in a given pair (``Independently Optimized'').
    ``No Statistical Testing'' refers to the setting where we compare models based on raw absolute accuracy.
    ``Statistical Testing'' refers to the setting where we compare models based on the 95\% confidence intervals in relative accuracy.
    For \textsc{Clinical-Camel-70B} and \textsc{OpenBioLLM-8B}, we show the 3-shot win/tie/loss rates (\%) on the textual medical knowledge QA tasks (trends are similar for the textual clinical note QA tasks; see Section \ref{sec:prompting-results-2-aggregate}).
    For \textsc{LLaVA-Med-7B}, we show the 3-shot win/tie/loss rates (\%) on all visual medical QA tasks.
    \textit{Without statistical testing, both win and loss rates can be overestimated due to the failure to account for statistical ties.}
    }
    \label{fig:medopt-test}
\end{figure*}

In Figure \ref{fig:medopt-test}, we show how the win, tie, loss rates (\%) change when we remove statistical testing, comparing the performances of each model pair solely based on raw absolute accuracy. 
As in Section \ref{sec:prompting-results-2-prompt}, we show the 3-shot results for \textsc{Clinical-Camel-70B} and \textsc{OpenBioLLM-8B} on the textual medical knowledge QA datasets, and the 3-shot results for \textsc{LLaVA-Med-7B} on all visual medical QA datasets. We note that the results here are shown \textit{after} performing model-specific prompt selection for each model independently.

In all three cases, we overall observe that due to the absence of ties when comparing based on raw absolute accuracy, both the win and loss rates can become overestimated. 
In particular, we observe that the win rate for \textsc{LLaVA-Med-7B} in the 3-shot setting increases from 12.5\% to 62.5\%, reversing the conclusions about its improvement over its base model \textsc{LLaVA-v0-7B} (Figure \ref{fig:medopt-test}(c)).
For \textsc{Clinical-Camel-70B} and \textsc{OpenBioLLM-8B}, we find that while both models lose more than they win against their base models with or without statistical testing, the failure to account for statistical ties also leads to an overly pessimistic conclusion.
In fact, with statistical testing, both models reach a tie in 60+\% of QA datasets when we account for statistical uncertainty (Figures \ref{fig:medopt-test}(a) and (b)), which is almost entirely ignored when comparing based on raw accuracy. 
Accounting for statistical uncertainty allows one to more accurately conclude that all three models reach a statistical tie in most cases, with little to no statistically significant improvements in performance.

\textit{These findings suggest that in order to draw reliable conclusions about the performance benefits from medical DAPT, it is essential to take statistical uncertainty into consideration.}

\subsubsection{Ignoring Prompt Sensitivity and Statistical Uncertainty Can Overestimate the Performance Benefits from Medical DAPT}
\label{sec:prompting-results-2-aggregate}

\begin{figure*}[t!]
    \centering
    \includegraphics[width=\linewidth]{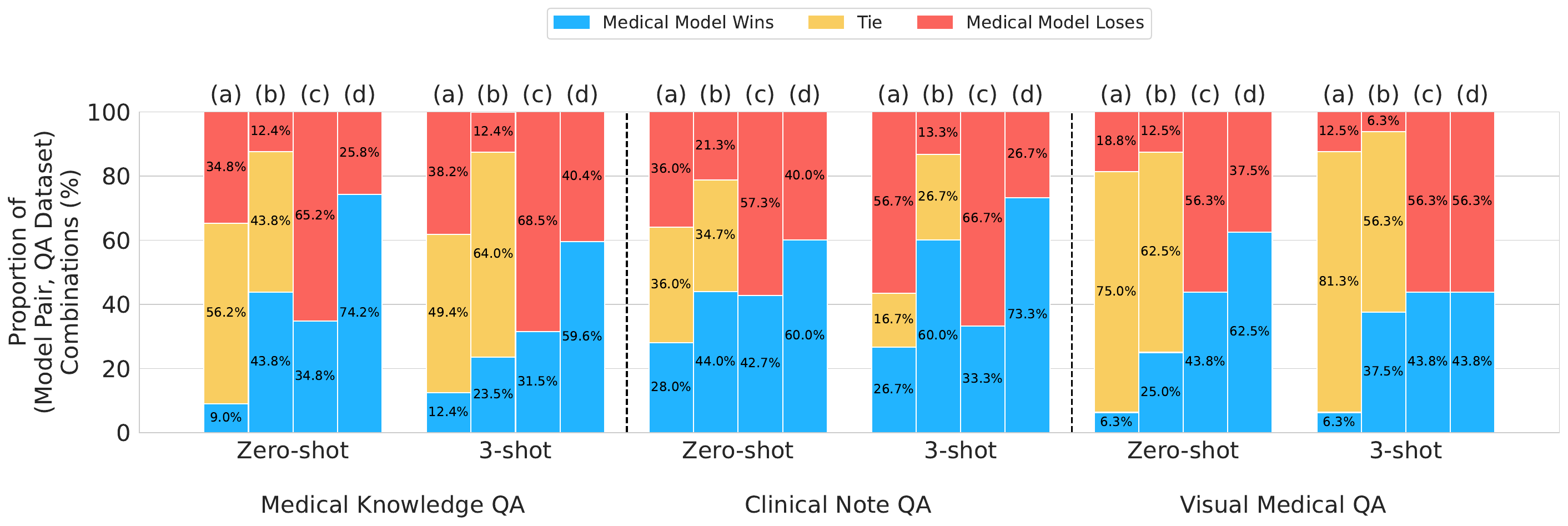}
    \vspace{-15pt}
    \caption{Optimizing the prompt for only the medical model and comparing models without accounting for statistical uncertainty can overestimate the performance improvements from medical DAPT. 
    We show the win/tie/loss rates (\%) of the medical models across all (model pair, QA dataset) combinations, when (a) optimizing the prompt for each model, with statistical testing; (b) optimizing the prompt only for the medical model, with statistical testing; (c) optimizing the prompt for each model, without statistical testing; and (d) optimizing the prompt only for the medical model, without statistical testing. 
    We show the results for greedy decoding. 
    The results for constrained decoding are similar (Figure \ref{fig:opt-logprob-ci-acc}).}
    \label{fig:opt-ci-acc}
\end{figure*}

Based on the findings discussed in Sections \ref{sec:prompting-results-2-prompt}--\ref{sec:prompting-results-2-uncertainty}, we investigate how the lack of (i) independent prompt selection and (ii) statistical testing
can change the broader conclusions about the effectiveness of medical DAPT in improving zero-/few-shot prompting performance.
In total, we consider four different scenarios, resulting from the inclusion or exclusion of these two aspects of the experimental setup. We aggregate all of the zero-shot and 3-shot results on textual medical knowledge QA, textual clinical note QA, and visual QA.

When we aggregate the results over all (model pair, QA dataset) combinations, we find that for both LLMs and VLMs, the performance improvement from medical DAPT 
can be substantially overestimated when (i) the prompt is only tailored to the medical model; and (ii) the models are compared only based on their absolute accuracies (Figure \ref{fig:opt-ci-acc}). 
For example, the aggregate zero-shot win rate substantially increases from 9.0\% to 74.2\% on textual medical knowledge QA, 28.0\% to 60.0\% on textual clinical note QA, and 6.3\% to 62.5\% on visual medical QA, when only performing prompt selection for the medical model and comparing based on raw absolute accuracy.
We see a similar trend in the win/tie/loss rates when the model predictions are generated via constrained decoding (Figure \ref{fig:opt-logprob-ci-acc}). 

\textit{In summary, these results highlight the importance of accounting for LLM/VLM sensitivity to the prompting details (Section \ref{sec:prompting-results-2-prompt}) and statistical uncertainty (Section \ref{sec:prompting-results-2-uncertainty}), in order to draw reliable conclusions about the effectiveness of medical DAPT (Section \ref{sec:prompting-results-2-aggregate}).}

\section{Supervised Fine-Tuning (SFT) Evaluation Results}\label{sec:sft-results}
We summarize the main findings from the SFT experiments outlined in Section \ref{sec:sft}. 
Overall, we find that after SFT, medical LLMs \textit{do} show statistically significant improvements on the textual medical knowledge QA datasets (Section \ref{sec:sft-results-knowledge}) but not on the clinical note QA datasets (Section \ref{sec:sft-results-clinical}), 
while medical VLMs do not show consistent improvements on the visual medical QA datasets (Section \ref{sec:sft-results-vqa}).

\begin{table}[t!]
    \centering
    \caption{The win, tie, and loss rates (\%) of all medical LLMs on textual medical knowledge QA (left) and textual clinical note QA (right), after fine-tuning them on the training set of each QA dataset. For each medical model, we boldface the win rate if it wins more than it loses to its general-domain base model, and vice versa. We exclude the results for the \textsc{Med42} models on the textual medical knowledge QA datasets, as they have already been trained on all of those datasets during medical DAPT.}
    \label{tab:win-tie-loss-rates-llm-sft}
    \resizebox{0.65\linewidth}{!}{
    
    \begin{tabular}{@{}l@{\hskip 20pt}c@{\hskip 10pt}c@{\hskip 10pt}c@{\hskip 15pt}c@{\hskip 10pt}c@{\hskip 10pt}c@{\hskip 10pt}}
        \toprule
        \multirow{2}{*}{\textbf{Model}} & \multicolumn{3}{c}{\textbf{\shortstack{Medical\\Knowledge}}} & \multicolumn{3}{c}{\textbf{\shortstack{Clinical\\Note}}} \\
        \cmidrule{2-4} \cmidrule{5-7}
        & Win & Tie & Loss & Win & Tie & Loss \\
        \midrule
        \textsc{Med42-v2-70B}             & - & - & - & \textbf{37.5} & 62.5 & 0 \\
        \textsc{OpenBioLLM-70B}        & \textbf{50.0} & 50.0 & 0 & \textbf{12.5} & 87.5 & 0 \\
        \textsc{Med42-v1-70B}             & - & - & - & 0 & 100.0 & 0 \\
        \textsc{MediTron-70B}             & \textbf{75.0} & 25.0 & 0 & 0 & 100.0 & 0 \\
        \textsc{Clinical-Camel-70B} & 0 & 100.0 & 0 & 0 & 100.0 & 0 \\
        \midrule
        \textsc{Med42-v2-8B}              & - & - & - & 12.5 & 75.0 & 12.5 \\
        \textsc{OpenBioLLM-8B}         & \textbf{25.0} & 75.0 & 0 & 0 & 100.0 & 0 \\
        \textsc{MediTron-7B}              & \textbf{75.0} & 25.0 & 0 & 0 & 100.0 & 0 \\
        \textsc{BioMistral-7B}          & \textbf{50.0} & 50.0 & 0 & 0 & 100.0 & 0 \\
        \textsc{BioMedGPT-LM-7B}         & \textbf{75.0} & 25.0 & 0 & 0 & 87.5 & \textbf{12.5} \\
        \midrule
        \textbf{Aggregate}                                 & \textbf{53.8} & 46.2 & 0 & \textbf{6.7} & 90.7 & 2.7 \\
        \bottomrule
    \end{tabular}
    }
\end{table}

\subsection{Evaluation of Medical LLMs on Textual Medical Knowledge QA}
\label{sec:sft-results-knowledge}

\begin{figure*}[t!]
    \centering
    \begin{tabular}{@{}c@{}c@{\hskip 7pt}c@{}c@{}}
        \multicolumn{4}{c}{
            \begin{subfigure}{0.97\linewidth}
                \includegraphics[width=\linewidth]{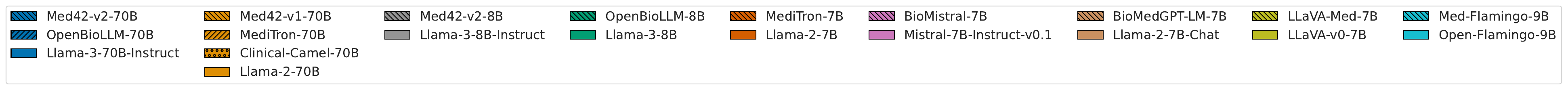}
            \end{subfigure}
        }
        \\
        \begin{subfigure}{0.03\linewidth}
            \makebox[\linewidth]{\raisebox{55pt}{{(a)}}}
        \end{subfigure} &
        \begin{subfigure}{0.45\linewidth}
            \includegraphics[width=\linewidth]{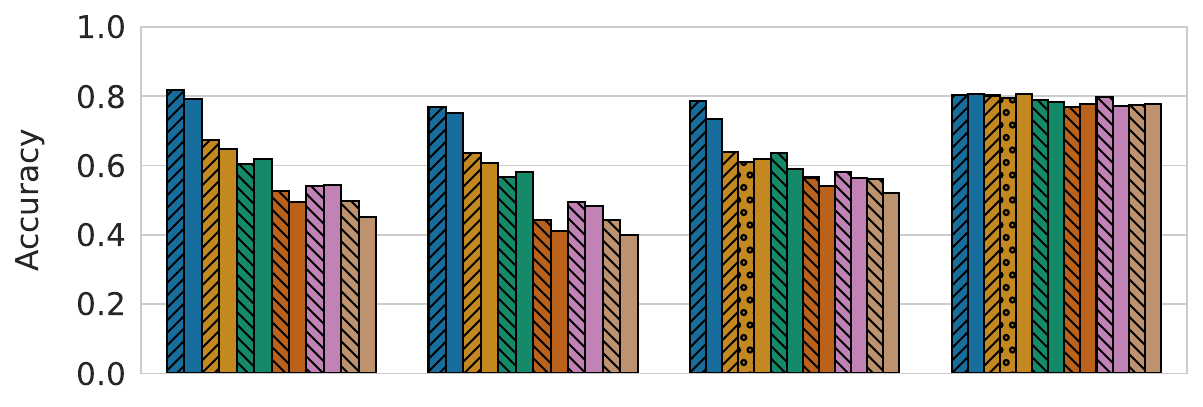}
        \end{subfigure} &
        \begin{subfigure}{0.03\linewidth}
            \makebox[\linewidth]{\raisebox{55pt}{{(b)}}}
        \end{subfigure} &
        \begin{subfigure}{0.45\linewidth}
            \includegraphics[width=\linewidth]{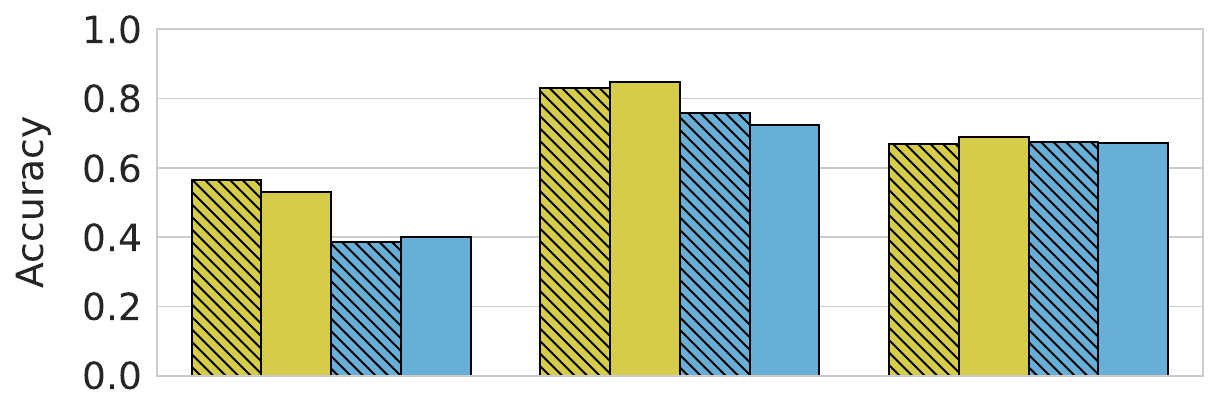}
        \end{subfigure}
        \\
        \begin{subfigure}{0.03\linewidth}
        \end{subfigure} &
        \begin{subfigure}{0.45\linewidth}
            \includegraphics[width=\linewidth]{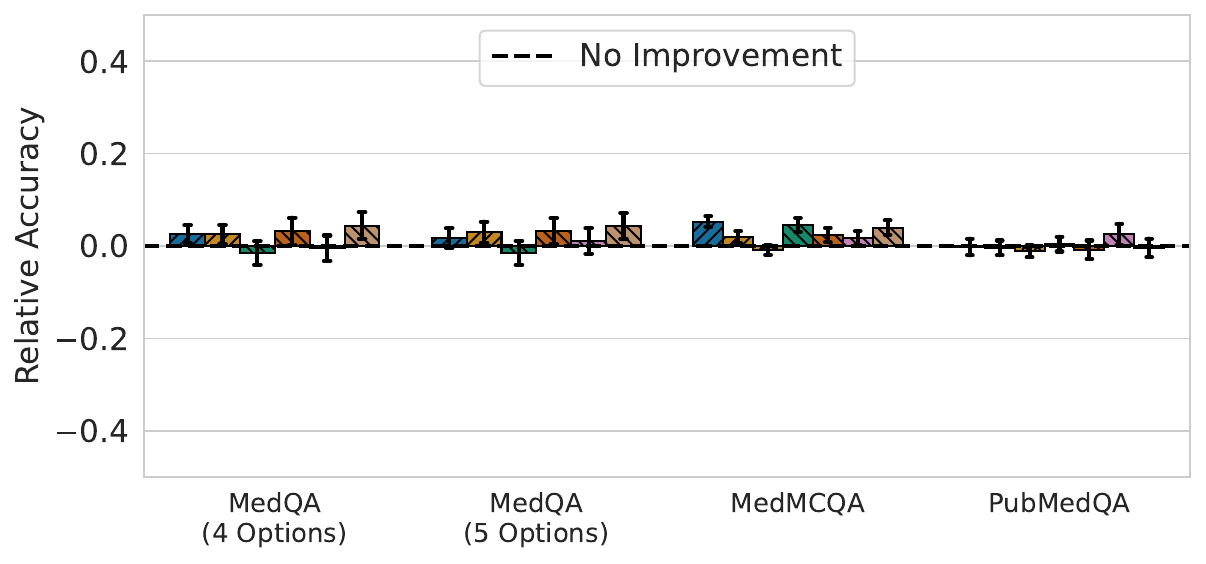}
        \end{subfigure} &
        \begin{subfigure}{0\linewidth}
        \end{subfigure} &
        \begin{subfigure}{0.45\linewidth}
            \includegraphics[width=\linewidth]{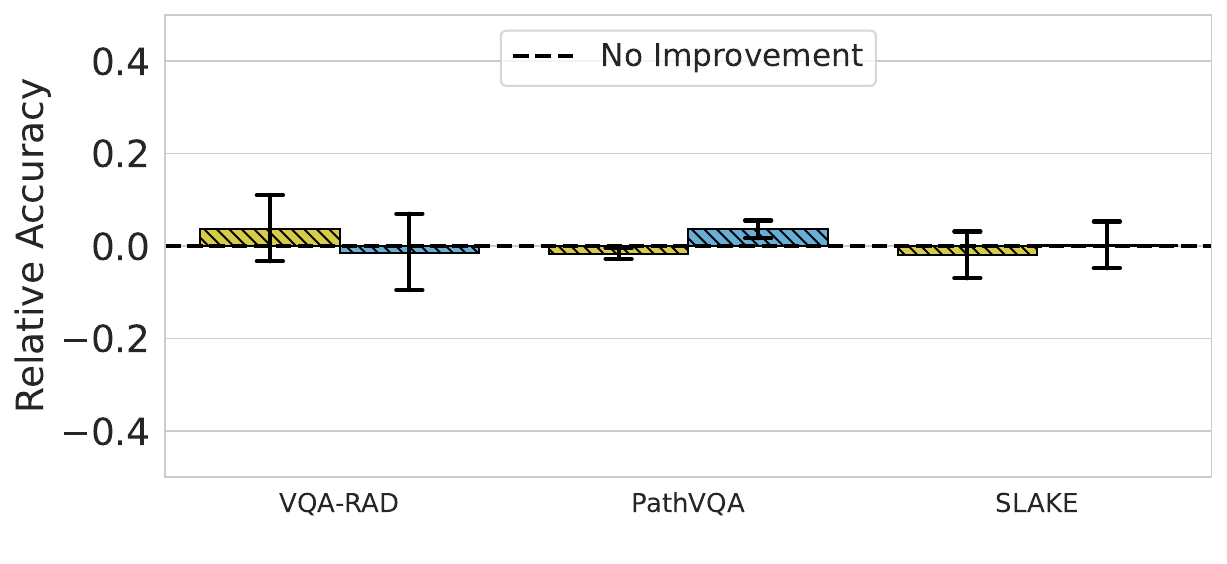}
        \end{subfigure}
        \\
        \begin{subfigure}{0.03\linewidth}
            \makebox[\linewidth]{\raisebox{55pt}{{(c)}}}
        \end{subfigure} &
        \multicolumn{3}{c}{
            \begin{subfigure}{0.95\linewidth}
                \includegraphics[width=\linewidth]{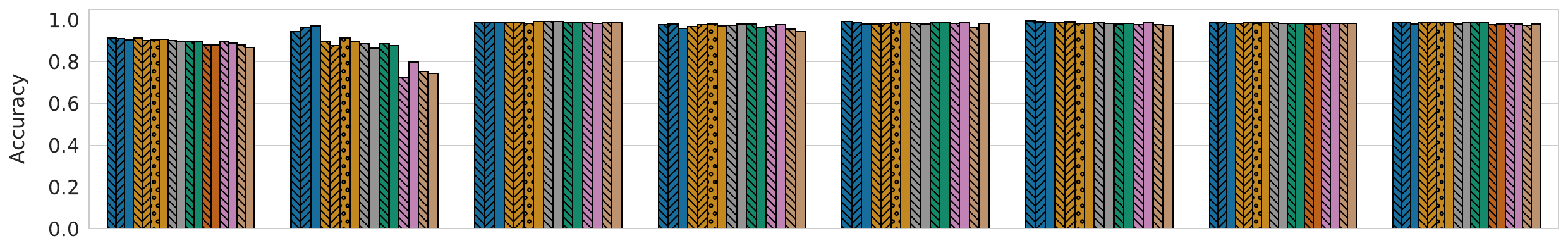}
            \end{subfigure}
        }
        \\
        \begin{subfigure}{0.03\linewidth}
        \end{subfigure} &
        \multicolumn{3}{c}{
            \begin{subfigure}{0.95\linewidth}
                \includegraphics[width=\linewidth]{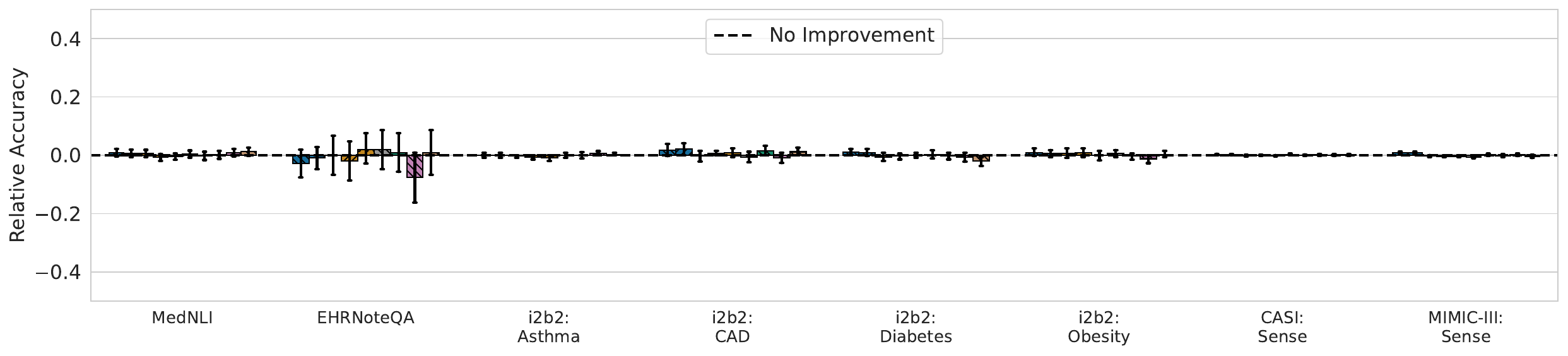}
            \end{subfigure}
        }
    \end{tabular}
    \vspace{-10pt}
    \caption{Absolute and relative accuracies of (a) all LLMs on textual medical knowledge QA, (b) all VLMs on visual medical QA, and (c) all LLMs on textual clinical note QA, after SFT on each task. \textit{Overall, medical LLMs do show improvement in SFT performance on the textual medical knowledge QA datasets, but not on the textual clinical note QA datasets. 
    Medical VLMs do not show significant improvements on the visual medical QA datasets.}
    }
    \label{fig:sft-acc-ci}
\end{figure*}

In Figure \ref{fig:sft-acc-ci}(a), we show the absolute and relative exact-match accuracies achieved by the medical and general-domain LLMs on the textual medical knowledge QA datasets, after fine-tuning each model.
In Table \ref{tab:win-tie-loss-rates-llm-sft} (left), we also show the win, tie, and loss rates (\%) of the medical LLMs on these datasets. 
As discussed in Section \ref{sec:sft}, we do not evaluate on the MMLU-Medical datasets, which only have 5 examples in the training set per subject (Table \ref{tab:datasets}).
We also exclude the results for the \textsc{Med42} models, since they have already been trained on all medical knowledge QA datasets via medical DAPT (Table \ref{tab:models}).

Overall, we find that \textbf{medical LLMs \textit{do} consistently improve over their general-domain base models in the SFT regime.}
All medical LLMs generally perform better than their base models in absolute terms, albeit with marginal improvements (on average, 1.6\% improvement in absolute accuracy) (top row of Figure \ref{fig:sft-acc-ci}(a)).
In most cases, these improvements are nonetheless statistically significant, with the 95\% confidence intervals in relative accuracy lying above zero (bottom row of Figure \ref{fig:sft-acc-ci}(a)).
In fact, all medical LLMs achieve loss rates of 0\%, with 5 out of 7 models achieving win rates of 50+\% (Table \ref{tab:win-tie-loss-rates-llm-sft}, left). 
Even \textsc{BioMedGPT-LM-7B} (\llamatwosevenchat{brown}), which significantly underperforms its base model \textsc{Llama-2-7B-Chat} in the zero-/few-shot prompting regime (Sections \ref{sec:prompting-results-1}--\ref{sec:prompting-results-2}), achieves a win rate of 75.0\%.
We also find that medical models based on more recent general-domain base models (e.g., \textsc{OpenBioLLM-70B} and \textsc{OpenBioLLM-8B}, based on \textsc{Llama-3}) tend to show lower win rates than the other medical models based on older general-domain base models (e.g., \textsc{MediTron} and \textsc{BioMedGPT-LM}, based on \textsc{Llama-2}).

\textit{Together with our findings from Sections \ref{sec:prompting-results-1}--\ref{sec:prompting-results-2}, these results suggest that additional training on biomedical text indeed provides a better initialization of the parameters for downstream fine-tuning on textual medical knowledge QA tasks, while the off-the-shelf capabilities of medically adapted LLMs are overall no better than those of their base models}.

\subsection{Evaluation of Medical LLMs on Textual Clinical Note QA}
\label{sec:sft-results-clinical}

\begin{figure*}[t!]
    \centering
    \begin{tabular}{@{}c@{}c@{}}
        \multicolumn{2}{c}{
            \begin{subfigure}{0.95\linewidth}
                \includegraphics[width=\linewidth]{figs/llm-acc-ci-legend-clinical.pdf}
            \end{subfigure}
        }
        \\
        \begin{subfigure}{0.45\linewidth}
            \includegraphics[width=\linewidth]{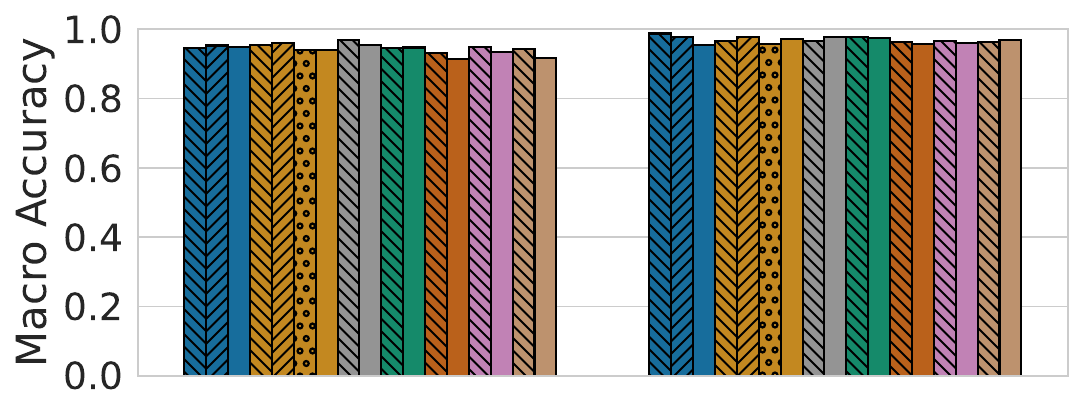}
        \end{subfigure} &
        \begin{subfigure}{0.45\linewidth}
            \includegraphics[width=\linewidth]{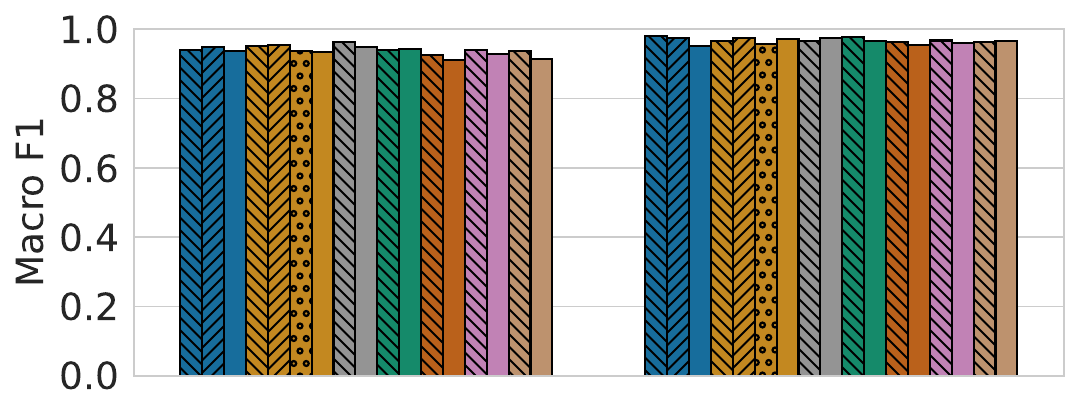}
        \end{subfigure}
        \\
        \begin{subfigure}{0.45\linewidth}
            \includegraphics[width=\linewidth]{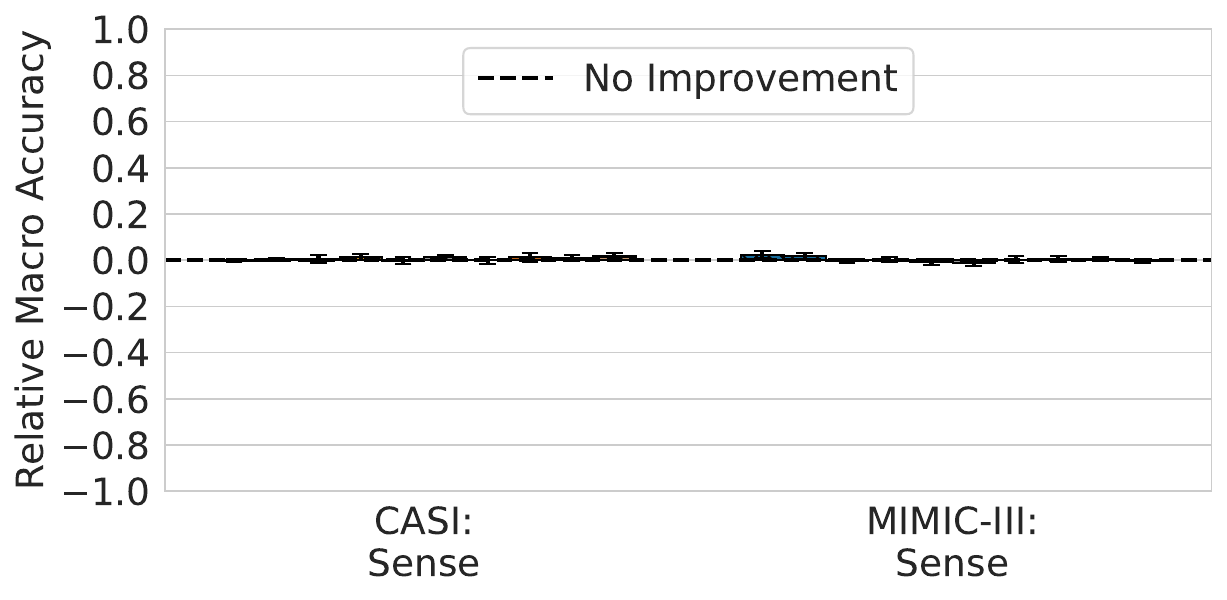}
        \end{subfigure} &
        \begin{subfigure}{0.45\linewidth}
            \includegraphics[width=\linewidth]{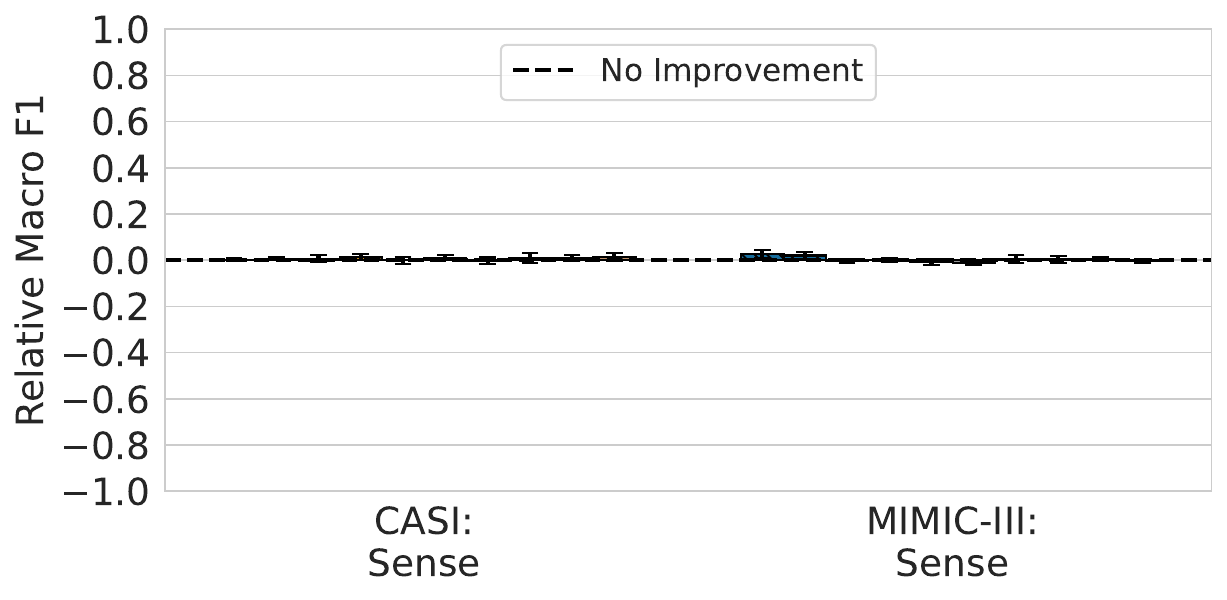}
        \end{subfigure}
    \end{tabular}
    \vspace{-10pt}
    \caption{Even after accounting for the imbalance in the distribution of clinical acronyms in the CASI and MIMIC-III sense disambiguation datasets, medical LLMs fail to consistently improve over their base models, after SFT. 
    In each panel, the top row shows the absolute macro exact-match accuracies and F1 scores---averaged over clinical acronyms---on the test set, and the bottom row shows the relative macro exact-match accuracies and F1 scores along with 95\% confidence intervals derived via bootstrapping on the test set (Section~\ref{sec:eval-setup}).}
    \label{fig:sft-acc-ci-sense}
\end{figure*}

In Figure \ref{fig:sft-acc-ci}(c), we show the absolute and relative exact-match accuracies achieved by the medical and general-domain LLMs on the textual clinical note QA datasets, after fine-tuning each model.
For the CASI and MIMIC-III sense disambiguation datasets, we also show the absolute and relative macro exact-match accuracies and F1 scores---averaged over clinical acronyms---in Figure \ref{fig:sft-acc-ci-sense} to account for the imbalance in the acronym distribution (Section \ref{sec:prompting-results-1-clinical}).
In Table \ref{tab:win-tie-loss-rates-llm-sft} (right), we also show the win, tie, and loss rates (\%) of the medical LLMs on these datasets. 
As in the zero-/few-shot prompting evaluations discussed in Section \ref{sec:prompting-results-1-clinical}, we exclude the results for \textsc{MediTron-7B} (and its base model \textsc{Llama-2-7B}) on the EHRNoteQA and i2b2 datasets, as its small context window size of 2k tokens is insufficient for handling the full-length clinical notes in these datasets.

Across all model pairs and datasets, we find that the \textbf{fine-tuning performances of the medical LLMs and those of their base models are nearly indistinguishable} (Figure \ref{fig:sft-acc-ci}(c)).
In Figure \ref{fig:sft-acc-ci-sense}, we make similar observations on the CASI and MIMIC-III sense disambiguation datasets even after accounting for the imbalance in the clinical acronym distribution.
In Table \ref{tab:win-tie-loss-rates-llm-sft} (right), we find that across all (model pair, textual clinical note QA dataset) combinations, 90.7\% of cases result in a statistical tie, with the medical models winning against their base models in only 6.7\% of cases. 
In fact, \textbf{7 out of 10 medical LLMs are indistinguishable from their base models in terms of the win/tie/loss rates}. 
Meanwhile, we observe that while the LLMs do not consistently perform well \textit{off-the-shelf} on clinical note QA (Section \ref{sec:prompting-results-1-clinical}), both medical and general-domain models uniformly show strong performance after being fine-tuned for each task, across all model scales.

\textit{These results suggest that additional training on \textit{biomedical} text data, as commonly done for the medical LLMs that we evaluate (Table \ref{tab:models}), is not particularly beneficial for improving downstream supervised fine-tuning performance on QA tasks involving \textit{clinical notes}.}

\begin{table}[t!]
    \centering
    \caption{The win, tie, and loss rates (\%) of all medical VLMs on visual medical QA, after fine-tuning them on the training set of each QA dataset. For each medical model, we boldface the win rate if it wins more than it loses to its general-domain base model, and vice versa.}
    \label{tab:win-tie-loss-rates-vqa-sft}
    \resizebox{0.5\linewidth}{!}{
    \begin{tabular}{@{}l@{\hskip 20pt}c@{\hskip 25pt}c@{\hskip 10pt}c@{}}
        \toprule
        \multirow{2}{*}{\textbf{Model}} & \multicolumn{3}{c}{\textbf{Visual Medical QA}} \\
        \cmidrule{2-4}
        & Win & Tie & Loss \\
        \midrule
        \textsc{LLaVA-Med-7B}    & 0 & 66.7 & 33.3 \\
        \textsc{Med-Flamingo-9B} & \textbf{33.3} & 66.7 & 0 \\
        \midrule
        \textbf{Aggregate}                            & 16.7 & 66.7 & 16.7 \\
        \bottomrule
    \end{tabular}
    }
\end{table}

\subsection{Evaluation of Medical VLMs on Visual Medical QA}
\label{sec:sft-results-vqa}

In Figure \ref{fig:sft-acc-ci}(b), we show the absolute and relative exact-match accuracies achieved by the medical and general-domain VLMs on the visual medical QA datasets, after fine-tuning each model. 
In Table \ref{tab:win-tie-loss-rates-vqa-sft}, we also show the win, tie, and loss rates (\%) of the medical VLMs on these datasets. 
As discussed in Section \ref{sec:sft}, we do not evaluate on the MMMU-Medical datasets, which only have 5 examples in the training set per subject (Table \ref{tab:datasets}).

Overall, \textbf{medical VLMs do not consistently improve over their general-domain base models in the SFT regime}.
We observe that \textsc{LLaVA-Med-7B} (\llava{light green}) shows no statistically significant improvements over its base model, while \textsc{Med-Flamingo-9B} (\flamingo{cyan}) shows improvement on PathVQA (+3.5\% in raw accuracy) but reaches a tie in others (Figure \ref{fig:sft-acc-ci}(b)). 
When we aggregate the results across all (model pair, visual medical QA dataset) combinations, the medical VLMs achieve win/tie/loss rates of 16.7\%/66.7\%/16.7\% (Table \ref{tab:win-tie-loss-rates-vqa-sft}), indicating limited collective improvement.
Meanwhile, given that these conclusions are based on a relatively small number of models (only 2) and datasets (only 3), we note that the generalizability of this finding may be limited.

\textit{In summary, the results in Sections \ref{sec:sft-results-knowledge}--\ref{sec:sft-results-vqa} show that in the SFT regime, medical LLMs show improvements on textual QA tasks focused on assessing medical knowledge but not on those based on clinical notes, while medical VLMs generally fail to show statistically significant improvements on the visual medical QA datasets.}

\section{Discussion and Conclusion}\label{sec:discussion}
In this work, we investigated the effectiveness of DAPT 
for training medically specialized LLMs and autoregressive VLMs 
suitable for various medical (visual) QA tasks.
To that end, we compared several pairs of state-of-the-art medical LLMs/VLMs 
to their general-domain counterparts,
whose only differences lie in medical DAPT 
and are exactly identical in model architecture and scale. 
Our work differs from prior works 
by providing direct apples-to-apples comparisons 
of medical and general-domain base models, 
while accounting for LLM/VLM sensitivity to prompting details
and assessing the statistical significance of the results.
We also ensure that our evaluations are comprehensive 
by (i) comparing the models in both the zero-/few-shot prompting and SFT regimes;
and (ii) covering both QA tasks focused on assessing medical knowledge and those based on real-world clinical notes.

In the zero-/few-shot prompting regime, 
we found that across both model classes and all model scales, the performance benefits from medical DAPT 
largely disappear when we 
(i) tailor the prompt format and choice of few-shot examples 
to each medical and general-domain model separately;
and (ii) account for statistical uncertainty in model comparison (Section \ref{sec:prompting-results-1}).
In particular, we found that when we optimize the prompt only for the medical model 
and compare each model pair based on their absolute accuracies without accounting for uncertainty, 
the performance improvements from medical DAPT can be significantly overestimated (Section \ref{sec:prompting-results-2}), 
potentially leading to unreliable conclusions about the benefits of medical DAPT on zero-/few-shot prompting performance.

In the SFT regime, we found that medical LLMs overall do show statistically significant improvements on textual QA tasks focused on assessing medical knowledge (Section \ref{sec:sft-results-knowledge}) but not on clinically relevant QA tasks that based on real-world clinical notes (Section \ref{sec:sft-results-clinical}).
For medical VLMs, we found that they overall fail to show significant improvements on the visual medical QA tasks that we consider, which we present with the caveat that our findings are based on a relatively small number of VLMs and datasets (Section \ref{sec:sft-results-vqa}).

Our findings suggest that for state-of-the-art general-domain LLMs and VLMs, 
the performance benefits from additionally pretraining 
on medical data from public sources such as PubMed may be limited. 
In fact, almost all of the medical models used in our evaluation 
use PubMed as the primary source of pretraining data 
for medical adaptation (Table~\ref{tab:models}), 
while open-source datasets commonly used 
for pretraining the general-domain base models in the first place 
(e.g., the Pile \citep{the-pile}, S2ORC \citep{s2orc}) 
often already include PubMed data. 
Thus, we emphasize that any claims of improvement 
due to proposed medical DAPT procedures 
should be backed up by rigorous head-to-head comparisons 
against the corresponding general-domain base models. 

\paragraph{Limitations.}
First, there is a vast and growing set of papers on applying medical DAPT to various general-domain base models, and we could not hope to compare all publicly available models here. 
While we selected the models to cover a wide range of general-domain base models and model scales (7B--70B) (Table~\ref{tab:models}) and included some of the latest models (e.g., \textsc{OpenBioLLM} and \textsc{Llama-3}), 
it is always possible that some newly released models do in fact yield better zero-/few-shot or SFT performance on medical QA.

Second, we focus in this paper on the narrower task of closed-ended medical QA.  In part, this choice reflects the fact that such benchmarks are well-standardized and highly publicized. However, they do not reflect the breadth of possible applications of LLMs and VLMs in medical domains. For instance, \citet{med-palm-2} show that medical LLMs such as \textsc{Med-PaLM-2} can produce physician-level answers to open-ended consumer health queries; 
\citet{liu2022retrieve} explore the use of LLMs for generating discharge instructions based on a patient's health records; and
\citet{li2023chatdoctor} demonstrate the potential of using LLMs to recommend treatments for patients based on patient-doctor conversations. 
Some would argue that such tasks are a more realistic application of such models in practice, and it is certainly possible that an analysis like ours would find improved performance on such tasks, though we do not investigate these tasks in the present work.

Third, due to computational constraints, we do not consider full fine-tuning of models for evaluations in the SFT regime.
Prior works show that parameter-efficient fine-tuning performance (e.g., LoRA) can be worse than that of full fine-tuning, and that the gap in performance varies significantly across different tasks \citep{med42-v1,lora-learns-less-forgets-less}.
While our head-to-head comparisons based on parameter-efficient fine-tuning revealed that medical models show limited improvement over their base models across all model scales, the overall conclusions may be different in the full fine-tuning regime.

While we acknowledge the limitations above, we do not believe they detract from the value of this work.  We hope that our results call attention to a need for rigorous head-to-head evaluations when making similar claims of improved performance via medical DAPT, whether with other models, on other clinical tasks, or with respect to SFT versus zero-/few-shot prompting performance.

\acks{We gratefully acknowledge DARPA (FA8750-23-2-1015), ONR (N00014-23-1-2368), NSF (IIS2211955), UPMC, Highmark Health, Abridge, Ford Research, Mozilla, the PwC Center, Amazon AI, JP Morgan Chase, the Block Center, the Center for Machine Learning and Health, and the CMU Software Engineering Institute (SEI) via Department of Defense contract FA8702-15-D-0002, for their generous support of our research. We also thank Monica Agrawal, Ahmed Alaa, and Shan Chen for helpful discussions on an earlier version of this manuscript.
}

\newpage
\appendix

\renewcommand\thetable{A\arabic{table}}
\renewcommand\thefigure{A\arabic{figure}}
\setcounter{table}{0}
\setcounter{figure}{0}
\section{Additional Details on Datasets}\label{sec:details-datasets}
\begin{table}[h!]
    \centering
    \caption{The numbers of train, validation, and test QA examples for all textual medical knowledge QA, textual clinical note QA, and visual medical QA datasets used for evaluation.}
    \label{tab:datasets}
    \resizebox{0.75\linewidth}{!}{
    
    \begin{tabular}{@{}c@{\hskip 10pt}l@{\hskip 10pt}c@{\hskip 7pt}c@{\hskip 7pt}c@{\hskip 3pt}}
        \toprule
        \textbf{QA Type} & \textbf{Dataset} & \textbf{Train} & \textbf{Validation} & \textbf{Test} \\
        \midrule
        \multirow{12}{*}{\textbf{Textual Medical Knowledge QA}} & MedQA (4 \& 5 Options) & 10178 & 1272 & 1273 \\
        & MedMCQA & 146257 & 36565 & 4183 \\
        & PubMedQA & 211269 & 500 & 500 \\
        & MMLU (Anatomy) & 5 & 14 & 135 \\
        & MMLU (Clinical Knowledge) & 5 & 29 & 265 \\
        & MMLU (College Biology) & 5 & 16 & 144 \\
        & MMLU (College Medicine) & 5 & 22 & 173 \\
        & MMLU (High School Biology) & 5 & 32 & 310 \\
        & MMLU (Medical Genetics) & 5 & 11 & 100 \\
        & MMLU (Nutrition) & 5 & 33 & 306 \\
        & MMLU (Professional Medicine) & 5 & 31 & 272 \\
        & MMLU (Virology) & 5 & 18 & 166 \\
        \midrule
        \multirow{8}{*}{\textbf{Textual Clinical Note QA}} & MedNLI & 11232 & 1395 & 1422 \\
        & EHRNoteQA & 319 & 105 & 105 \\
        & i2b2 2008 Challenge (Asthma) & 457 & 115 & 357 \\
        & i2b2 2008 Challenge (CAD) & 441 & 111 & 335 \\
        & i2b2 2008 Challenge (Diabetes) & 457 & 115 & 358 \\
        & i2b2 2008 Challenge (Obesity) & 443 & 111 & 334 \\
        & CASI Sense Disambiguation & 9978 & 3304 & 3304 \\
        & MIMIC-III Sense Disambiguation & 10384 & 3458 & 3458 \\
        \midrule
        \multirow{8}{*}{\textbf{Visual Medical QA}} & VQA-RAD & 820 & 205 & 272 \\
        & PathVQA & 9806 & 3135 & 3391 \\
        & SLAKE & 1943 & 422 & 415 \\
        & MMMU (Basic Medical Science) & 5 & 5 & 25 \\
        & MMMU (Clinical Medicine) & 5 & 5 & 25 \\
        & MMMU (Diag. \& Lab Medicine) & 5 & 5 & 25 \\
        & MMMU (Pharmacy) & 5 & 5 & 25 \\
        & MMMU (Public Health) & 5 & 5 & 25 \\
        \bottomrule
    \end{tabular}
    }
\end{table}

\noindent For VQA-RAD \citep{vqa-rad}, PathVQA \citep{pvqa}, and SLAKE \citep{slake}, we only show the number of \textit{closed-ended} examples, since we focus on closed-ended QA. 
For the datasets that required additional splits from the official train-validation-test split (e.g., due to the lack of a public test set), we include all of the fixed random seeds.

\renewcommand\thetable{B\arabic{table}}
\renewcommand\thefigure{B\arabic{figure}}
\setcounter{table}{0}
\setcounter{figure}{0}
\section{Additional Details on Model-Specific Prompt Selection}\label{sec:details-prompting-selection}
To define the prompt format search space discussed in Section \ref{sec:prompting}, we construct a context-free grammar of plausible prompt formats following \citet{quantifying-lm-prompt-design} (see Section 3.1 and Appendix A of the paper for reference). Using the Backus-Naur notation, we first define the basic fields $H_q$ for the question header (e.g., \texttt{``\#\#\# Question:''}), $H_c$ for the answer choice header ('e.g., \texttt{``\#\#\# Options:''}), and $H_a$ for the answer header (e.g., \texttt{``\#\#\# Answer:''}) as
\begin{align}
    H_q(f_{\text{case}}, d_q, s_1) &::= f_{\text{case}}(d_q) s_1 \langle\text{text}\rangle, \nonumber \\
    H_c(f_{\text{case}}, d_c, s_1) &::= f_{\text{case}}(d_c) s_1, \nonumber \\
    H_a(f_{\text{case}}, d_a, s_1) &::= f_{\text{case}}(d_a) s_1 \langle\text{text}\rangle, \nonumber
\end{align}
where $f_\text{case} \in \mathcal{F}_{\text{case}}$ denotes the casing function (e.g., \texttt{x} $\mapsto$ ``\#\#\# '' + \texttt{x}, \texttt{x} $\mapsto$ \texttt{x.upper()}), $d_q \in D_q$ denotes the question descriptor (e.g., \texttt{``Question''}), $d_c \in D_c$ denotes the answer choice descriptor (e.g., \texttt{``Options''}), $d_a \in D_a$ denotes the answer descriptor (e.g., \texttt{``Answer''}), $s_1 \in S_1$ denotes the header separator (e.g., \texttt{`:'}), and $\langle\text{text}\rangle$ denotes a text placeholder. 
For formatting the list of answer choices, we also define the basic fields $C$ for formatting each answer choice (e.g., \texttt{``(A) yes''}) and $L$ for the concatenation of all answer choices as follows:
\begin{align}
    C(f_{\text{wrap}},f_{\text{index}}, i) &::= f_{\text{wrap}}(f_{\text{index}}(i))\langle\text{text}\rangle, \nonumber \\
    L(f_{\text{wrap}},f_{\text{index}},n,s_2) &::= C(f_{\text{wrap}},f_{\text{index}},0) s_2 \ldots s_2 C(f_{\text{wrap}},f_{\text{index}},n-1), \nonumber
\end{align}
where $f_{\text{wrap}} \in \mathcal{F}_{\text{wrap}}$ denotes the wrapper function for the answer choice letter (e.g., \texttt{x} $\mapsto$ \texttt{``('' + x + ``)''}), $f_{\text{index}} \in \mathcal{F}_{\text{index}}$ denotes the numbering function that converts an integer index into a number format (e.g., 0 $\rightarrow$ ``A''), $i \in \mathbb{Z}^+$ denotes the index of a particular answer choice from the list, $s_2 \in S_2$ denotes the answer choice separator, $n$ denotes the number of answer choices, and $\langle$text$\rangle$ denotes a text placeholder. The full prompt format $P(f_{\text{case}},f_{\text{wrap}},f_{\text{index}},d_q,d_c,d_a,s_1,s_2,n)$ is then constructed by concatenating all of the headers and the answer choices, while adding space $t \in T$ (e.g., \texttt{``\textbackslash n''}) in-between:
\begin{align}\label{eq:prompt-format}
    P &::= H_q t H_c t L t H_a,
\end{align}
where we exclude the arguments for simplicity. 
To define the prompt format search space, we instantiate the grammar with the descriptors, separators, spaces, and functions shown below.\\

\noindent \textbf{Descriptors:}
\begin{align}
    D_q &= \{\texttt{``Question''}, \texttt{``''}\}; \nonumber \\
    D_c &= \{\texttt{``Options''}, \texttt{``Choices''}, \texttt{``''}\}; \nonumber \\
    D_a &= \{\texttt{``Answer''}, \texttt{``The answer is''}\}. \nonumber
\end{align}

\noindent \textbf{Separators:}
\begin{align}
    S_1 &= \{\texttt{``: '', `` : '', `` :: '', ``:\textbackslash n'', ``= '', `` = '', `` == '', ``=\textbackslash n'', `` - '',} \nonumber \\
    &\quad\;\;\; \texttt{`` \text{-}- '', ``\text{-}-\text{-}'', ``\textbackslash n'', ``\textbackslash n\textbackslash n''}\}; \nonumber \\
    S_2 &= \{\texttt{``\textbackslash n'', ``\textbackslash \textbackslash'', ``; '', `` || '', `` '', ``;\textbackslash n'', ``;\textbackslash n\textbackslash n'', ``, ''}\}. \nonumber
\end{align}

\noindent \textbf{Spaces:}
\begin{align}
    T &= \{\texttt{``\textbackslash n'', ``\textbackslash \textbackslash'', `` || '', `` ''}\}. \nonumber
\end{align}

\noindent \textbf{Casing, Wrapper, and Numbering Functions:}
\begin{align}
    \mathcal{F}_{\text{case}} &= \{\texttt{x} \mapsto \texttt{x}, \texttt{x} \mapsto \texttt{x.title()}, \texttt{x} \mapsto \texttt{x.upper()}, \texttt{x} \mapsto \texttt{x.lower()},  \texttt{x} \mapsto \texttt{``\#\#\# '' + x}, \nonumber \\
    &\quad\;\;\; \texttt{x} \mapsto \texttt{``**'' + x + ``**''} \}; \nonumber \\
    \mathcal{F}_{\text{wrap}} &= \{\texttt{x} \mapsto \texttt{``('' + x + ``)''}, \texttt{x} \mapsto \texttt{ x + ``.''}, \texttt{x} \mapsto \texttt{ x + ``)''}, \texttt{x} \mapsto \texttt{``['' + x + ``]''}, \nonumber \\
    &\quad\;\;\; \texttt{x} \mapsto \texttt{ x + `` )''}, \texttt{x} \mapsto \texttt{``<'' + x + ``>''}\}; \nonumber \\
    \mathcal{F}_{\text{index}} &= \{\texttt{x} \mapsto \texttt{chr(ord(``A'') + x)} \}. \nonumber
\end{align}
To randomly sample a prompt format accepted by the grammar, we randomly sample each of these components and construct the full prompt format as in Eq.~\eqref{eq:prompt-format}. Below, we show an example QA pair from the MedQA dataset, formatted according to the formats sampled from the prompt format space defined by the above context-free grammar.

\begin{custombox}[frametitle={Example 1}]
A key factor facilitating the application of nested case-control studies from the MACS was:

\noindent OPTIONS -- A ) Data collection

\noindent B ) Establishment of a repository of biologic specimens

\noindent C ) Participant interest

\noindent D ) Administration of the questionnaire by staff

\noindent THE ANSWER IS -- B ) Establishment of a repository of biologic specimens
\end{custombox}

\begin{custombox}[frametitle={Example 2}]
QUESTION -- A key factor facilitating the application of nested case-control studies from the MACS was:

\noindent CHOICES -- [A] Data collection; [B] Establishment of a repository of biologic specimens; [C] Participant interest; [D] Administration of the questionnaire by staff

\noindent ANSWER -- [B] Establishment of a repository of biologic specimens
\end{custombox}

\renewcommand\thetable{C\arabic{table}}
\renewcommand\thefigure{C\arabic{figure}}
\setcounter{table}{0}
\setcounter{figure}{0}
\section{Additional Details on Zero-/Few-shot Prompting}\label{sec:details-prompting}
Here, we summarize the prompting details made available for the medical LLMs and VLMs used in our evaluation (Appendix \ref{sec:reproducibility}), and the default prompt formats used for each LLM (Appendix \ref{sec:llm-prompt-templates}) and VLM (Appendix \ref{sec:vlm-prompt-templates}), which have been reproduced based on the former.

\subsection{Reproducibility of Prompting Details}\label{sec:reproducibility}
In Table \ref{tab:prompting-details}, we provide a summary of all of the prompting details available (in the context of closed-ended medical QA) for all medical LLMs and VLMs used in our evaluation. We share these details to demonstrate our best efforts with reproducing the original prompting setups considered for performing our evaluations. In particular, we focus on whether the following four components are explicitly made available, either in the original publications or the publicly released code repository: (i) system prompt; (ii) zero-/few-shot prompt format (used for closed-ended QA tasks); (iii) the choice of few-shot examples; and (iv) details on how the text generations are sampled (e.g., softmax temperature, top-$p$, beam size, random seeds used for sampling).
Below, we provide detailed clarifications for each model. 

\begin{table*}[t!]
    \centering
    \caption{Summary of prompting details shared for each medical LLM and VLM in the original papers. For each column, a {\cmark} indicates that the information was fully provided, and a {\tri} indicates that the information was partially provided (e.g., random sampling without information about the seeds).
    Otherwise, the information was either not provided or irrelevant (e.g., no few-shot example details due to lack of few-shot evaluations).
    }
    \label{tab:prompting-details}
    \resizebox{0.9\linewidth}{!}{
    \begin{tabular}{@{}l@{\hskip 7pt}c@{\hskip 7pt}c@{\hskip 7pt}c@{\hskip 7pt}c@{\hskip 7pt}c@{}}
        \toprule
        Model & \shortstack{System Prompt} & \shortstack{Prompt Format} & \shortstack{Few-Shot Examples} & \shortstack{Sampling Details} \\
        \midrule
        \textsc{Med42-v2} & \cmark & \tri & & \cmark \\
        \textsc{Med42-v1} & \cmark & \cmark & & \cmark \\
        \textsc{OpenBioLLM} & \cmark & \tri & & \\
        \textsc{Clinical-Camel} & & \tri & & \tri \\
        \textsc{BioMistral} & \cmark & \cmark & & \tri \\
        \textsc{MediTron} & \cmark & \tri & \cmark & \cmark \\
        \textsc{BioMedGPT-LM} & & & & \\
        \midrule
        \textsc{LLaVA-Med} & \tri & \tri & & \tri \\
        \textsc{Med-Flamingo} & \cmark & \tri & & \\
        \bottomrule
    \end{tabular}
    }
\end{table*}

\paragraph{\textsc{Med42-v2}~\citep{med42-v2}.} We follow the instructions provided in the model cards on HuggingFace, for the 70B-parameter\footnote{\href{https://huggingface.co/m42-health/Llama3-Med42-70B}{huggingface.co/m42-health/Llama3-Med42-70B}} and 8B-parameter\footnote{\href{https://huggingface.co/m42-health/Llama3-Med42-8B}{huggingface.co/m42-health/Llama3-Med42-8B}} models. We use the recommended system prompt and the \textsc{Llama-3}-based conversational prompt format. 
Meanwhile, in Table \ref{tab:prompting-details}, we treat the prompt format as partially missing, as the exact format that was used to format each question (``user'' query) and answer (``assistant'' response) for evaluation on closed-ended multiple-choice questions is not provided.
As \citet{med42-v1} include the prompt format used for \textsc{Med42-v1}, we use this format since both models are from the same authors.
In the original evaluation, the model predictions are selected based on the log-likelihood of each answer choice, which differs from our evaluation setup.
We include the default prompt format used for \textsc{Med42-v2} in Appendix \ref{sec:med42-v2-template}.

\paragraph{\textsc{Med42-v1}~\citep{med42-v1}.} We follow the instructions provided in the model card\footnote{\href{https://huggingface.co/m42-health/med42-70b}{huggingface.co/m42-health/med42-70b}} on HuggingFace.
We use the recommended system prompt and the custom conversational prompt format with special \texttt{<|system|>}, \texttt{<|prompter|>}, and \texttt{<|assistant|>} tokens.
For the prompt format, we refer to what is provided in Appendix A.1 of \citet{med42-v1}. 
In the original evaluation, the model predictions are selected based on the log-likelihood of each answer choice, which differs from our evaluation setup (Section \ref{sec:eval-setup}).
We include the default prompt format used for \textsc{Med42-v2} in Appendix \ref{sec:med42-v1-template}.

\paragraph{\textsc{OpenBioLLM}~\citep{OpenBioLLMs}.} We follow the instructions provided in the model cards on HuggingFace, for the 70B-parameter\footnote{\href{https://huggingface.co/aaditya/Llama3-OpenBioLLM-70B}{huggingface.co/aaditya/Llama3-OpenBioLLM-70B}} and 8B-parameter\footnote{\href{https://huggingface.co/aaditya/Llama3-OpenBioLLM-8B}{huggingface.co/aaditya/Llama3-OpenBioLLM-8B}} models. We use the recommended system prompt and the \textsc{Llama-3}-based conversational prompt format. In Table~\ref{tab:prompting-details}, we treat the prompt format as partially missing, as the exact format used to format each question (``user'' query) and answer (``assistant'' response) for closed-ended multiple-choice QA is not provided. At the time of writing, there are no additional details about the models that have been publicly released, beyond what is provided in the model cards. We include the default prompt format used for \textsc{OpenBioLLM} in Appendix~\ref{sec:openbiollm-template}.

\paragraph{\textsc{Clinical-Camel}~\citep{clinical-camel}.} We use the conversation format used in the official GitHub repository, which corresponds to that of \textsc{Llama-2}~\citep{llama-2}. 
As the system prompts and few-shot examples used for the main evaluations in the paper are not provided, we use our own manually designed default system prompt and search over different choices of few-shot examples.
For sampling, the evaluation code\footnote{\href{https://github.com/bowang-lab/clinical-camel/blob/14d960f/evaluation/get_model_answer.py\#L54}{github.com/bowang-lab/clinical-camel/blob/14d960f/evaluation/get\_model\_answer.py\#L54}} uses default temperature setting of 0.7 (albeit without the random seeds), which differs from our evaluation setup.
We include the default prompt format used for \textsc{Clinical-Camel} in Appendix~\ref{sec:clinical-camel-template}.

\paragraph{\textsc{BioMistral}~\citep{biomistral}.} We use the system prompt and zero-/few-shot prompt format provided in Appendix F of the paper. At the time of writing, the code repository is not publicly available, and the few-shot example details are not fully disclosed. In Section 4.3 of \citet{biomistral}, the authors mention that the output vocabulary is constrained to be one of the answer choices in lettered format (e.g., one of [``A'',``B'',``C'',``D'']) to force the model to avoid generating irrelevant tokens in its output. However, it is unclear whether (i) the filtered token with the highest probability was treated as the model's prediction or (ii) a token was randomly sampled based on the re-normalized token probabilities. 
We include the default prompt format used for \textsc{BioMistral} in Appendix~\ref{sec:biomistral-template}.

\paragraph{\textsc{MediTron} \citep{meditron}.} We use the system prompts---tailored specifically to MedQA, MedMCQA, PubMedQA, and the MMLU datasets---provided in Table 2 of the paper. For the prompt formats, we use the ones provided in the official GitHub repository, as the prompt formats (those with special \texttt{`<|im\_start|>'} and \texttt{`<|im\_end|>'} tokens, following the ChatML format\footnote{ \href{https://github.com/openai/openai-python/blob/release-v0.28.0/chatml.md}{github.com/openai/openai-python/blob/release-v0.28.0/chatml.md}}) shown in the paper are only applicable to the fine-tuned models (see this discussion\footnote{\href{https://github.com/epfLLM/meditron/issues/13\#issuecomment-1845955741}{github.com/epfLLM/meditron/issues/13\#issuecomment-1845955741}} from the official GitHub repository). In particular, we refer to the prompt formats provided in the code\footnote{\href{https://github.com/epfLLM/meditron/blob/a7c7cda3014e537f0df2ec58f836fbe920d6283b/evaluation/benchmarks.py\#L622}{github.com/epfLLM/meditron/blob/a7c7/evaluation/benchmarks.py\#L622}} code and used for evaluation\footnote{\href{https://github.com/epfLLM/meditron/blob/a7c7cda3014e537f0df2ec58f836fbe920d6283b/evaluation/inference.py\#L188}{github.com/epfLLM/meditron/blob/a7c7/evaluation/inference.py\#L188}} to determine the default prompt format for both the 70B- and 7B-parameter models. However, we were unable to reliably reproduce the zero-/few-shot prompting performance using this prompt format, and therefore perform a grid search over the prompt formats as well for model-specific prompt selection. In the evaluation code\footnote{\href{https://github.com/epfLLM/meditron/blob/a7c7cda3014e537f0df2ec58f836fbe920d6283b/evaluation/inference.py\#L188}{github.com/epfLLM/meditron/blob/a7c7/evaluation/inference.py\#L188}}, \citet{meditron} provide the random seeds used for sampling the few-shot examples; however, we also search over the set of few-shot examples to consider a larger number of few-shot example choices. 
For sampling, we use the same greedy decoding approach as considered in the paper (``Top Token Selection'' in Section 4.3 of the paper).
We include the default prompt format used for \textsc{MediTron} in Appendix~\ref{sec:meditron-template}.

\paragraph{\textsc{BioMedGPT-LM}~\citep{biomedgpt}.} While \textsc{BioMedGPT-LM} was evaluated on medical knowledge QA tasks such as MedMCQA and PubMedQA, it was only evaluated in the SFT regime, and the prompt formats used for these datasets are not available to the best of our knowledge.
Meanwhile, the official GitHub repository provides Jupyter notebook examples\footnote{\href{https://github.com/PharMolix/OpenBioMed/blob/main/examples/biomedgpt_inference.ipynb}{github.com/PharMolix/OpenBioMed/blob/main/examples/biomedgpt\_inference.ipynb}} with a conversation format used in the context of other QA tasks. We thus use this format by default but search over the prompt formats for model-specific prompt selection, since it is not specifically designed for closed-ended multiple-choice QA tasks. Moreover, as the system prompt provided is not semantically applicable to the QA tasks that we consider (e.g., \texttt{``You are working as an excellent assistant in chemistry and molecule di- scovery.''}, we use our own manually designed default system prompt. We include the default prompt format used for \textsc{BioMedGPT-LM} in Appendix~\ref{sec:biomedgpt-template}.

\paragraph{\textsc{LLaVA-Med} \citep{llava-med}.} For \textsc{LLaVA-Med}, we use the system prompt and conversational prompt format in the ``simple\_conv\_med'' template\footnote{\href{https://github.com/microsoft/LLaVA-Med/blob/b9a98a736d2ef05bcf5ff345be6403fb3a664eaf/llava/conversation.py\#L243}{github.com/microsoft/LLaVA-Med/blob/b9a9/llava/conversation.py\#L243}} from the official GitHub repository (for \textsc{LLaVA-v0} \citep{llava}, we use the ``simple\_conv'' template\footnote{\href{https://github.com/microsoft/LLaVA-Med/blob/b9a98a736d2ef05bcf5ff345be6403fb3a664eaf/llava/conversation.py\#L257}{github.com/microsoft/LLaVA-Med/blob/b9a9/llava/conversation.py\#L257}}) by default.
For formatting the visual questions, we refer to this file\footnote{\href{https://github.com/microsoft/LLaVA-Med/blob/b9a98a736d2ef05bcf5ff345be6403fb3a664eaf/llava/eval/eval_metrics/answer-file-llava-zeorshot.jsonl}{github.com/microsoft/LLaVA-Med/blob/b9a9/llava/eval/eval\_metrics/answer-file-llava-zeorshot.jsonl}} containing the raw visual QA results on VQA-RAD (\texttt{``Please choose from the following two options: [yes, no]''}).
Meanwhile, we make these choices with the following caveats, to the best of our knowledge. 
First, the exact choice of system prompt and conversational prompt format used for evaluation are not discussed in the paper or the code repository, and we choose the one that has a system prompt specific to \textsc{LLaVA-Med} (\texttt{``You are LLaVA-Med, a large language and vision assistant trained by a group of researchers at Microsoft \ldots''}) and follows the conversational format used for \textsc{Vicuna-v0} \citep{vicuna}, which forms its LLM backbone. Second, details on how the answer choices should be formatted in the context of closed-ended QA tasks is only shown in the VQA-RAD results file.
Given the uncertainty in such details, we also search over the prompt formats for model-specific prompt selection.
We note that \textsc{LLaVA-Med} was not pretrained on multi-image inputs or evaluated in the few-shot setting, and therefore details on the choice of few-shot examples are irrelevant.
For sampling, the evaluation code\footnote{\href{https://github.com/microsoft/LLaVA-Med/blob/b9a98a736d2ef05bcf5ff345be6403fb3a664eaf/llava/eval/model_vqa_med.py}{github.com/microsoft/LLaVA-Med/blob/b9a9/llava/eval/model\_vqa\_med.py}} uses a temperature setting of 0.7 (albeit without the random seeds), which differs from our evaluation setup. We include the default prompt formats used for \textsc{LLaVA-Med} and \textsc{LLaVA-v0} in Appendices~\ref{sec:llava-med-template}--\ref{sec:llava-template}.

\paragraph{\textsc{Med-Flamingo} \citep{med-flamingo}.} We by default use the system prompt and prompt format provided in the demo code\footnote{\href{https://github.com/snap-stanford/med-flamingo/blob/7bcbb6c3932c814a6f7ae4b838ae4fada39b42a4/scripts/demo.py}{github.com/snap-stanford/med-flamingo/blob/7bcb/scripts/demo.py}} from the official GitHub repository.
However, we search over the prompt formats when performing model-specific prompt selection, as the example prompt in the demo does not show details for formatting answer choices in a closed-ended QA context.
The choice of few-shot examples and sampling details used for the original evaluations on VQA-RAD and PathVQA are not available.
We include the default prompt formats used for \textsc{Med-Flamingo} and \textsc{Open-Flamingo} in Appendices~\ref{sec:med-flamingo-template}--\ref{sec:open-flamingo-template}.

\subsection{Default LLM Prompt Formats}\label{sec:llm-prompt-templates}
Here, we share the \textit{default} prompt formats for each LLM, with MMLU (Clinical Knowledge) \citep{mmlu} as a running example. We denote the system prompt in \system{orange}, the few-shot examples in \fsexample{green}, and the question in \question{pink}. 
For models without a specific system prompt and prompt format designed for closed-ended medical QA (Section \ref{sec:reproducibility}), we use a manually designed prompt format by default. 
This includes all general-domain LLMs. 
For example, the default 1-shot prompt for non-instruction-tuned models is as follows:

\begin{mdframed}
\system{The following is a multiple-choice question about medical knowledge. Answer the question by choosing one of the options from A to D.}

\noindent \fsexample{\#\#\# Question: Glycolysis is the name given to the pathway involving the conversion of:\\ 
(A) glycogen to glucose-1-phosphate.\\
(B) glycogen or glucose to fructose.\\
(C) glycogen or glucose to pyruvate or lactate.\\
(D) glycogen or glucose to pyruvate or acetyl CoA.\\
\noindent \#\#\# Answer: (C) glycogen or glucose to pyruvate or lactate.}

\noindent \question{\#\#\# Question: What size of cannula would you use in a patient who needed a rapid blood transfusion (as of 2020 medical knowledge)?\\
(A) 18 gauge.\\
(B) 20 gauge.\\
(C) 22 gauge.\\
(D) 24 gauge.\\
\noindent \#\#\# Answer:}
\end{mdframed}

\noindent For instruction-tuned models, which expect a specific \textit{conversational} format, we apply the above format to each ``user'' query and ``assistant'' response and remove the \texttt{`\#\#\#'} and \texttt{`Answer:'} tags. For example, the input prompt to \textsc{Llama-3-70B-Instruct} is as follows:

\begin{mdframed}
<|begin\_of\_text|><|start\_header\_id|> system <|end\_header\_id|> \\
\system{The following is a multiple-choice question about medical knowledge. Answer the question by choosing one of the options from A to D.}<|eot\_id|>\\
<|start\_header\_id|>user<|end\_header\_id|> 
\noindent \fsexample{Question: Glycolysis is the name given to the pathway involving the conversion of:\\ 
(A) glycogen to glucose-1-phosphate.\\
(B) glycogen or glucose to fructose.\\
(C) glycogen or glucose to pyruvate or lactate.\\
(D) glycogen or glucose to pyruvate or acetyl CoA.}<|eot\_id|>\\
<|start\_header\_id|>assistant<|end\_header\_id|>\\
\noindent \fsexample{(C) glycogen or glucose to pyruvate or lactate.}<|eot\_id|>\\
<|start\_header\_id|>user<|end\_header\_id|>

\noindent \question{Question: What size of cannula would you use in a patient who needed a rapid blood transfusion (as of 2020 medical knowledge)?\\
(A) 18 gauge.\\
(B) 20 gauge.\\
(C) 22 gauge.\\
(D) 24 gauge.}<|eot\_id|>\\
<|start\_header\_id|>assistant<|end\_header\_id|>
\end{mdframed}
In the following subsections, we show the system prompt and prompt formats used in the 1-shot setting for models that have a dedicated format. We exclude the model-specific special tokens (e.g., \texttt{`[INST]'}) for ease of presentation, and add \texttt{`[User]'} and \texttt{`[Model]'} to demarcate each question and answer for the instruction-tuned models.

\subsubsection{\textsc{Med42-v2} \citep{med42-v2}}\label{sec:med42-v2-template}
\begin{mdframed}
\system{You are a helpful, respectful and honest medical assistant. You are a second version of Med42 developed by the AI team at M42, UAE. Always answer as helpfully as possible, while being safe. Your answers should not include any harmful, unethical, racist, sexist, toxic, dangerous, or illegal content. Please ensure that your responses are socially unbiased and positive in nature. If a question does not make any sense, or is not factually coherent, explain why instead of answering something not correct. If you don’t know the answer to a question, please don’t share false information.}

\noindent \textbf{[User]} \fsexample{Question: Glycolysis is the name given to the pathway involving the conversion of:\\
(A) glycogen to glucose-1-phosphate.\\
(B) glycogen or glucose to fructose.\\
(C) glycogen or glucose to pyruvate or lactate.\\
(D) glycogen or glucose to pyruvate or acetyl CoA.}\\
\fsexample{
\textcolor{black}{\textbf{[Model]}} (C) glycogen or glucose to pyruvate or lactate.
}

\noindent \textbf{[User]} \question{Question: What size of cannula would you use in a patient who needed a rapid blood transfusion (as of 2020 medical knowledge)?\\
(A) 18 gauge.\\
(B) 20 gauge.\\
(C) 22 gauge.\\
(D) 24 gauge.
}
\end{mdframed}

\subsubsection{\textsc{Med42-v1} \citep{med42-v1}}\label{sec:med42-v1-template}
\begin{mdframed}
<|system|>: \system{You are a helpful medical assistant created by M42 Health in the UAE.}
\noindent <|prompter|>: \fsexample{Question: Glycolysis is the name given to the pathway involving the conversion of:\\
(A) glycogen to glucose-1-phosphate.\\
(B) glycogen or glucose to fructose.\\
(C) glycogen or glucose to pyruvate or lactate.\\
(D) glycogen or glucose to pyruvate or acetyl CoA.}\\
\noindent <|assistant|>: \fsexample{(C) glycogen or glucose to pyruvate or lactate.}\\
\noindent <|prompter|>: \question{Question: What size of cannula would you use in a patient who needed a rapid blood transfusion (as of 2020 medical knowledge)?\\
(A) 18 gauge.\\
(B) 20 gauge.\\
(C) 22 gauge.\\
(D) 24 gauge.
}
\end{mdframed}

\subsubsection{\textsc{OpenBioLLM} \citep{OpenBioLLMs}}\label{sec:openbiollm-template}
\begin{mdframed}
\system{You are an expert and experienced from the healthcare and biomedical domain with extensive medical knowledge and practical experience. Your name is OpenBioLLM, and you were developed by Saama AI Labs. who's willing to help answer the user's query with explanation. In your explanation, leverage your deep medical expertise such as relevant anatomical structures, physiological processes, diagnostic criteria, treatment guidelines, or other pertinent medical concepts. Use precise medical terminology while still aiming to make the explanation clear and accessible to a general audience.}

\noindent \textbf{[User]} \fsexample{Question: Glycolysis is the name given to the pathway involving the conversion of:\\
(A) glycogen to glucose-1-phosphate.\\
(B) glycogen or glucose to fructose.\\
(C) glycogen or glucose to pyruvate or lactate.\\
(D) glycogen or glucose to pyruvate or acetyl CoA.}\\
\fsexample{
\textcolor{black}{\textbf{[Model]}} (C) glycogen or glucose to pyruvate or lactate.
}

\noindent \textbf{[User]} \question{Question: What size of cannula would you use in a patient who needed a rapid blood transfusion (as of 2020 medical knowledge)?\\
(A) 18 gauge.\\
(B) 20 gauge.\\
(C) 22 gauge.\\
(D) 24 gauge.
}

\end{mdframed}

\subsubsection{\textsc{Clinical-Camel} \citep{clinical-camel}}\label{sec:clinical-camel-template}
\begin{mdframed}
\system{The following is a multiple-choice question about medical knowledge. Answer the question by choosing one of the options from A to D.}\\
\fsexample{\textcolor{black}{\textbf{[User]}} Question: Glycolysis is the name given to the pathway involving the conversion of:\\
(A) glycogen to glucose-1-phosphate.\\
(B) glycogen or glucose to fructose.\\
(C) glycogen or glucose to pyruvate or lactate.\\
(D) glycogen or glucose to pyruvate or acetyl CoA.
}\\
\fsexample{\textcolor{black}{\textbf{[Model]}} (C) glycogen or glucose to pyruvate or lactate.}\\
\question{\textcolor{black}{\textbf{[User]}} Question: What size of cannula would you use in a patient who needed a rapid blood transfusion (as of 2020 medical knowledge)?\\
(A) 18 gauge.\\
(B) 20 gauge.\\
(C) 22 gauge.\\
(D) 24 gauge.
}
\end{mdframed}

\subsubsection{\textsc{BioMistral} \citep{biomistral}}\label{sec:biomistral-template}
\begin{mdframed}
\system{The following are multiple choice questions (with answers) about medical knowledge.}

\noindent \fsexample{**Question:** Glycolysis is the name given to the pathway involving the conversion of:\\
(A) glycogen to glucose-1-phosphate.\\
(B) glycogen or glucose to fructose.\\
(C) glycogen or glucose to pyruvate or lactate.\\
(D) glycogen or glucose to pyruvate or acetyl CoA.\\
**Answer:** (C
}

\noindent \question{**Question:** What size of cannula would you use in a patient who needed a rapid blood transfusion (as of 2020 medical knowledge)?\\
(A) 18 gauge.\\
(B) 20 gauge.\\
(C) 22 gauge.\\
(D) 24 gauge.\\
**Answer:** (
}
\end{mdframed}

\subsubsection{\textsc{MediTron} \citep{meditron}}\label{sec:meditron-template}
\begin{mdframed}
\system{You are a medical doctor answering real-world medical entrance exam questions. Based on your understanding of basic and clinical science, medical knowledge, and mechanisms underlying health, disease, patient care, and modes of therapy, answer the following multiple-choice question. Select one correct answer from A to D. Base your answer on the current and standard practices referenced in medical guidelines.}

\noindent \fsexample{Question: Glycolysis is the name given to the pathway involving the conversion of:\\
Options:\\
A. glycogen to glucose-1-phosphate.\\
B. glycogen or glucose to fructose.\\
C. glycogen or glucose to pyruvate or lactate.\\
D. glycogen or glucose to pyruvate or acetyl CoA.\\
The answer is: C}\\
\noindent \question{Question: What size of cannula would you use in a patient who needed a rapid blood transfusion (as of 2020 medical knowledge)?\\
Options:\\
A. 18 gauge.\\
B. 20 gauge.\\
C. 22 gauge.\\
D. 24 gauge.\\
The answer is:
}
\end{mdframed}

\subsubsection{\textsc{BioMedGPT-LM} \citep{biomedgpt}}\label{sec:biomedgpt-template}
\begin{mdframed}
\system{The following is a multiple-choice question about medical knowledge. Answer the question by choosing one of the options from A to D.}

\noindent \fsexample{\#\#\# Human: Glycolysis is the name given to the pathway involving the conversion of:\\
(A) glycogen to glucose-1-phosphate.\\
(B) glycogen or glucose to fructose.\\
(C) glycogen or glucose to pyruvate or lactate.\\
(D) glycogen or glucose to pyruvate or acetyl CoA.\\
\#\#\# Assistant: (C) glycogen or glucose to pyruvate or lactate.
}

\noindent \question{\#\#\# Human: What size of cannula would you use in a patient who needed a rapid blood transfusion (as of 2020 medical knowledge)?\\
(A) 18 gauge.\\
(B) 20 gauge.\\
(C) 22 gauge.\\
(D) 24 gauge.\\
\#\#\# Assistant: 
}
\end{mdframed}

\subsection{Default VLM Prompt Formats}\label{sec:vlm-prompt-templates}
Here, we share the \textit{default} prompt formats that we use for each VLM, using VQA-RAD \citep{vqa-rad} as a running example. We denote the system prompt in \system{orange}, the few-shot examples in \fsexample{green}, and the question \question{pink}.
We show the format in the 1-shot setting.

\subsubsection{\textsc{LLaVA-Med} \citep{llava-med}}\label{sec:llava-med-template}

\begin{mdframed}
\system{You are LLaVA-Med, a large language and vision assistant trained by a group of researchers at Microsoft, based on the general domain LLaVA architecture. You are able to understand the visual content that the user provides, and assist the user with a variety of medical and clinical tasks using natural language.}\\
\system{Follow the instructions carefully and explain your answers in detail.}

\noindent \fsexample{\#\#\# Human: Does this patient have multiple lesions in their chest? Please choose from the following options: [yes, no]. <image>\\
\#\#\# Assistant: no
}

\noindent \question{\#\#\# Human: Is there evidence of an aortic aneurysm? Please choose from the following options: [yes, no]. <image>\\
\#\#\# Assistant: 
}
\end{mdframed}

\subsubsection{\textsc{LLaVA-v0} \citep{llava}}\label{sec:llava-template}
\begin{mdframed}
\system{A chat between a curious human and an artificial intelligence assistant. The assistant gives helpful, detailed, and polite answers to the human's questions.}

\noindent \fsexample{\#\#\# Human: Does this patient have multiple lesions in their chest? Please choose from the following options: [yes, no]. <image>\\
\#\#\# Assistant: no
}

\noindent \question{\#\#\# Human: Is there evidence of an aortic aneurysm? Please choose from the following options: [yes, no]. <image>\\
\#\#\# Assistant:
}
\end{mdframed}

\subsubsection{\textsc{Med-Flamingo} \citep{med-flamingo}}\label{sec:med-flamingo-template}
\begin{mdframed}
\system{You are a helpful medical assistant. You are being provided with images, a question about the image and an answer. Follow the examples and answer the last question.}

\noindent \fsexample{<image> Does this patient have multiple lesions in their chest?\\
(A) yes\\
(B) no\\
Answer: (B) no <|endofchunk|>
}

\noindent \question{<image> Is there evidence of an aortic aneurysm?\\
(A) yes\\
(B) no\\
Answer: 
}
\end{mdframed}

\subsubsection{\textsc{Open-Flamingo} \citep{open-flamingo}}\label{sec:open-flamingo-template}
\begin{mdframed}
\system{The following is a multiple-choice visual question requiring medical knowledge. Answer the question by choosing one of the provided answer options.}

\noindent \fsexample{<image> Does this patient have multiple lesions in their chest?\\
(A) yes\\
(B) no}\\
\fsexample{Answer: (B) no <|endofchunk|>
}
\noindent \question{<image> Is there evidence of an aortic aneurysm?\\
(A) yes\\
(B) no\\
Answer: 
}
\end{mdframed}

\renewcommand\thetable{D\arabic{table}}
\renewcommand\thefigure{D\arabic{figure}}
\setcounter{table}{0}
\setcounter{figure}{0}
\section{Zero-/Few-Shot Prompting Evaluation Results with Constrained Decoding}\label{sec:constrained-decoding}
\begin{table}[h!]
    \centering
    \caption{The zero-shot and 3-shot win/tie/loss rates (\%) of all medical LLMs on textual medical knowledge QA, after independently optimizing the prompt for each model. For each medical model, we boldface the win rate if it wins more than it loses to its general-domain base model, and vice versa. Here, we show the results when model predictions are generated via constrained decoding. The results for greedy decoding are shown in Table \ref{tab:win-tie-loss-rates-knowledge}.}
    \label{tab:win-tie-loss-rates-logprob-knowledge}
    \resizebox{0.65\linewidth}{!}{
    
    \begin{tabular}{@{}l@{\hskip 20pt}c@{\hskip 10pt}c@{\hskip 10pt}c@{\hskip 20pt}c@{\hskip 10pt}c@{\hskip 10pt}c@{\hskip 10pt}}
        \toprule
        \multirow{2}{*}{\textbf{Model}} & \multicolumn{3}{c}{\textbf{Zero-Shot}} & \multicolumn{3}{c}{\textbf{3-Shot}} \\
        \cmidrule{2-4} \cmidrule{5-7}
        & Win & Tie & Loss & Win & Tie & Loss \\
        \midrule
        \textsc{OpenBioLLM-70B}        & 7.7 & 76.9 & \textbf{15.4} & 23.1 & 53.8 & 23.1 \\
        \textsc{MediTron-70B}             & \textbf{30.8} & 46.2 & 23.1 & 15.4 & 69.2 & 15.4 \\
        \textsc{Clinical-Camel-70B} & \textbf{18.2} & 72.7 & 9.1 & 9.1 & 72.7 & \textbf{18.2} \\
        \midrule
        \textsc{OpenBioLLM-8B}         & 0 & 53.8 & \textbf{46.2} & 7.7 & 69.2 & \textbf{23.1} \\
        \textsc{MediTron-7B}              & \textbf{23.1} & 76.9 & 0 & \textbf{7.7} & 92.3 & 0 \\
        \textsc{BioMistral-7B}          & \textbf{30.8} & 69.2 & 0 & \textbf{7.7} & 92.3 & 0 \\
        \textsc{BioMedGPT-LM-7B}         & 7.7 & 84.6 & 7.7 & \textbf{7.7} & 69.2 & \textbf{23.1} \\
        \midrule
        \textbf{Aggregate}                                 & \textbf{16.9} & 68.5 & 14.6 & 11.2 & 74.2 & \textbf{14.6} \\
        \bottomrule
    \end{tabular}
    }
\end{table}

\noindent Here, we provide the constrained decoding results for the zero-/few-shot prompting evaluations described in Section \ref{sec:prompting} (see ``Models'' in Section \ref{sec:eval-setup} for a description on constrained decoding). 
In Appendices \ref{sec:prompting-results-1-knowledge-logprob}--\ref{sec:prompting-results-1-vqa-logprob}, corresponding to Sections \ref{sec:prompting-results-1-knowledge}--\ref{sec:prompting-results-1-vqa}, we show the results on
the textual medical knowledge QA, textual clinical note QA, and visual medical QA datasets, when model-specific prompt selection is performed for each model independently. 
In Appendix \ref{sec:prompting-results-2-logprob}, corresponding to Section \ref{sec:prompting-results-2-aggregate}, we show the aggregate results on how the win, tie, and loss rates (\%) of the medical models are affected by the inclusion (or lack thereof) of model-specific prompt selection and statistical testing.

\subsection{Evaluation of Medical LLMs on Textual Medical Knowledge QA (Section~\ref{sec:prompting-results-1-knowledge})}
\label{sec:prompting-results-1-knowledge-logprob}

\begin{figure*}[t!]
    \centering
    \begin{tabular}{@{}c@{}c@{}}
        \multicolumn{2}{c}{
            \begin{subfigure}{0.98\linewidth}
                \includegraphics[width=\linewidth]{figs/llm-acc-ci-legend-v3.pdf}
            \end{subfigure}
        }
        \\
        \begin{subfigure}{0.02\linewidth}
            \makebox[\linewidth]{\raisebox{75pt}{{(a)}}}
        \end{subfigure} &
        \begin{subfigure}{0.96\linewidth}
            \includegraphics[width=\linewidth]{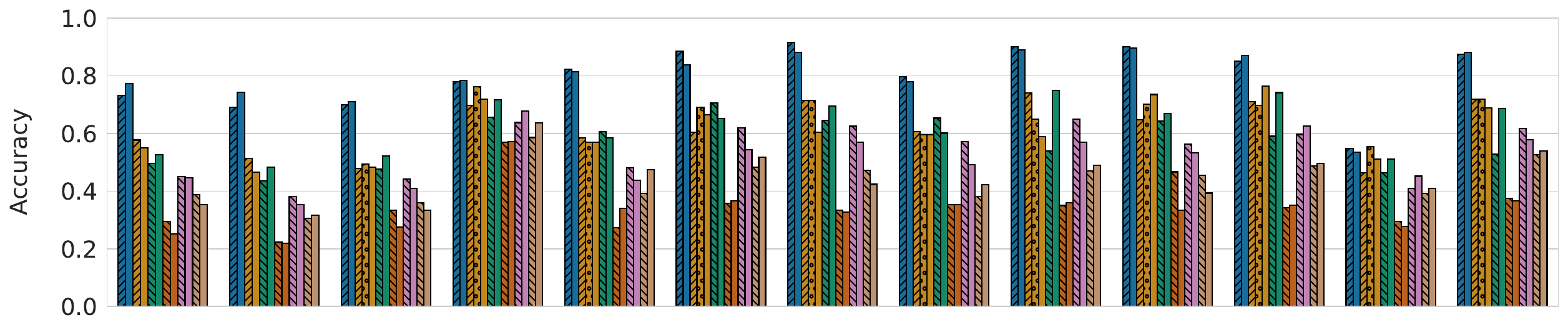}
        \end{subfigure}
        \\
        \begin{subfigure}{0.02\linewidth}
        \end{subfigure} &
        \begin{subfigure}{0.96\linewidth}
            \includegraphics[width=\linewidth]{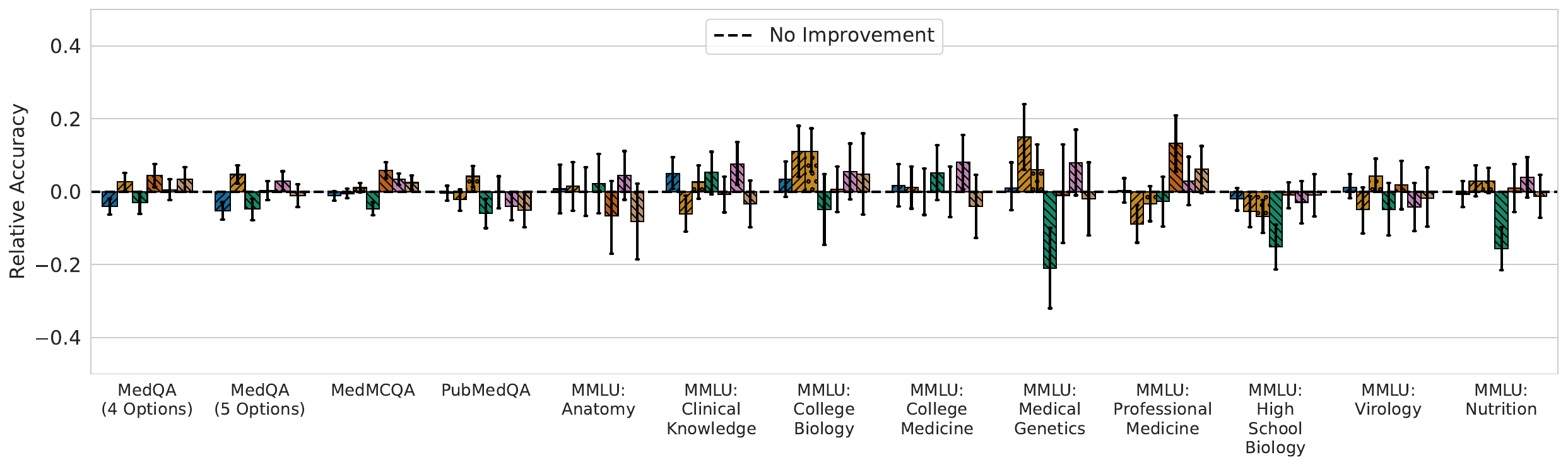}
        \end{subfigure}
        \\
        \begin{subfigure}{0.02\linewidth}
            \makebox[\linewidth]{\raisebox{75pt}{{(b)}}}
        \end{subfigure} &
        \begin{subfigure}{0.96\linewidth}
            \includegraphics[width=\linewidth]{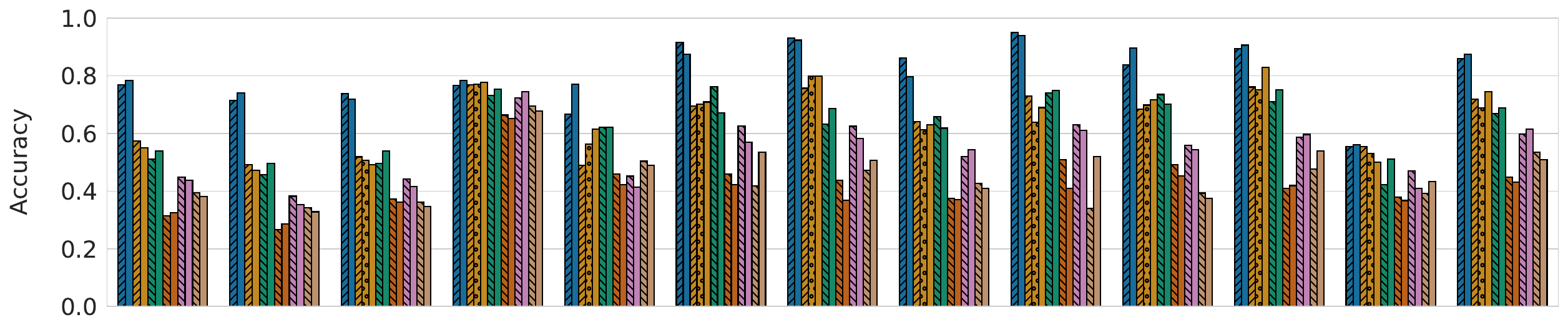}
        \end{subfigure}
        \\
        \begin{subfigure}{0.02\linewidth}
        \end{subfigure} &
        \begin{subfigure}{0.96\linewidth}
            \includegraphics[width=\linewidth]{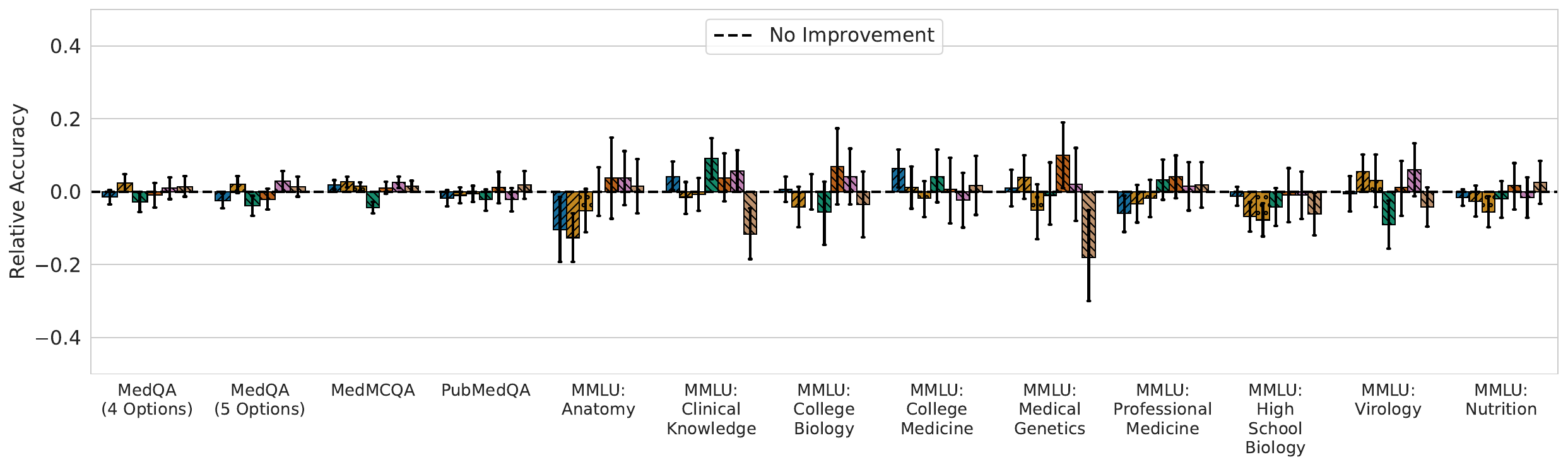}
        \end{subfigure}
    \end{tabular}
    \vspace{-10pt}
    \caption{After independently selecting the best prompt format and examples for each model, medical LLMs (textured bars) fail to consistently improve over their base models (solid bars) on textual medical knowledge QA tasks, in both (a) zero-shot and (b) 3-shot settings. 
    In each panel, the top row shows the absolute exact-match accuracies on the test set, and the bottom row shows the relative exact-match accuracies along with 95\% confidence intervals derived via bootstrapping on the test set (Section~\ref{sec:eval-setup}). 
    Here, model predictions are generated via constrained decoding.
    Greedy decoding results are shown in Figure \ref{fig:llm-acc-ci-knowledge}.
    }
    \label{fig:llm-acc-ci-logprob-knowledge}
\end{figure*}

In Figures \ref{fig:llm-acc-ci-logprob-knowledge}(a) and (b), we show the absolute and relative exact-match accuracies achieved by the medical and general-domain LLMs on the textual medical knowledge QA datasets, from zero-shot and 3-shot prompting, respectively. 
In Table \ref{tab:win-tie-loss-rates-logprob-knowledge}, we also show the win, tie, and loss rates (\%) of the medical LLMs, where win rate refers to the proportion of QA datasets where a medical model shows a statistically significant improvement over its base model. 
We exclude the results for \textsc{Clinical-Camel-70B} on both versions of MedQA, and the \textsc{Med42} models on all textual medical knowledge QA datasets, as they have already been trained on these datasets via medical DAPT (Table \ref{tab:models}). 

Overall, we observe that our findings discussed in Section \ref{sec:prompting-results-1-knowledge} also hold in the constrained decoding setting.
In particular, we observe that
\begin{enumerate}[topsep=0.5ex,itemsep=-0.5ex]
    \item medical and general-domain LLMs generally achieve higher accuracy across all datasets, going from the zero-shot to 3-shot setting;
    \item more recent general-domain models (e.g., \textsc{Llama-3-8B}) tend to show stronger performance even without any medical adaptation via DAPT; and
    \item the majority of medical LLMs show limited improvement over their base models, reaching a tie in most cases (aggregate zero-shot/3-shot tie rates of 68.5\%/74.2\%).
\end{enumerate}
These results suggest that even when we constrain the outputs of each model to always produce a valid answer choice (which is not guaranteed with greedy decoding, especially in the zero-shot setting), the performance benefits from medical DAPT on textual medical knowledge QA datasets are overall limited in the zero-/few-shot prompting regime.

\subsection{Evaluation of Medical LLMs on Textual Clinical Note QA (Section~\ref{sec:prompting-results-1-clinical})}
\label{sec:prompting-results-1-clinical-logprob}

\begin{figure*}[t!]
    \centering
    \begin{tabular}{@{}c@{}c@{\hskip 7pt}c@{}c@{}}
        \multicolumn{4}{c}{
            \begin{subfigure}{0.95\linewidth}
                \includegraphics[width=\linewidth]{figs/llm-acc-ci-legend-clinical.pdf}
            \end{subfigure}
        }
        \\
        \begin{subfigure}{0.03\linewidth}
            \makebox[\linewidth]{\raisebox{75pt}{{(a)}}}
        \end{subfigure} &
        \begin{subfigure}{0.64\linewidth}
            \includegraphics[width=\linewidth]{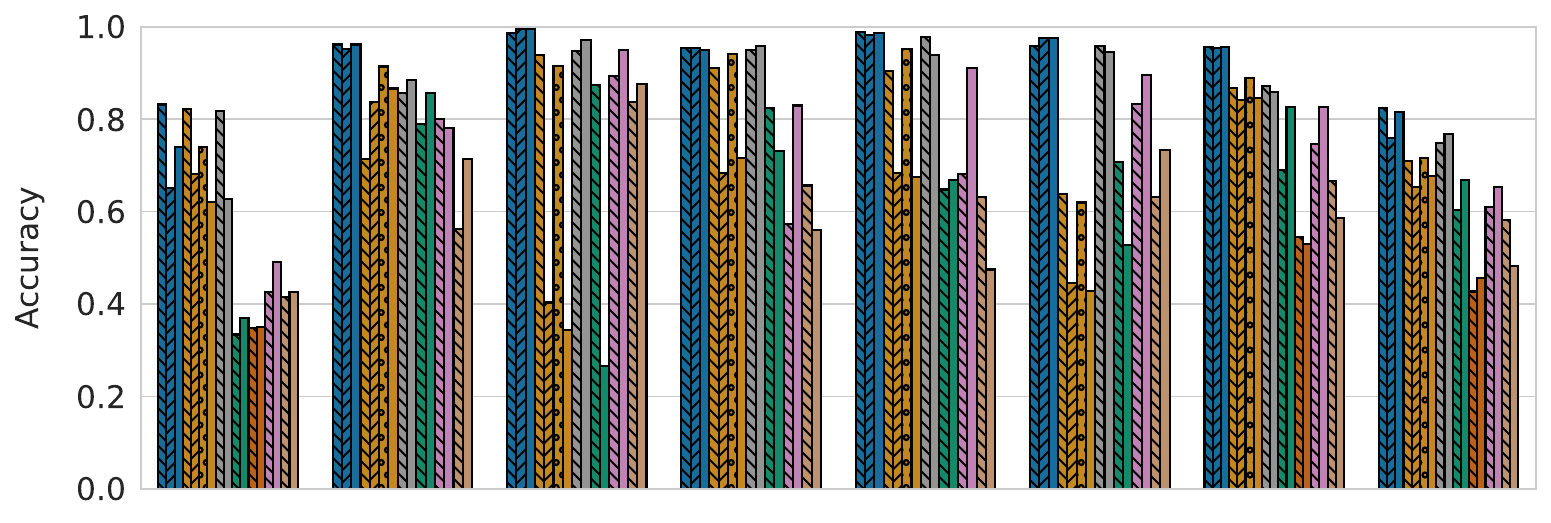}
        \end{subfigure} &
        \begin{subfigure}{0.03\linewidth}
            \makebox[\linewidth]{\raisebox{75pt}{{(b)}}}
        \end{subfigure} &
        \begin{subfigure}{0.26\linewidth}
            \includegraphics[width=\linewidth]{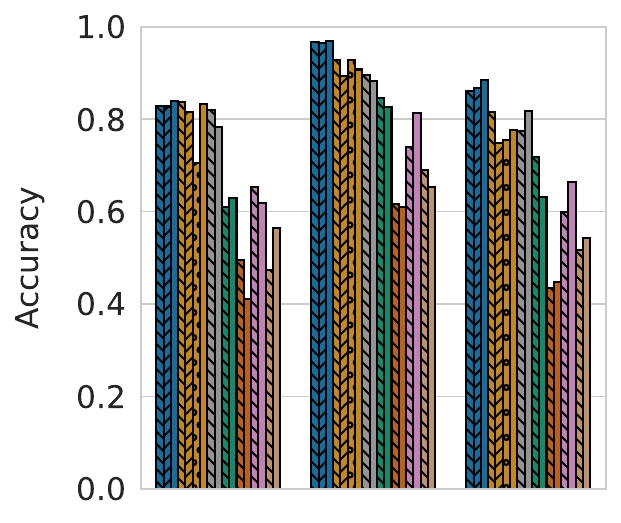}
        \end{subfigure}
        \\
        \begin{subfigure}{0\linewidth}
        \end{subfigure} &
        \begin{subfigure}{0.64\linewidth}
            \includegraphics[width=\linewidth]{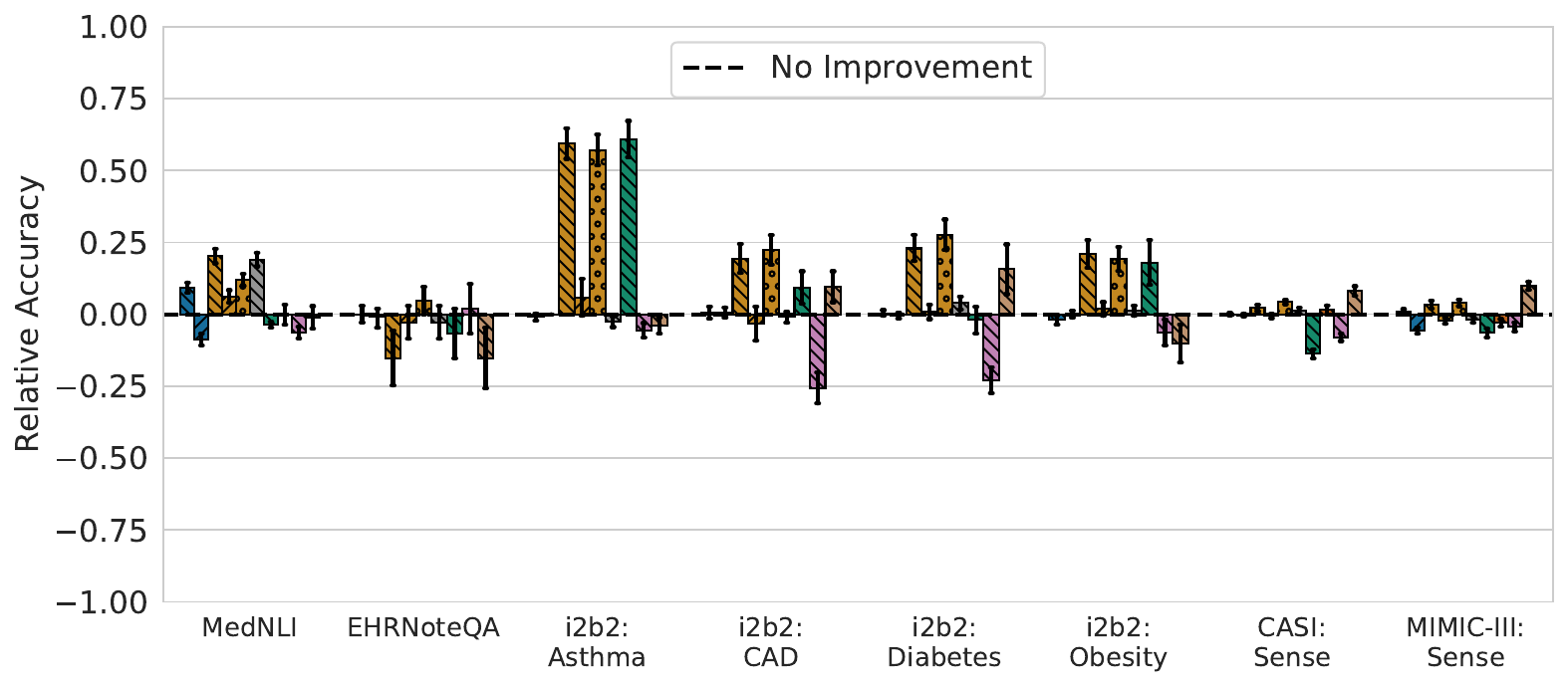}
        \end{subfigure} &
        \begin{subfigure}{0\linewidth}
        \end{subfigure} &
        \begin{subfigure}{0.26\linewidth}
            \includegraphics[width=\linewidth]{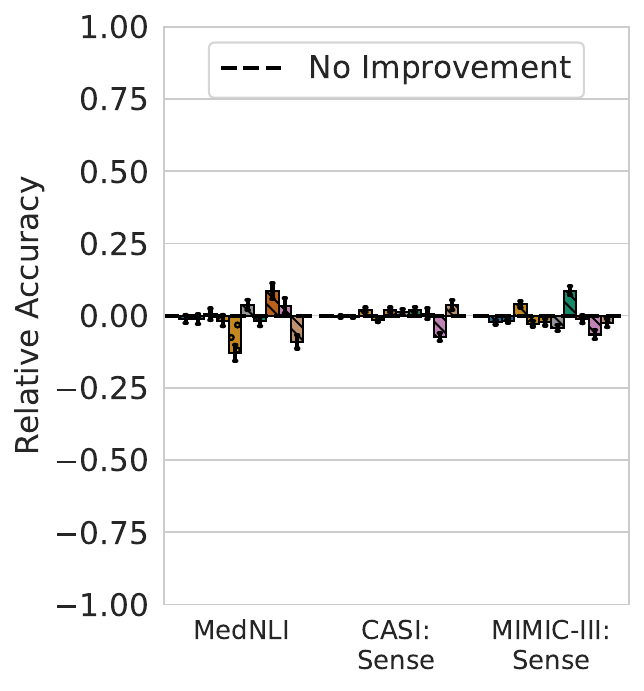}
        \end{subfigure}
    \end{tabular}
    \vspace{-10pt}
    \caption{After independently selecting the best prompt format and examples for each model, medical LLMs (textured bars) fail to consistently improve over their base models (solid bars) on textual clinical note QA tasks, in both (a) zero-shot and (b) 3-shot settings. 
    In the 3-shot setting, we exclude the results on the EHRNoteQA and i2b2 datasets given the context window limitations (Section \ref{sec:prompting}). 
    In each panel, the top row shows the absolute exact-match accuracies on the test set, and the bottom row shows the relative exact-match accuracies along with 95\% confidence intervals derived via bootstrapping on the test set (Section~\ref{sec:eval-setup}). 
    Here, model predictions are generated via constrained decoding. 
    Greedy decoding results are shown in Figure \ref{fig:llm-acc-ci-clinical}.
    }
    \label{fig:llm-acc-ci-clinical-logprob}
\end{figure*}

\begin{figure*}[t!]
    \centering
    \begin{tabular}{@{}c@{}c@{}c@{}}
        \multicolumn{3}{c}{
            \begin{subfigure}{0.95\linewidth}
                \includegraphics[width=\linewidth]{figs/llm-acc-ci-legend-clinical.pdf}
            \end{subfigure}
        }
        \\
        \begin{subfigure}{0.05\linewidth}
            \makebox[\linewidth]{\raisebox{65pt}{{(a)}}}
        \end{subfigure} &
        \begin{subfigure}{0.45\linewidth}
            \includegraphics[width=\linewidth]{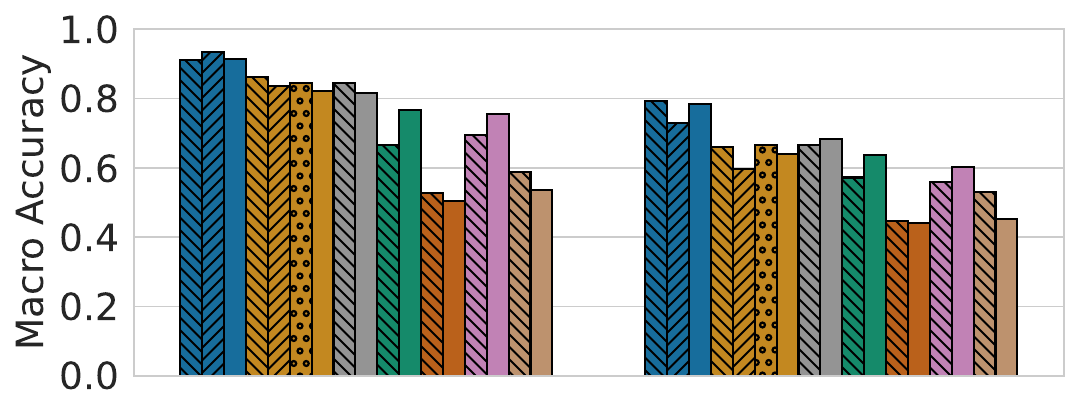}
        \end{subfigure} &
        \begin{subfigure}{0.45\linewidth}
            \includegraphics[width=\linewidth]{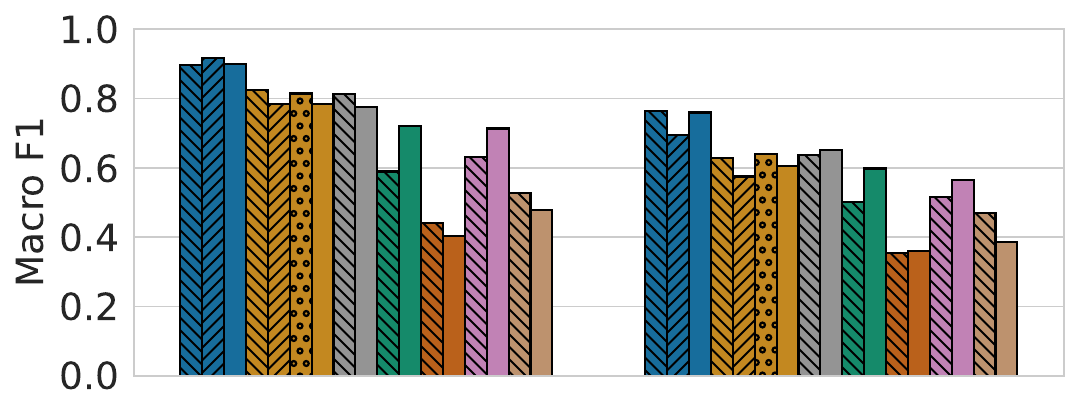}
        \end{subfigure}
        \\
        \begin{subfigure}{0\linewidth}
        \end{subfigure} &
        \begin{subfigure}{0.45\linewidth}
            \includegraphics[width=\linewidth]{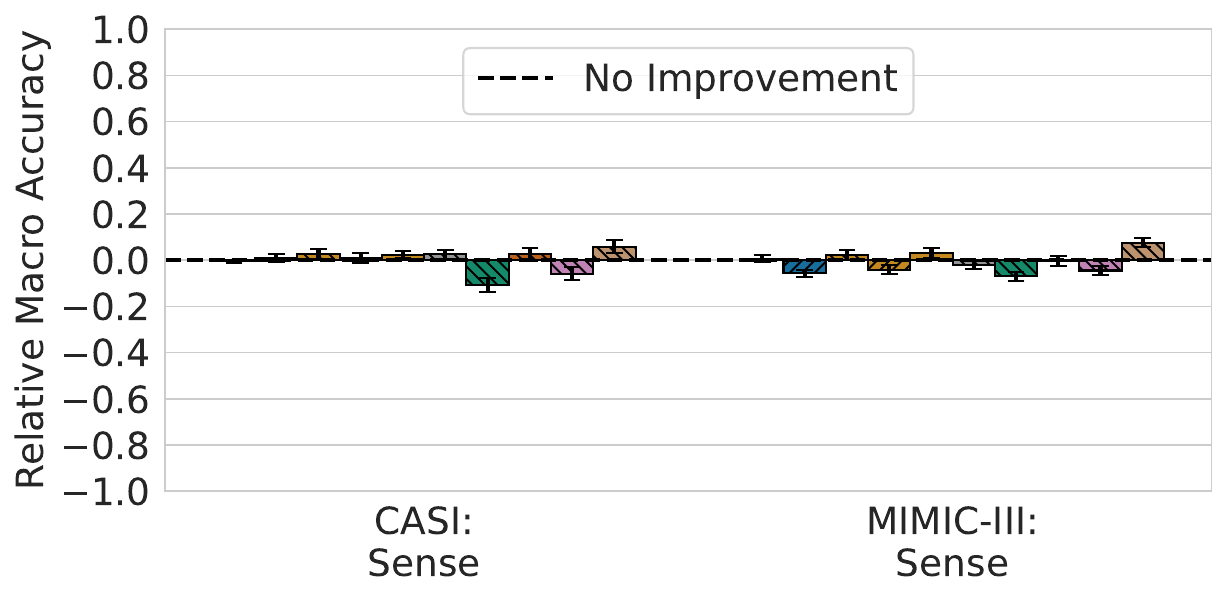}
        \end{subfigure} &
        \begin{subfigure}{0.45\linewidth}
            \includegraphics[width=\linewidth]{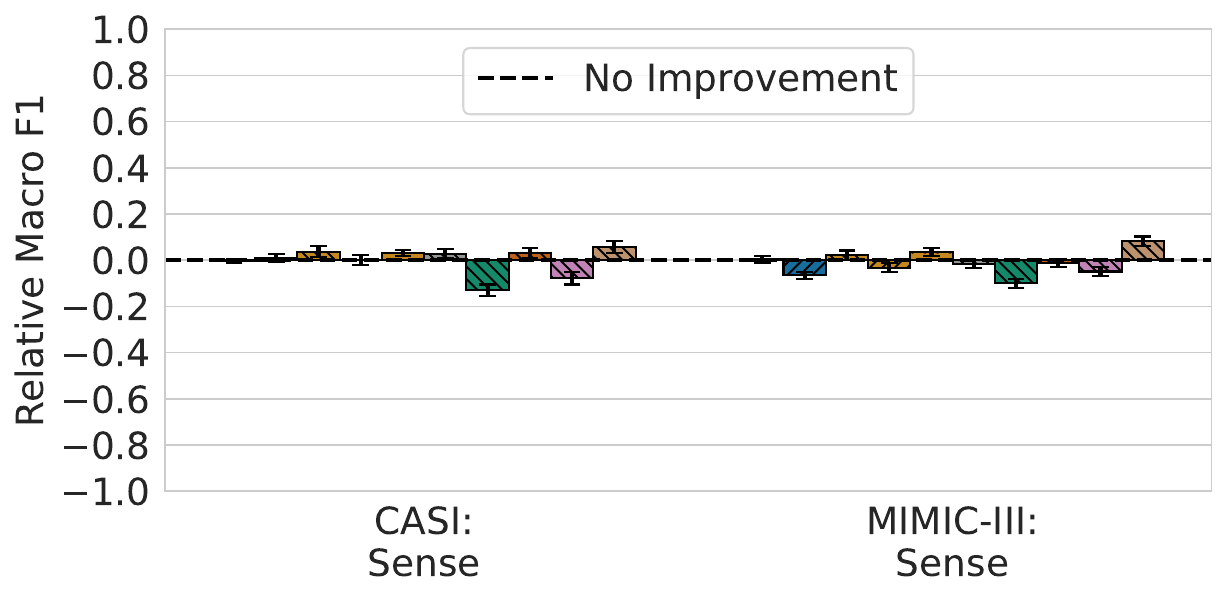}
        \end{subfigure} \\
        \begin{subfigure}{0.05\linewidth}
            \makebox[\linewidth]{\raisebox{65pt}{{(b)}}}
        \end{subfigure} &
        \begin{subfigure}{0.45\linewidth}
            \includegraphics[width=\linewidth]{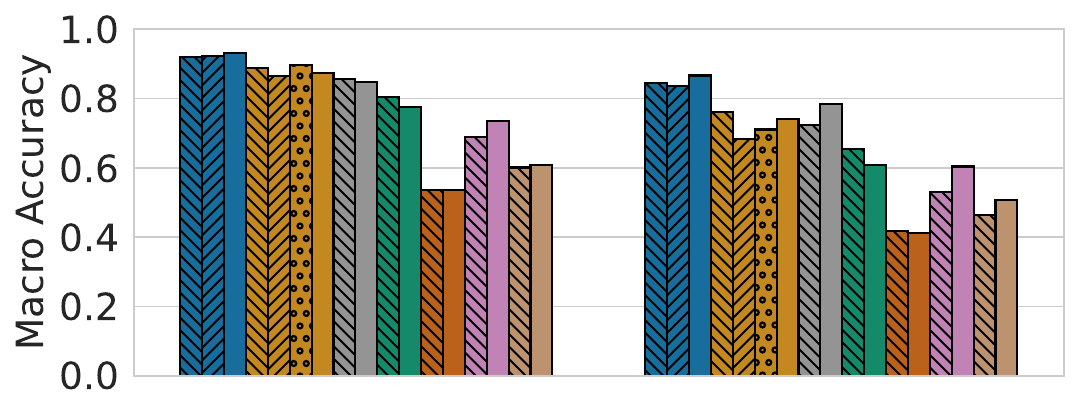}
        \end{subfigure} &
        \begin{subfigure}{0.45\linewidth}
            \includegraphics[width=\linewidth]{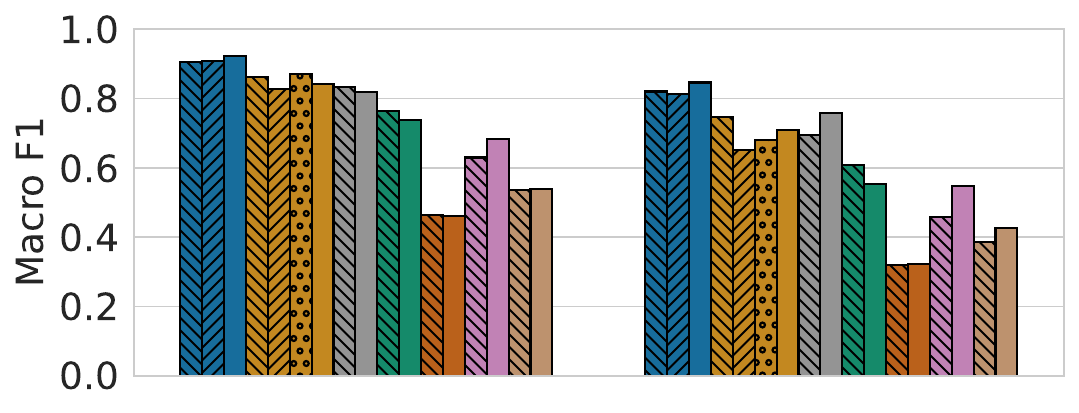}
        \end{subfigure} \\
        \begin{subfigure}{0\linewidth}
        \end{subfigure} &
        \begin{subfigure}{0.45\linewidth}
            \includegraphics[width=\linewidth]{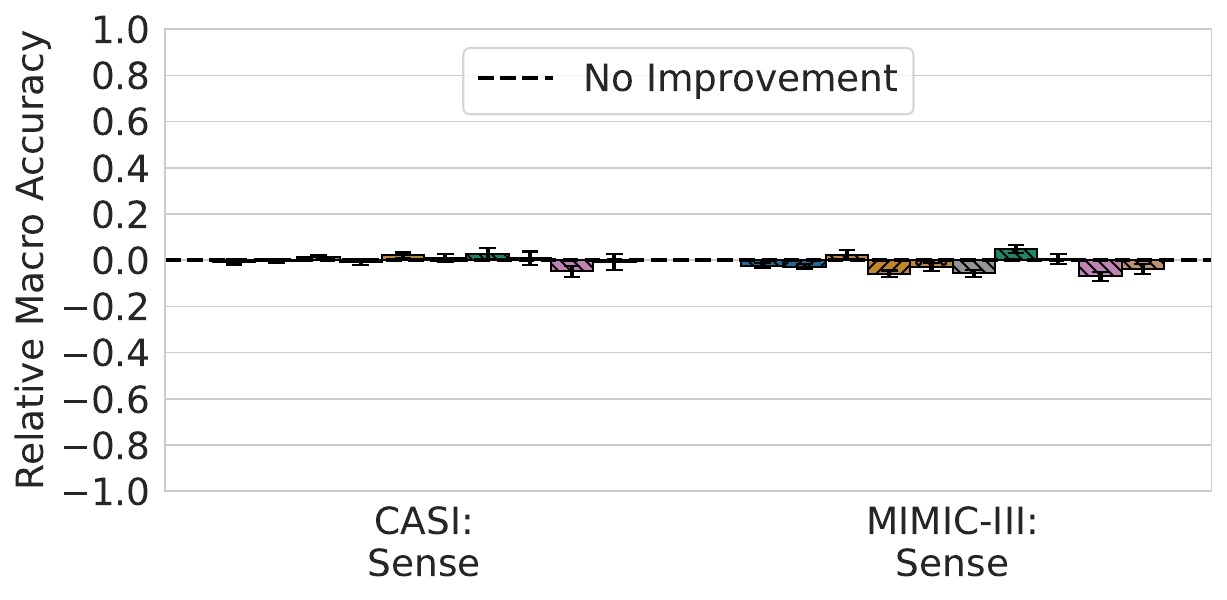}
        \end{subfigure} &
        \begin{subfigure}{0.45\linewidth}
            \includegraphics[width=\linewidth]{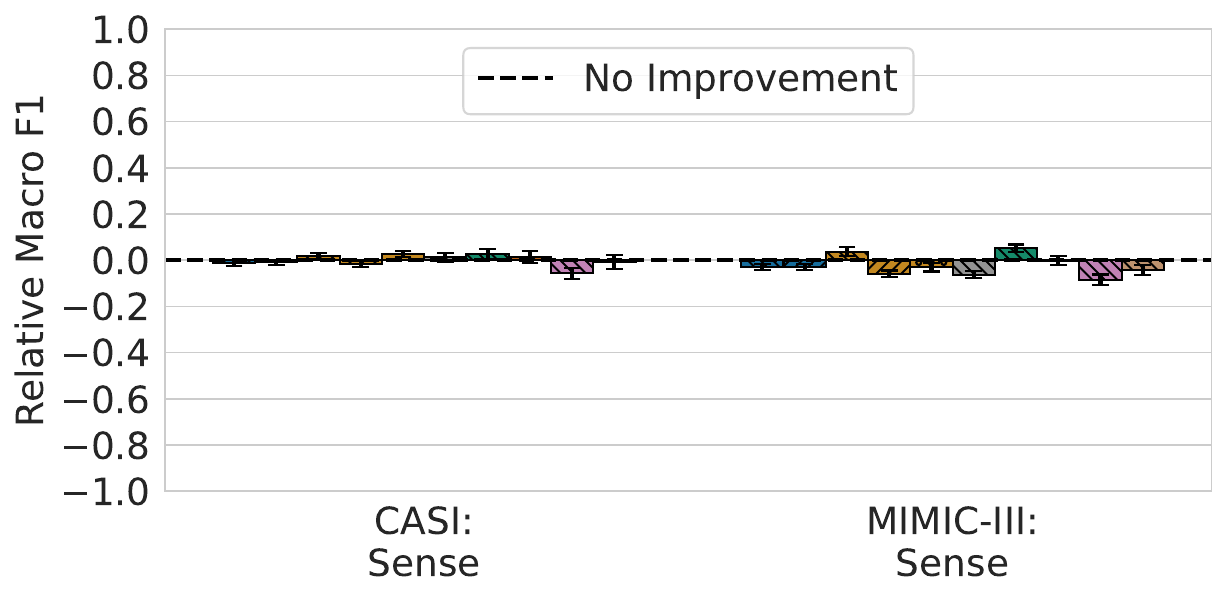}
        \end{subfigure}
    \end{tabular}
    \vspace{-10pt}
    \caption{Even after accounting for the imbalance in the distribution of clinical acronyms in the CASI and MIMIC-III datasets, medical LLMs (textured bars) fail to consistently improve over their base models (solid bars), in both (a) zero-shot and (b) 3-shot settings. 
    We show the results when the prompts are optimized for each model independently.
    In each panel, the top row shows the absolute macro exact-match accuracies and F1 scores---averaged over clinical acronyms---on the test set, and the bottom row shows the relative macro exact-match accuracies and F1 scores along with 95\% confidence intervals derived via bootstrapping on the test set (Section~\ref{sec:eval-setup}). 
    Here, model predictions are generated via constrained decoding. 
    Greedy decoding results are shown in Figure \ref{fig:llm-acc-ci-sense}.
    }
    \label{fig:llm-acc-ci-sense-logprob}
\end{figure*}

\begin{table}[t!]
    \centering
    \caption{The zero-shot and 3-shot win, tie, and loss rates (\%) of all medical LLMs on textual clinical note QA, after independently optimizing the prompt for each model. For each medical model, we boldface the win rate if it wins more than it loses to its general-domain base model, and vice versa. Here, we show the results when model predictions are generated via constrained decoding. The results for greedy decoding are shown in Table \ref{tab:win-tie-loss-rates-clinical}.}
    \label{tab:win-tie-loss-rates-logprob-clinical}
    \resizebox{0.65\linewidth}{!}{
    
    \begin{tabular}{@{}l@{\hskip 20pt}c@{\hskip 10pt}c@{\hskip 10pt}c@{\hskip 20pt}c@{\hskip 10pt}c@{\hskip 10pt}c@{\hskip 10pt}}
        \toprule
        \multirow{2}{*}{\textbf{Model}} & \multicolumn{3}{c}{\textbf{Zero-Shot}} & \multicolumn{3}{c}{\textbf{3-Shot}} \\
        \cmidrule{2-4} \cmidrule{5-7}
        & Win & Tie & Loss & Win & Tie & Loss \\
        \midrule
        \textsc{Med42-v2-70B}             & 12.5 & 75.0 & 12.5 & 0 & 66.7 & \textbf{33.3} \\
        \textsc{OpenBioLLM-70B}        & 0 & 75.0 & \textbf{25.0} & 0 & 33.3 & \textbf{66.7} \\
        \textsc{Med42-v1-70B}             & \textbf{87.5} & 0 & 0 & \textbf{66.7} & 33.3 & 0 \\
        \textsc{MediTron-70B}            & 12.5 & 75.0 & 12.5 & 0 & 33.3 & \textbf{66.7} \\
        \textsc{Clinical-Camel-70B} & \textbf{87.5} & 12.5 & 0 & 33.3 & 0 & \textbf{66.7} \\
        \midrule
        \textsc{Med42-v2-8B}              & \textbf{37.5} & 37.5 & 25.0 & \textbf{66.7} & 0 & 33.3 \\
        \textsc{OpenBioLLM-8B}         & 37.5 & 25.0 & 37.5 & \textbf{66.7} & 0 & 33.3 \\
        \textsc{MediTron-7B}              & 0 & 66.7 & \textbf{33.3} & \textbf{33.3} & 66.7 & 0 \\
        \textsc{BioMistral-7B}          & 0 & 12.5 & \textbf{87.5} & 33.3 & 0 & \textbf{66.7} \\
        \textsc{BioMedGPT-LM-7B}         & \textbf{50.0} & 12.5 & 37.5 & 33.3 & 0 & \textbf{66.7} \\
        \midrule
        \textbf{Aggregate}                                 & \textbf{34.7} & 37.3 & 28.0 & 33.3 & 23.3 & \textbf{43.3} \\
        \bottomrule
    \end{tabular}
    }
\end{table}

In Figures \ref{fig:llm-acc-ci-clinical-logprob}(a) and (b), we show the absolute and relative exact-match accuracies achieved by the medical and general-domain LLMs on the textual clinical note QA datasets, from zero-shot and 3-shot prompting, respectively. 
In Figures \ref{fig:llm-acc-ci-sense-logprob}(a) and (b), we show the absolute and relative \textit{macro} exact-match accuracies and F1 scores---averaged over clinical acronyms---on the CASI and MIMIC-III sense disambiguation datasets, from zero-shot and 3-shot prompting, respectively.
In Table \ref{tab:win-tie-loss-rates-logprob-clinical}, we also show the win, tie, and loss rates (\%) of the medical LLMs, where win rate refers to the proportion of QA datasets where a medical model shows a statistically significant improvement over its base model. 
As discussed in Section \ref{sec:prompting}, we exclude the EHRNoteQA and i2b2 datasets from few-shot prompting evaluations, as most of the clinical notes already occupy the full context window for most models even in the zero-shot setting.
Additionally, we exclude the zero-shot results on these datasets for \textsc{MediTron-7B} (and its base model \textsc{Llama-2-7B}), as its context window size of 2k tokens is insufficient even in the zero-shot setting.

Overall, we observe that our findings discussed in Section \ref{sec:prompting-results-1-clinical} also hold in the constrained decoding setting. In particular, we observe that
\begin{enumerate}[topsep=0.5ex,itemsep=-0.5ex]
    \item the extent of performance improvement achieved by medical LLMs varies significantly across different model pairs and datasets;
    \item the results vary the most on the i2b2 datasets (which involve full-length, minimally preprocessed clinical notes), with \textsc{Med42-v1-70B} and \textsc{Clinical-Camel-70B} showing the largest improvements as in the greedy decoding case;
    \item only two medical LLMs---\textsc{Med42-v1-70B} and \textsc{Med42-v2-8B}---consistently show improvements in both zero-shot and 3-shot settings; and
    \item medical LLMs and general-domain LLMs are virtually indistinguishable in terms of the aggregate win/tie/loss rates.
\end{enumerate}
These results suggest that even when we constrain the outputs of each model to always produce a valid answer choice (which is not guaranteed with greedy decoding, especially in the zero-shot setting), the performance benefits from medical DAPT on textual clinical note QA datasets are overall limited in the zero-/few-shot prompting regime.

\subsection{Evaluation of Medical VLMs on Visual Medical QA (Section~\ref{sec:prompting-results-1-vqa})}
\label{sec:prompting-results-1-vqa-logprob}

\begin{figure*}[t!]
    \centering
    \begin{tabular}{@{}c@{}c@{\hskip 3pt}c@{}c@{}}
        \multicolumn{4}{c}{
            \begin{subfigure}{0.8\linewidth}
                \includegraphics[width=\linewidth]{figs/vlm-acc-ci-legend-v2.pdf}
            \end{subfigure}
        }
        \\
        \begin{subfigure}{0.03\linewidth}
            \makebox[\linewidth]{\raisebox{45pt}{{(a)}}}
        \end{subfigure} &
        \begin{subfigure}{0.47\linewidth}
            \includegraphics[width=\linewidth]{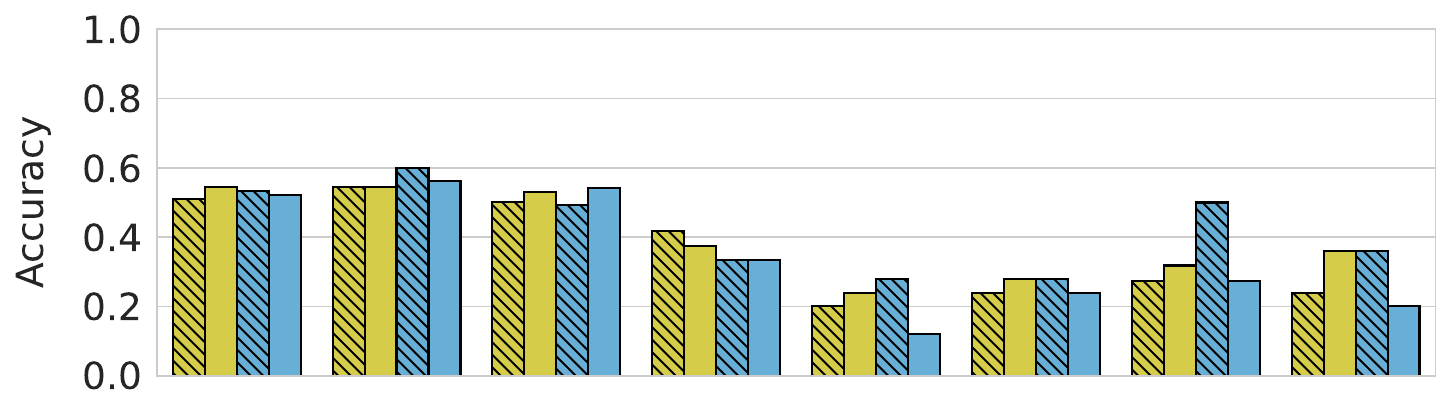}
        \end{subfigure} &
        \begin{subfigure}{0.03\linewidth}
            \makebox[\linewidth]{\raisebox{45pt}{{(b)}}}
        \end{subfigure} &
        \begin{subfigure}{0.47\linewidth}
            \includegraphics[width=\linewidth]{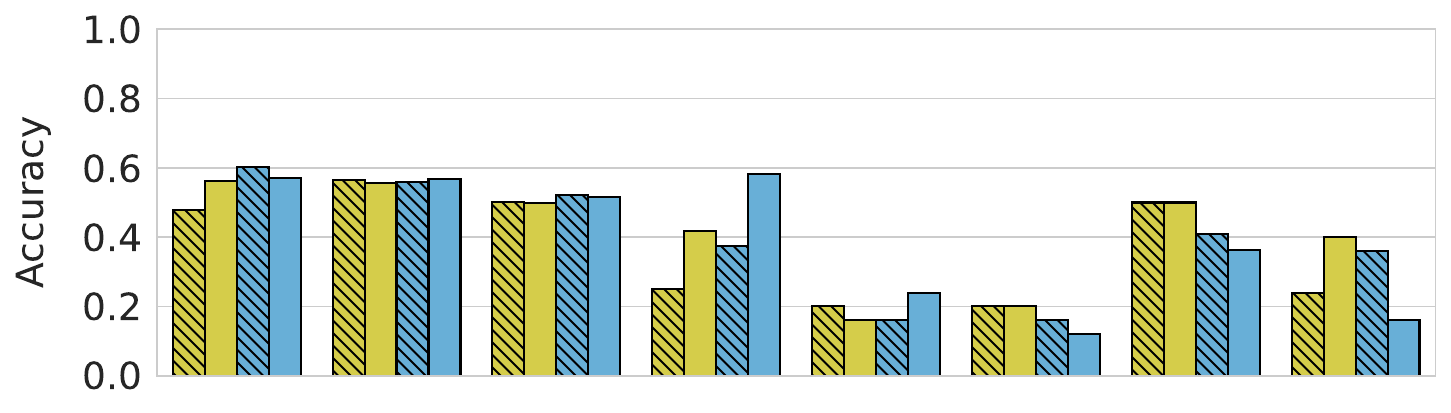}
        \end{subfigure}
        \\
        \begin{subfigure}{0.03\linewidth}
        \end{subfigure} &
        \begin{subfigure}{0.47\linewidth}
            \includegraphics[width=\linewidth]{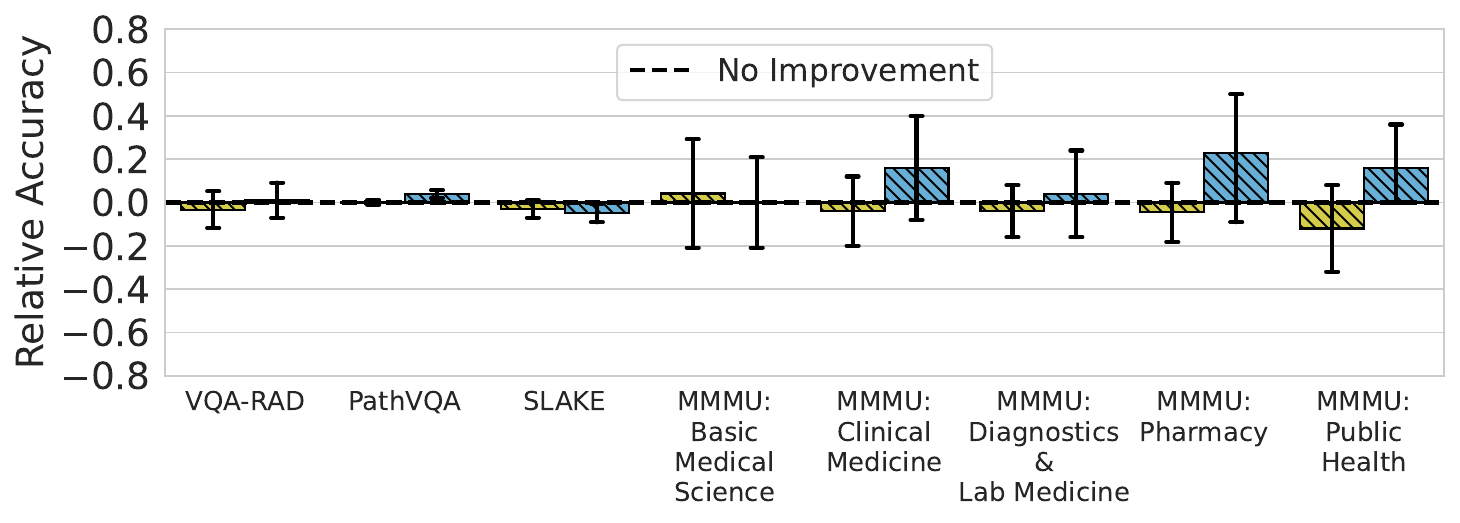}
        \end{subfigure} &
        \begin{subfigure}{0.03\linewidth}
        \end{subfigure} &
        \begin{subfigure}{0.47\linewidth}
            \includegraphics[width=\linewidth]{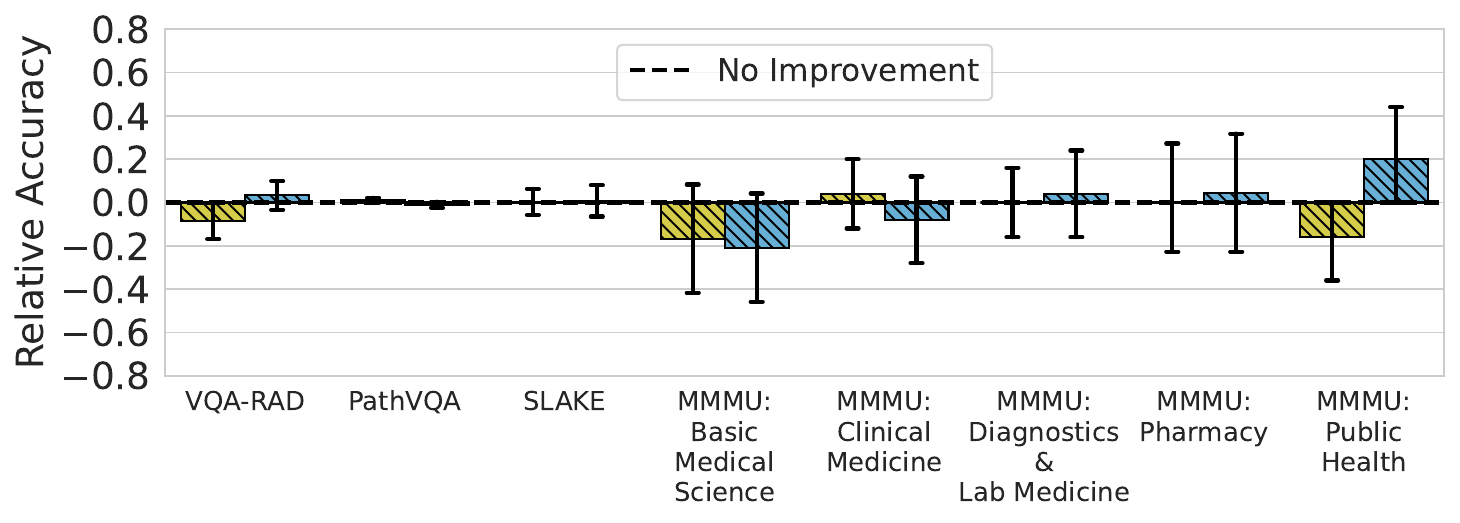}
        \end{subfigure}
    \end{tabular}
    \vspace{-10pt}
    \caption{
    After independently selecting the best prompt format and examples for each model, medical VLMs (textured bars) fail to consistently improve over their base models (solid bars) on visual medical QA tasks, in both (a) zero-shot and (b) 3-shot settings. 
    In each panel, the top row shows the absolute exact-match accuracies on the test set, and the bottom row shows the relative exact-match accuracies along with 95\% confidence intervals derived via bootstrapping on the test set (Section~\ref{sec:eval-setup}). 
    Here, model predictions are generated via constrained decoding. 
    Greedy decoding results are shown in Figure \ref{fig:vlm-acc-ci}.
    }
    \label{fig:vlm-logprob-acc-ci}
\end{figure*}

\begin{table}[t!]
    \centering
    \caption{The zero-shot and 3-shot win, tie, and loss rates (\%) of all medical VLMs on visual medical QA, after independently optimizing the prompt for each model. For each medical model, we boldface the win rate if it wins more than it loses to its general-domain base model, and vice versa. Here, we show the results when model predictions are generated via constrained decoding. The results for greedy decoding are shown in Table \ref{tab:win-tie-loss-rates-vlm}.}
    \label{tab:win-tie-loss-rates-logprob-vlm}
    \resizebox{0.65\linewidth}{!}{
    
    \begin{tabular}{@{}l@{\hskip 20pt}c@{\hskip 10pt}c@{\hskip 10pt}c@{\hskip 20pt}c@{\hskip 10pt}c@{\hskip 10pt}c@{\hskip 10pt}}
        \toprule
        \multirow{2}{*}{\textbf{Model}} & \multicolumn{3}{c}{\textbf{Zero-Shot}} & \multicolumn{3}{c}{\textbf{3-Shot}} \\
        \cmidrule{2-4} \cmidrule{5-7}
        & Win & Tie & Loss & Win & Tie & Loss \\
        \midrule
        \textsc{LLaVA-Med-7B}    & 0 & 100.0 & 0 & 0 & 87.5 & \textbf{12.5} \\
        \textsc{Med-Flamingo-9B} & 12.5 & 75.0 & 12.5 & 0 & 100.0 & 0 \\
        \midrule
        \textbf{Aggregate}                            & 6.3 & 87.5 & \textbf{6.3} & 0 & 93.8 & \textbf{6.3} \\
        \bottomrule
    \end{tabular}
    }
\end{table}

In Figures \ref{fig:vlm-logprob-acc-ci}(a) and (b), we show the absolute and relative exact-match accuracies achieved by the medical and general-domain VLMs on the visual medical QA datasets, from zero-shot and 3-shot prompting, respectively. 
Table \ref{tab:win-tie-loss-rates-logprob-vlm} shows the win, tie, and loss rates (\%) of the medical VLMs, where win rate refers to the proportion of visual medical QA datasets where a medical model shows a statistically significant improvement over its base model. 

Overall, we observe that our findings discussed in Section \ref{sec:prompting-results-1-vqa} also hold in the constrained decoding setting. In particular, we observe that
\begin{enumerate}[topsep=0.5ex,itemsep=-0.5ex]
    \item the absolute accuracies of the medical and general-domain models are generally similar on VQA-RAD, PathVQA, and SLAKE;
    \item the absolute accuracies do not consistently exhibit an increasing trend going from the zero-shot setting to the 3-shot setting; and
    \item the medical and general-domain VLMs are virtually indistinguishable in terms of the win/tie/loss rates (\%) in both zero-shot and 3-shot settings.
\end{enumerate}

These results suggest that even when we constrain the outputs of each model to always produce a valid answer choice (which is not guaranteed with greedy decoding, especially in the zero-shot setting), the performance benefits from medical DAPT on visual medical QA datasets are overall limited in the zero-/few-shot prompting regime.

\subsection{Ignoring Prompt Sensitivity and Statistical Uncertainty Can Overestimate the Performance Benefits from Medical DAPT (Section \ref{sec:prompting-results-2-aggregate})}
\label{sec:prompting-results-2-logprob}

\begin{figure*}[t!]
    \centering
    \includegraphics[width=\linewidth]{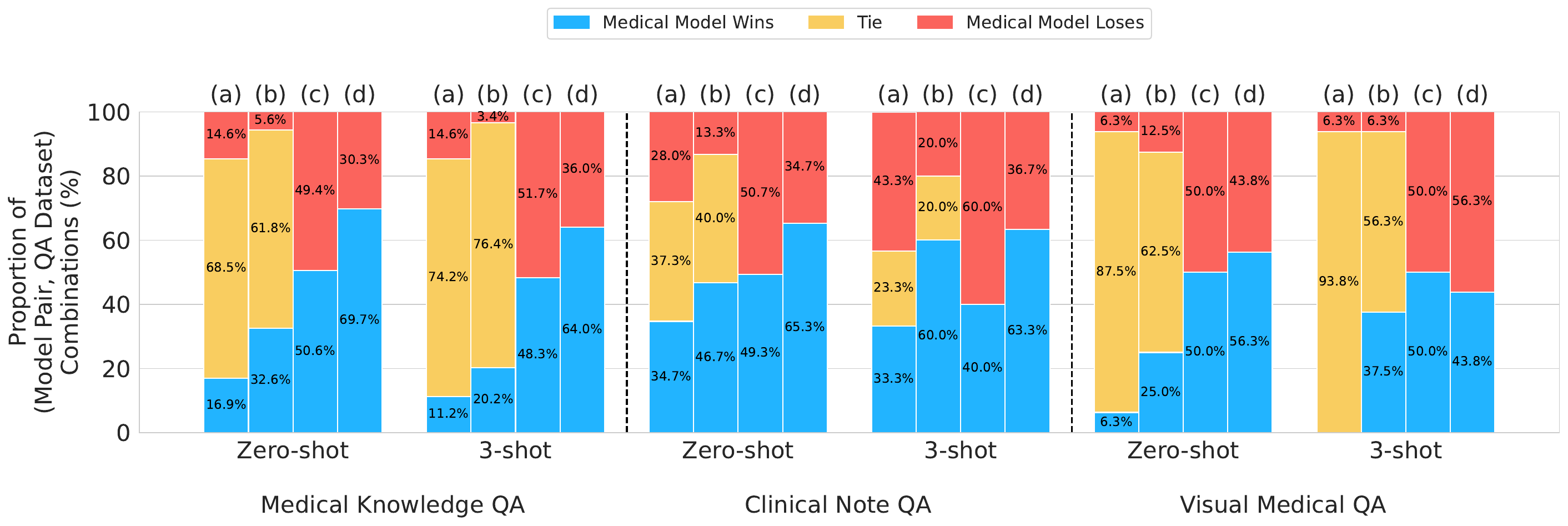}
    \vspace{-10pt}
    \caption{Optimizing the prompt for only the medical model and comparing models without accounting for statistical uncertainty can overestimate the performance improvements from medical DAPT. 
    We show the win/tie/loss rates (\%) of the medical models across all (model pair, QA dataset) combinations, when (a) optimizing the prompt for each model, with statistical testing; (b) optimizing the prompt only for the medical model, with statistical testing; (c) optimizing the prompt for each model, without statistical testing; and (d) optimizing the prompt only for the medical model, without statistical testing. 
    Here, we show the results for constrained decoding. 
    Greedy decoding results are shown in Figure \ref{fig:opt-ci-acc}.}
    \label{fig:opt-logprob-ci-acc}
\end{figure*}

In Figure \ref{fig:opt-logprob-ci-acc}, we show how the aggregate win/tie/loss rates (\%) of the medical models, computed over all (model pair, QA dataset) combinations, 
change as we vary the prompting setups as in Section \ref{sec:prompting-results-2-aggregate}. 
As with greedy decoding, we find that for both LLMs and VLMs, the performance improvements from medical DAPT can be substantially overestimated when (i) the prompt is only tailored to the medical model; and (ii) the models are compared only based on their absolute accuracies. 
For example, we observe that the zero-shot win rate substantially increases from 16.9\% to 68.0\% on textual medical knowledge QA, 32.5\% to 61.6\% on textual clinical note QA, and 6.3\% to 56.3\% on visual QA, when only performing prompt selection for the medical model and comparing based on raw absolute accuracy.
Thus, even when we constrain the outputs of each model always produce a valid answer choice (which is not guaranteed with greedy decoding, especially in the zero-shot setting), it is essential to account for LLM/VLM sensitivity to the prompting details and statistical uncertainty, in order to draw reliable conclusions about the effectiveness of medical DAPT.

\renewcommand\thetable{E\arabic{table}}
\renewcommand\thefigure{E\arabic{figure}}
\setcounter{table}{0}
\setcounter{figure}{0}
\section{Additional Details on Supervised Fine-Tuning (SFT)}\label{sec:details-sft}
\subsection{LoRA Fine-Tuning Details for LLMs} For all fine-tuning runs, we fix the $\alpha$ hyperparameter in LoRA to 16 and search over the ranks $r=[16,32,64]$ and the learning rates $\eta = [10^{-5},2 \times 10^{-5},5 \times 10^{-5},10^{-4},2 \times 10^{-4}]$, resulting in 15 trials per model and dataset.
For each run, we train the model for a maximum of 10 epochs and apply early stopping regularization with a patience of 1 epoch.
We track the validation cross-entropy loss on the output tokens after every epoch, and save the checkpoint that achieved the lowest validation loss during each fine-tuning run.
We then select the model that achieved the lowest validation loss across all trials to use for final evaluation on the test set.
Given the high variability in the remaining training details (e.g., per-device batch size $B$, the number of NVIDIA A6000 GPUs used for each run, the distributed training setup (DeepSpeed ZeRO Stage), use of QLoRA instead of LoRA) across different models and datasets, we detail them throughout Tables \ref{tab:sft-medqa-usmle}--\ref{tab:sft-obesity}, organized by dataset.

\subsection{LoRA Fine-Tuning Details for \textsc{LLaVA-Med-7B} and \textsc{LLaVA-v0-7B}} For all fine-tuning runs, we fix the $\alpha$ hyperparameter in LoRA to 16 and search over the ranks $r=[16,32,64]$, the learning rates $\eta=[10^{-5},2 \times 10^{-5}, 5 \times 10^{-5}]$, and LoRA dropout $p=[0,0.1]$, resulting in 18 trials per model and dataset.
We fix the per-device batch size $B$ to 32. 
For each run, we train the model for a maximum of 10 epochs and apply early stopping regularization with a patience of 1.
We track the validation cross-entropy loss on the output tokens after every epoch, and save the checkpoint that achieved the lowest validation loss during each fine-tuning run.
We then select the model with the lowest validation loss across all trials for final evaluation on the test set.
We train all models using Stage 2 of DeepSpeed ZeRO \citep{deepspeed-zero}, using 2 NVIDIA A6000 GPUs per trial.

\subsection{Fine-Tuning Details for \textsc{Med-Flamingo-9B} and \textsc{Open-Flamingo-9B}} For all fine-tuning runs, we search over the learning rates $\eta=[10^{-5},2\times 10^{-5},5 \times 10^{-5}]$ and the weight decay coefficients $\lambda=[0,0.05,0.1]$, resulting in 9 trials per model and dataset.
We fix the per-device batch size $B$ to 16.
For each run, we train the model for a maximum of 10 epochs and apply early stopping regularization with a patience of 1.
We track the validation cross-entropy loss on the output tokens after every epoch, and save the checkpoint that achieved the lowest validation loss during each fine-tuning run.
We then select the model with the lowest validation loss across all trials for final evaluation on the test set.
We train all models using PyTorch FSDP \citep{pt-fsdp}, using 6 NVIDIA A6000 GPUs per trial.

\noindent
\begin{minipage}{0.49\linewidth}
    \centering
    \captionof{table}{MedQA (4 Opt.) SFT configs.}
    \label{tab:sft-medqa-usmle}
    \resizebox{0.9\linewidth}{!}{
    \begin{tabular}{l@{\hskip 20pt}c@{\hskip 10pt}c@{\hskip 10pt}c}
        \toprule
        \textbf{Model} & $B$ & \# GPUs & ZeRO \\
        \midrule
        \textsc{OpenBioLLM-70B} & 8 & 8 & 3 \\
        \textsc{Llama-3-70B-Instruct} & 8 & 8 & 3 \\
        \textsc{MediTron-70B} & 8 & 8 & 3 \\
        \textsc{Llama-2-70B} & 8 & 8 & 3 \\
        \midrule
        \textsc{OpenBioLLM-8B} & 16 & 2 & 2 \\
        \textsc{Llama-3-8B} & 16 & 2 & 2 \\
        \midrule
        \textsc{MediTron-7B} & 32 & 2 & 2 \\
        \textsc{Llama-2-7B} & 32 & 2 & 2 \\
        \textsc{BioMistral-7B} & 32 & 2 & 2 \\
        \textsc{Mistral-7B-Instruct-v0.1} & 32 & 2 & 2 \\
        \textsc{BioMedGPT-LM-7B} & 32 & 2 & 2 \\
        \textsc{Llama-2-7B-Chat} & 32 & 2 & 2 \\
        \bottomrule
    \end{tabular}
    }
\end{minipage}
\hfill
\begin{minipage}{0.49\linewidth}
    \centering
    \captionof{table}{MedQA (5 Opt.) SFT configs.}
    \label{tab:sft-medqa}
    \resizebox{0.9\linewidth}{!}{
    \begin{tabular}{l@{\hskip 20pt}c@{\hskip 10pt}c@{\hskip 10pt}c}
        \toprule
        \textbf{Model} & $B$ & \# GPUs & ZeRO \\
        \midrule
        \textsc{OpenBioLLM-70B} & 8 & 8 & 3 \\
        \textsc{Llama-3-70B-Instruct} & 8 & 8 & 3 \\
        \textsc{MediTron-70B} & 8 & 8 & 3 \\
        \textsc{Llama-2-70B} & 8 & 8 & 3 \\
        \midrule
        \textsc{OpenBioLLM-8B} & 8 & 2 & 2 \\
        \textsc{Llama-3-8B} & 8 & 2 & 2 \\
        \midrule
        \textsc{MediTron-7B} & 32 & 2 & 2 \\
        \textsc{Llama-2-7B} & 32 & 2 & 2 \\
        \textsc{BioMistral-7B} & 32 & 2 & 2 \\
        \textsc{Mistral-7B-Instruct-v0.1} & 32 & 2 & 2 \\
        \textsc{BioMedGPT-LM-7B} & 32 & 2 & 2 \\
        \textsc{Llama-2-7B-Chat} & 32 & 2 & 2 \\
        \bottomrule
    \end{tabular}
    }
\end{minipage}

\vspace{10pt}

\noindent
\begin{minipage}{0.49\linewidth}
    \centering
    \captionof{table}[t]{MedMCQA SFT configs.}
    \label{tab:sft-medmcqa}
    \resizebox{0.9\linewidth}{!}{
    \begin{tabular}{l@{\hskip 20pt}c@{\hskip 10pt}c@{\hskip 10pt}c}
        \toprule
        \textbf{Model} & $B$ & \# GPUs & ZeRO \\
        \midrule
        \textsc{OpenBioLLM-70B} & 2 & 8 & 3 \\
        \textsc{Llama-3-70B-Instruct} & 2 & 8 & 3 \\
        \textsc{MediTron-70B} & 2 & 8 & 3 \\
        \textsc{Clinical-Camel-70B} & 2 & 8 & 3 \\
        \textsc{Llama-2-70B} & 2 & 8 & 3 \\
        \midrule
        \textsc{OpenBioLLM-8B} & 16 & 4 & 2 \\
        \textsc{Llama-3-8B} & 16 & 4 & 2 \\
        \midrule
        \textsc{MediTron-7B} & 32 & 4 & 2 \\
        \textsc{Llama-2-7B} & 32 & 4 & 2 \\
        \textsc{BioMistral-7B} & 32 & 4 & 2 \\
        \textsc{Mistral-7B-Instruct-v0.1} & 32 & 4 & 2 \\
        \textsc{BioMedGPT-LM-7B} & 32 & 4 & 2 \\
        \textsc{Llama-2-7B-Chat} & 32 & 4 & 2 \\
        \bottomrule
    \end{tabular}
    }
\end{minipage}
\hfill
\begin{minipage}{0.49\linewidth}
    \centering
    \captionof{table}[t]{PubMedQA SFT configs.}
    \label{tab:sft-pubmedqa}
    \resizebox{0.95\linewidth}{!}{
    \begin{tabular}{l@{\hskip 20pt}c@{\hskip 10pt}c@{\hskip 10pt}c}
        \toprule
        \textbf{Model} & $B$ & \# GPUs & ZeRO \\
        \midrule
        \textsc{OpenBioLLM-70B} (QLoRA) & 2 & 8 & 2 \\
        \textsc{Llama-3-70B-Instruct} (QLoRA) & 2 & 8 & 2 \\
        \textsc{MediTron-70B} (QLoRA) & 2 & 8 & 2 \\
        \textsc{Clinical-Camel-70B} (QLoRA) & 2 & 8 & 2 \\
        \textsc{Llama-2-70B} (QLoRA) & 2 & 8 & 2\\
        \midrule
        \textsc{OpenBioLLM-8B} & 8 & 4 & 2 \\
        \textsc{Llama-3-8B} & 8 & 4 & 2 \\
        \midrule
        \textsc{MediTron-7B} & 32 & 4 & 2 \\
        \textsc{Llama-2-7B} & 32 & 4 & 2 \\
        \textsc{BioMistral-7B} & 32 & 4 & 2 \\
        \textsc{Mistral-7B-Instruct-v0.1} & 32 & 4 & 2 \\
        \textsc{BioMedGPT-LM-7B} & 32 & 4 & 2 \\
        \textsc{Llama-2-7B-Chat} & 32 & 4 & 2 \\
        \bottomrule
    \end{tabular}
    }
\end{minipage}

\noindent
\begin{minipage}{0.49\linewidth}
    \centering
    \captionof{table}{MedNLI SFT configs.}
    \label{tab:sft-mednli}
    \resizebox{0.85\linewidth}{!}{
    \begin{tabular}{l@{\hskip 20pt}c@{\hskip 10pt}c@{\hskip 10pt}c}
        \toprule
        \textbf{Model} & $B$ & \# GPUs & ZeRO \\
        \midrule
        \textsc{Med42-v2-70B} & 8 & 8 & 3 \\
        \textsc{OpenBioLLM-70B} & 8 & 8 & 3 \\
        \textsc{Llama-3-70B-Instruct} & 8 & 8 & 3 \\
        \textsc{MediTron-70B} & 8 & 8 & 3 \\
        \textsc{Clinical-Camel-70B} & 8 & 8 & 3 \\
        \textsc{Med42-v1-70B} & 8 & 8 & 3 \\
        \textsc{Llama-2-70B} & 8 & 8 & 3 \\
        \midrule
        \textsc{Med42-v2-8B} & 16 & 2 & 2 \\
        \textsc{Llama-3-8B-Instruct} & 16 & 2 & 2 \\
        \textsc{OpenBioLLM-8B} & 16 & 2 & 2 \\
        \textsc{Llama-3-8B} & 16 & 2 & 2 \\
        \midrule
        \textsc{MediTron-7B} & 32 & 2 & 2 \\
        \textsc{Llama-2-7B} & 32 & 2 & 2 \\
        \textsc{BioMistral-7B} & 32 & 2 & 2 \\
        \textsc{Mistral-7B-Instruct-v0.1} & 32 & 2 & 2 \\
        \textsc{BioMedGPT-LM-7B} & 32 & 2 & 2 \\
        \textsc{Llama-2-7B-Chat} & 32 & 2 & 2 \\
        \bottomrule
    \end{tabular}
    }
\end{minipage}
\hfill
\begin{minipage}{0.49\linewidth}
    \centering
    \captionof{table}{CASI SFT configs.}
    \label{tab:sft-casi}
    \resizebox{0.85\linewidth}{!}{
    \begin{tabular}{l@{\hskip 20pt}c@{\hskip 10pt}c@{\hskip 10pt}c}
        \toprule
        \textbf{Model} & $B$ & \# GPUs & ZeRO \\
        \midrule
        \textsc{Med42-v2-70B} & 8 & 8 & 3 \\
        \textsc{OpenBioLLM-70B} & 8 & 8 & 3 \\
        \textsc{Llama-3-70B-Instruct} & 8 & 8 & 3 \\
        \textsc{MediTron-70B} & 8 & 8 & 3 \\
        \textsc{Clinical-Camel-70B} & 8 & 8 & 3 \\
        \textsc{Med42-v1-70B} & 8 & 8 & 3 \\
        \textsc{Llama-2-70B} & 8 & 8 & 3 \\
        \midrule
        \textsc{Med42-v2-8B} & 8 & 2 & 2 \\
        \textsc{Llama-3-8B-Instruct} & 8 & 2 & 2 \\
        \textsc{OpenBioLLM-8B} & 8 & 2 & 2 \\
        \textsc{Llama-3-8B} & 8 & 2 & 2 \\
        \midrule
        \textsc{MediTron-7B} & 16 & 2 & 2 \\
        \textsc{Llama-2-7B} & 16 & 2 & 2 \\
        \textsc{BioMistral-7B} & 16 & 2 & 2 \\
        \textsc{Mistral-7B-Instruct-v0.1} & 16 & 2 & 2 \\
        \textsc{BioMedGPT-LM-7B} & 16 & 2 & 2 \\
        \textsc{Llama-2-7B-Chat} & 16 & 2 & 2 \\
        \bottomrule
    \end{tabular}
    }
\end{minipage}

\vspace{10pt}

\noindent
\begin{minipage}[t]{0.49\linewidth}
    \centering
    \captionof{table}{MIMIC-III SFT configs.}
    \label{tab:sft-mimic-iii}
    \resizebox{0.85\linewidth}{!}{
    \begin{tabular}{l@{\hskip 20pt}c@{\hskip 10pt}c@{\hskip 10pt}c}
        \toprule
        \textbf{Model} & $B$ & \# GPUs & ZeRO \\
        \midrule
        \textsc{Med42-v2-70B} & 8 & 8 & 3 \\
        \textsc{OpenBioLLM-70B} & 8 & 8 & 3 \\
        \textsc{Llama-3-70B-Instruct} & 8 & 8 & 3 \\
        \textsc{MediTron-70B} & 8 & 8 & 3 \\
        \textsc{Clinical-Camel-70B} & 8 & 8 & 3 \\
        \textsc{Med42-v1-70B} & 8 & 8 & 3 \\
        \textsc{Llama-2-70B} & 8 & 8 & 3 \\
        \midrule
        \textsc{Med42-v2-8B} & 8 & 2 & 2 \\
        \textsc{Llama-3-8B-Instruct} & 8 & 2 & 2 \\
        \textsc{OpenBioLLM-8B} & 8 & 2 & 2 \\
        \textsc{Llama-3-8B} & 8 & 2 & 2 \\
        \midrule
        \textsc{MediTron-7B} & 32 & 2 & 2 \\
        \textsc{Llama-2-7B} & 32 & 2 & 2 \\
        \textsc{BioMistral-7B} & 32 & 2 & 2 \\
        \textsc{Mistral-7B-Instruct-v0.1} & 32 & 2 & 2 \\
        \textsc{BioMedGPT-LM-7B} & 32 & 2 & 2 \\
        \textsc{Llama-2-7B-Chat} & 32 & 2 & 2 \\
        \bottomrule
    \end{tabular}
    }
\end{minipage}
\hfill
\begin{minipage}[t]{0.49\linewidth}
    \centering
    \captionof{table}{EHRNoteQA SFT configs.}
    \label{tab:sft-ehrnoteqa}
    \resizebox{0.85\linewidth}{!}{
    \begin{tabular}{l@{\hskip 20pt}c@{\hskip 10pt}c@{\hskip 10pt}c}
        \toprule
        \textbf{Model} & $B$ & \# GPUs & ZeRO \\
        \midrule
        \textsc{Med42-v2-70B} & 4 & 8 & 3 \\
        \textsc{OpenBioLLM-70B} & 4 & 8 & 3 \\
        \textsc{Llama-3-70B-Instruct} & 4 & 8 & 3 \\
        \textsc{MediTron-70B} & 4 & 8 & 3 \\
        \textsc{Clinical-Camel-70B} & 4 & 8 & 3 \\
        \textsc{Med42-v1-70B} & 4 & 8 & 3 \\
        \textsc{Llama-2-70B} & 4 & 8 & 3 \\
        \midrule
        \textsc{Med42-v2-8B} & 4 & 2 & 2 \\
        \textsc{Llama-3-8B-Instruct} & 4 & 2 & 2 \\
        \textsc{OpenBioLLM-8B} & 4 & 2 & 2 \\
        \textsc{Llama-3-8B} & 4 & 2 & 2 \\
        \midrule
        \textsc{BioMistral-7B} & 32 & 2 & 2 \\
        \textsc{Mistral-7B-Instruct-v0.1} & 32 & 2 & 2 \\
        \textsc{BioMedGPT-LM-7B} & 32 & 2 & 2 \\
        \textsc{Llama-2-7B-Chat} & 32 & 2 & 2 \\
        \bottomrule
    \end{tabular}
    }
\end{minipage}

\vspace{10pt}

\noindent
\begin{minipage}[t]{0.49\linewidth}
    \centering
    \captionof{table}{i2b2 (Asthma) SFT configs.}
    \label{tab:sft-asthma}
    \resizebox{0.85\linewidth}{!}{
    \begin{tabular}{l@{\hskip 20pt}c@{\hskip 10pt}c@{\hskip 10pt}c}
        \toprule
        \textbf{Model} & $B$ & \# GPUs & ZeRO \\
        \midrule
        \textsc{Med42-v2-70B} & 2 & 8 & 3 \\
        \textsc{OpenBioLLM-70B} & 2 & 8 & 3 \\
        \textsc{Llama-3-70B-Instruct} & 2 & 8 & 3 \\
        \textsc{MediTron-70B} & 2 & 8 & 3 \\
        \textsc{Clinical-Camel-70B} & 2 & 8 & 3 \\
        \textsc{Med42-v1-70B} & 2 & 8 & 3 \\
        \textsc{Llama-2-70B} & 2 & 8 & 3 \\
        \midrule
        \textsc{Med42-v2-8B} & 2 & 2 & 2 \\
        \textsc{Llama-3-8B-Instruct} & 2 & 2 & 2 \\
        \textsc{OpenBioLLM-8B} & 2 & 2 & 2 \\
        \textsc{Llama-3-8B} & 2 & 2 & 2 \\
        \midrule
        \textsc{BioMistral-7B} & 4 & 2 & 2 \\
        \textsc{Mistral-7B-Instruct-v0.1} & 4 & 2 & 2 \\
        \textsc{BioMedGPT-LM-7B} & 4 & 2 & 2 \\
        \textsc{Llama-2-7B-Chat} & 4 & 2 & 2 \\
        \bottomrule
    \end{tabular}
    }
\end{minipage}
\hfill
\begin{minipage}[t]{0.49\linewidth}
    \centering
    \captionof{table}{i2b2 (CAD) SFT configs.}
    \label{tab:sft-cad}
    \resizebox{0.85\linewidth}{!}{
    \begin{tabular}{l@{\hskip 20pt}c@{\hskip 10pt}c@{\hskip 10pt}c}
        \toprule
        \textbf{Model} & $B$ & \# GPUs & ZeRO \\
        \midrule
        \textsc{Med42-v2-70B} & 2 & 8 & 3 \\
        \textsc{OpenBioLLM-70B} & 2 & 8 & 3 \\
        \textsc{Llama-3-70B-Instruct} & 2 & 8 & 3 \\
        \textsc{MediTron-70B} & 2 & 8 & 3 \\
        \textsc{Clinical-Camel-70B} & 2 & 8 & 3 \\
        \textsc{Med42-v1-70B} & 2 & 8 & 3 \\
        \textsc{Llama-2-70B} & 2 & 8 & 3 \\
        \midrule
        \textsc{Med42-v2-8B} & 2 & 2 & 2 \\
        \textsc{Llama-3-8B-Instruct} & 2 & 2 & 2 \\
        \textsc{OpenBioLLM-8B} & 2 & 2 & 2 \\
        \textsc{Llama-3-8B} & 2 & 2 & 2 \\
        \midrule
        \textsc{BioMistral-7B} & 4 & 2 & 2 \\
        \textsc{Mistral-7B-Instruct-v0.1} & 4 & 2 & 2 \\
        \textsc{BioMedGPT-LM-7B} & 4 & 2 & 2 \\
        \textsc{Llama-2-7B-Chat} & 4 & 2 & 2 \\
        \bottomrule
    \end{tabular}
    }
\end{minipage}

\vspace{10pt}

\noindent
\begin{minipage}[t]{0.49\linewidth}
    \centering
    \captionof{table}{i2b2 (Diabetes) SFT configs.}
    \label{tab:sft-diabetes}
    \resizebox{0.85\linewidth}{!}{
    \begin{tabular}{l@{\hskip 20pt}c@{\hskip 10pt}c@{\hskip 10pt}c}
        \toprule
        \textbf{Model} & $B$ & \# GPUs & ZeRO \\
        \midrule
        \textsc{Med42-v2-70B} (QLoRA) & 1 & 8 & 3 \\
        \textsc{OpenBioLLM-70B} (QLoRA) & 1 & 8 & 3 \\
        \textsc{Llama-3-70B-Instruct} (QLoRA) & 1 & 8 & 3 \\
        \textsc{MediTron-70B} (QLoRA) & 1 & 8 & 3 \\
        \textsc{Clinical-Camel-70B} (QLoRA) & 1 & 8 & 3 \\
        \textsc{Med42-v1-70B} (QLoRA) & 1 & 8 & 3 \\
        \textsc{Llama-2-70B} (QLoRA) & 1 & 8 & 3 \\
        \midrule
        \textsc{Med42-v2-8B} & 2 & 2 & 2 \\
        \textsc{Llama-3-8B-Instruct} & 2 & 2 & 2 \\
        \textsc{OpenBioLLM-8B} & 2 & 2 & 2 \\
        \textsc{Llama-3-8B} & 2 & 2 & 2 \\
        \midrule
        \textsc{BioMistral-7B} & 4 & 2 & 2 \\
        \textsc{Mistral-7B-Instruct-v0.1} & 4 & 2 & 2 \\
        \textsc{BioMedGPT-LM-7B} & 4 & 2 & 2 \\
        \textsc{Llama-2-7B-Chat} & 4 & 2 & 2 \\
        \bottomrule
    \end{tabular}
    }
\end{minipage}
\hfill
\begin{minipage}[t]{0.49\linewidth}
    \centering
    \captionof{table}{i2b2 (Obesity) SFT configs.}
    \label{tab:sft-obesity}
    \resizebox{0.85\linewidth}{!}{
    \begin{tabular}{l@{\hskip 20pt}c@{\hskip 10pt}c@{\hskip 10pt}c}
        \toprule
        \textbf{Model} & $B$ & \# GPUs & ZeRO \\
        \midrule
        \textsc{Med42-v2-70B} & 1 & 8 & 3 \\
        \textsc{OpenBioLLM-70B} & 1 & 8 & 3 \\
        \textsc{Llama-3-70B-Instruct} & 1 & 8 & 3 \\
        \textsc{MediTron-70B} & 1 & 8 & 3 \\
        \textsc{Clinical-Camel-70B} & 1 & 8 & 3 \\
        \textsc{Med42-v1-70B} & 1 & 8 & 3 \\
        \textsc{Llama-2-70B} & 1 & 8 & 3 \\
        \midrule
        \textsc{Med42-v2-8B} & 2 & 2 & 2 \\
        \textsc{Llama-3-8B-Instruct} & 2 & 2 & 2 \\
        \textsc{OpenBioLLM-8B} & 2 & 2 & 2 \\
        \textsc{Llama-3-8B} & 2 & 2 & 2 \\
        \midrule
        \textsc{BioMistral-7B} & 4 & 2 & 2 \\
        \textsc{Mistral-7B-Instruct-v0.1} & 4 & 2 & 2 \\
        \textsc{BioMedGPT-LM-7B} & 4 & 2 & 2 \\
        \textsc{Llama-2-7B-Chat} & 4 & 2 & 2 \\
        \bottomrule
    \end{tabular}
    }
\end{minipage}

\clearpage
\bibliography{ref}

\end{document}